\documentclass[twoside,11pt]{article}

\usepackage{multirow,tikz,amsmath,comment}
\usepackage{algorithm}
\usepackage{algorithmic}
\usepackage{subfig}
\usepackage{enumitem}
\usepackage{epstopdf}
\usepackage{tabularx,booktabs}
\usepackage{lastpage}
\usepackage{caption}
\usepackage{microtype}
\usepackage{jmlr2e}
\usepackage{blindtext}

\usepackage{tikz,pgfplots}
\usepackage{amsmath}
\usepackage{mathtools}
\usepackage{amssymb}

\usepackage{threeparttable}

\usepackage{float}

\usepackage{wrapfig}

\newtheorem{Theorem}{Theorem}
\newtheorem{Proposition}{Proposition}
\newtheorem{Lemma}{Lemma}

\newtheorem{Definition}{Definition}
\newtheorem{Corollary}{Corollary}

\newtheorem{Example-set}{Example}
\newtheorem{Condition}{Condition}

\newtheorem{Remark}{Remark}

\usepackage{bbding}

\usepackage{centernot}

\newcommand{\CI}{\mathrel{\perp\mspace{-10mu}\perp}}
\newcommand{\nCI}{\centernot{\CI}}

\usetikzlibrary{fit,matrix,positioning, 
decorations.pathreplacing,calc,decorations,arrows,shadows,patterns}

\usetikzlibrary{arrows,fit,positioning,shapes,matrix}

\usetikzlibrary{backgrounds,shapes.geometric} 

\makeatletter
\newcommand{\srcsize}{\@setfontsize{\srcsize}{4.5pt}{4.5pt}}
\makeatother

\mathchardef\mhyphen="2D

\tikzset{
mycirclestyle/.style={
  circle,
  thin,
  inner sep=0pt,
  text width=6mm,
  minimum size=4mm,
  inner sep=0pt,
  align=center,
  draw=gray,
  fill=gray
  }
}
\tikzstyle{main node}=[circle,draw,minimum size=0.5mm]

\newenvironment{nospaceflalign}
 {\setlength{\abovedisplayskip}{1pt}\setlength{\belowdisplayskip}{1pt}%
  \csname flalign\endcsname}
 {\csname endflalign\endcsname\ignorespacesafterend}

\newdimen\arrowsize
\pgfarrowsdeclare{arcsq}{arcsq}
{
	\arrowsize=0.2pt
	\advance\arrowsize by .5\pgflinewidth
	\pgfarrowsleftextend{-4\arrowsize-.5\pgflinewidth}
	\pgfarrowsrightextend{.5\pgflinewidth}
}
{
	\arrowsize=0.8pt
	\advance\arrowsize by .5\pgflinewidth
	\pgfsetdash{}{0pt} 
	\pgfsetroundjoin   
	\pgfsetroundcap    
	\pgfpathmoveto{\pgfpoint{0\arrowsize}{0\arrowsize}}
	\pgfpatharc{-90}{-140}{4\arrowsize}
	\pgfusepathqstroke
	\pgfpathmoveto{\pgfpointorigin}
	\pgfpatharc{90}{140}{4\arrowsize}
	\pgfusepathqstroke
}

\newcommand{\feng}[1]{{\color{black} #1}}

\newcommand{\revision}[1]{{\color{black}  #1}}

\newcommand{\Revitwo}[1]{{\color{black}  #1}}

\newcommand*\samethanks[1][\value{footnote}]{\footnotemark[#1]}

\usepackage{lastpage}
\jmlrheading{25}{2024}{1-\pageref{LastPage}}{8/23; Revised
4/24}{6/24}{23-1052}{Feng Xie, Biwei Huang, Zhengming Chen, Ruichu Cai, Clark Glymour, Zhi Geng, and Kun Zhang}

\ShortHeadings{GIN Condition for Estimating Causal Structure with Latent Variables}
{Xie, Huang, Chen, Cai, Glymour, Geng, and Zhang}
\firstpageno{1}

\begin{document}

\title{Generalized Independent Noise Condition for Estimating Causal Structure with Latent Variables}

\author{\name Feng Xie\thanks{Equal contribution} \email     fengxie@btbu.edu.cn\\
 \addr Department of Applied Statistics, Beijing Technology and Business University\\Beijing, 102488, China
            \AND
 \name Biwei Huang\samethanks \email   bih007@ucsd.edu\\ 
 \addr Halicioglu Data Science Institute (HDSI), University of California San Diego\\
 La Jolla, San Diego, California, 92093, USA
		\AND
            \name Zhengming Chen  \email chenzhengming1103@gmail.com\\
 \addr School of Computer Science, Guangdong University of Technology\\ Guangzhou, 510006, China\\
\addr Machine Learning Department, Mohamed bin Zayed University of Artificial Intelligence\\ Abu Dhabi, UAE
        \AND
        \name Ruichu Cai  \email cairuichu@gmail.com\\
\addr School of Computer Science, Guangdong University of Technology\\ Guangzhou, 510006, China
            \AND
            \name Clark Glymour \email cg09@andrew.cmu.edu\\ 
\addr Department of Philosophy,
		Carnegie Mellon University\\
		Pittsburgh, PA 15213, USA
            \AND
            \name Zhi Geng \email zhigeng@btbu.edu.cn\\
 \addr Department of Applied Statistics, Beijing Technology and Business University\\ Beijing, 102488, China
		\AND 
            \name Kun Zhang \thanks{Corresponding author} \email kunz1@cmu.edu\\ 
	    \addr Department of Philosophy, 
	    Carnegie Mellon University\\
	    Pittsburgh, PA 15213, USA\\
            \addr Machine Learning Department, Mohamed bin Zayed University of Artificial Intelligence\\ Abu Dhabi, UAE
	   }
\editor{Ilya Shpitser}

\maketitle

\begin{abstract}
We investigate the challenging task of learning causal structure in the presence of latent variables, including locating latent variables, determining their quantity, and identifying causal relationships among both latent and observed variables. To address this, we propose a Generalized Independent Noise (GIN) condition for linear non-Gaussian acyclic causal models that incorporate latent variables, which establishes the independence between a linear combination of certain measured variables and some other measured variables. Specifically, for two observed random vectors
$\bf{Y}$ and $\bf{Z}$, GIN holds if and only if $\omega^{\intercal}\mathbf{Y}$ and $\mathbf{Z}$ are statistically independent, where $\omega$ is a non-zero parameter vector determined by the cross-covariance between $\mathbf{Y}$ and $\mathbf{Z}$. We then give necessary and sufficient graphical criteria of the GIN condition in linear non-Gaussian acyclic causal models. From a graphical perspective, roughly speaking, 
\revision{GIN implies the existence of a set $\mathcal{S}$ such that $\mathcal{S}$ is causally earlier (w.r.t. the causal ordering) than $\mathbf{Y}$, and that every
active (collider-free) path between $\mathbf{Y}$ and $\mathbf{Z}$ must contain a node from $\mathcal{S}$.}
Interestingly, we find that the independent noise condition (i.e., if there is no confounder, causes are independent of the residual derived from regressing the effect on the causes) can be seen as a special case of GIN. With such a connection between GIN and latent causal structures,  we further leverage the proposed GIN condition, together with a well-designed search procedure, to efficiently estimate Linear, Non-Gaussian Latent Hierarchical Models (LiNGLaHs), where latent confounders may also be causally related and may even follow a hierarchical structure. 
We show that the underlying causal structure of a LiNGLaH is identifiable in light of GIN conditions under mild assumptions. 
Experimental results on both synthetic and three real-world data sets show the effectiveness of the proposed approach.
\end{abstract}


\section{Introduction}
Discovering causal relationships from observational (non-experimental) data, known as causal discovery, has received much attention over the past two decades~\citep{pearl2009causality,spirtes2000causation,peters2017elements} and has played a key role in understanding system mechanisms, such as explanation, prediction, decision making, and control~\citep{sachs2005causal,spirtes2010introduction,yu2019multi,shi2021temporal,zhang2018learning}. Most causal discovery approaches focus on the situation without latent variables, such as the PC algorithm~\citep{spirtes1991PC}, Greedy Equivalence Search (GES)~\citep{chickering2002optimal}, and
methods based on the Linear, Non-Gaussian Acyclic Model (LiNGAM)~\citep{shimizu2006linear}, the linear-Gaussian causal model with equal error variances \citep{peters2014identifiability}, the Additive Noise Model (ANM)~\citep{hoyer2009ANM,peters2011causal}, and the Post-NonLinear causal model (PNL)~\citep{Zhang06_iconip, zhang2009PNL}. 
However, although these methods have been used in a range of fields, they may fail to produce convincing results in cases with latent variables (or more specifically, confounders), without properly taking into account the influences from latent variables as well as other practical issues \citep{spirtes2016causal}.

Existing methods for causal discovery with latent variables usually use the following two types of strategies. One typical strategy to handle this problem is by utilizing conditional independence relations to learn the causal graph over the observed variables up to an equivalence class ~\citep{pearl2009causality,spirtes2000causation}. Well-known algorithms along this line include FCI~\citep{spirtes1995causal}, RFCI~\citep{colombo2012learning}, and FCI+~\citep{claassen2013learning}. Another strategy involves utilizing the data-generating mechanism-based approaches in the linear setting, such as non-Gaussianity-based methods~\citep{hoyer2008estimation,entner2010discovering,chen2013causality,tashiro2014parcelingam,shimizu2014bayesian,wang2020causal,salehkaleybar2020learning,maeda2020rcd}, sparse plus low-rank matrix decomposition-based approaches~\citep{RankSparsity_11, RankSparsity_12,frot2019robust}. These approaches focus on estimating causal relationships among observed variables rather than those among latent variables. However, in some real-world scenarios, the observed variables may not necessarily be the underlying causal variables~\citep{bollen1989structural,Bartholomewbook}, making it desirable to develop approaches capable of locating latent variables and identifying causal relationships among them as well.

On the other hand, Factor Analysis is a classical framework commonly used for inferring latent factors \citep{Bartholomewbook}. However, with factor analysis-based approaches, the estimated factors may not necessarily correspond to the underlying causal variables, and their relationships are often not explicitly modeled \citep{Silva-linearlvModel}. Later, it was shown that by utilizing vanishing Tetrad conditions \citep{spearman1928pearson} and, more generally, t-separation~\citep{Sullivant-T-separation}, one can identify latent variables in linear-Gaussian models \citep{Silva-linearlvModel}. 
\revision{Furthermore, by leveraging an extended version of t-separation \citep{spirtes2013calculation-t-separation}, two more efficient and faster algorithms were developed: the FindOneFactorClusters (FOFC) algorithm for one-factor models \citep{Kummerfeld2016} and the FindTwoFactorClusters (FTFC) algorithm for two-factor models \citep{kummerfeld2014causal}.
However, these methods may not be capable of identifying causal directions between latent variables and they impose strong constraints on the graph structure, requiring that each observed variable has only one or two latent parents and that each latent variable (in one-factor models) has at least three pure observed variables.}
This limitation arises from their reliance solely on second-order information from measured variables, without considering higher-order statistics. 

\begin{figure}[htp]
	\begin{center}
		\begin{tikzpicture}[scale=1.3, line width=0.5pt, inner sep=0.2mm, shorten >=.1pt, shorten <=.1pt]
		\draw (2, 2.4) node(L1) [circle, fill=gray!60,draw] {{\footnotesize\,$L_1$\,}};
		\draw (1, 1.6) node(L2) [circle, fill=gray!60,draw] {{\footnotesize\,$L_2$\,}};
		\draw (1.8, 1.6) node(L3) [circle, fill=gray!60,draw] {{\footnotesize\,$L_3$\,}};
		\draw (3.0, 1.6) node(L4) [circle, fill=gray!60,draw] {{\footnotesize\,$L_4$\,}};
		
		\draw (-0.0, 0.8) node(X12) [] {{\footnotesize\,$X_{12}$\,}};
		\draw (0.6, 0.8) node(L5) [circle, fill=gray!60,draw] {{\footnotesize\,$L_5$\,}};
		\draw (1.3, 0.8) node(L6) [circle, fill=gray!60,draw] {{\footnotesize\,{$L_6$}\,}};
		\draw (2.1, 0.8) node(L7) [circle, fill=gray!60,draw] {{\footnotesize\,{$L_7$}\,}};
		\draw (2.8, 0.8) node(L8) [circle, fill=gray!60,draw] {{\footnotesize\,$L_8$\,}};
		\draw (3.5, 0.8) node(L9) [circle, fill=gray!60,draw] {{\footnotesize\,$L_9$\,}};
		\draw (4.1, 0.8) node(X13) [] {{\footnotesize\,$X_{13}$\,}};
		\draw (-0.4, 0) node(X1) [] {{\footnotesize\,$X_1$\,}};
		\draw (0.2, 0) node(X2) [] {{\footnotesize\,$X_2$\,}};
		\draw (0.8, 0) node(X3) [] {{\footnotesize\,$X_3$\,}};
		\draw (1.3,0) node(X4) [] {{\footnotesize\,$X_4$\,}};
		\draw (1.8, 0) node(X5) [] {{\footnotesize\,$X_5$\,}};
		\draw (2.3, 0) node(X6) [] {{\footnotesize\,$X_6$\,}};
		\draw (2.7, 0) node(X7) [] {{\footnotesize\,$X_7$\,}};
		\draw (3.2, 0) node(X8) [] {{\footnotesize\,$X_8$\,}};
		\draw (3.6, 0) node(X9) [] {{\footnotesize\,$X_9$\,}};
		\draw (4.2, 0) node(X10) [] {{\footnotesize\,$X_{10}$\,}};
		\draw (4.9, 0) node(X11) [] {{\footnotesize\,$X_{11}$\,}};
		\draw[-arcsq] (L1) -- (L2) node[pos=0.5,sloped,above] {};
		\draw[-arcsq] (L1) -- (L3) node[pos=0.5,sloped,above] {};
		\draw[-arcsq] (L1) -- (L4) node[pos=0.5,sloped,above] {};

		\draw[-arcsq] (L1) -- (L8) node[pos=0.5,sloped,above] {};
		\draw[-arcsq] (L2) -- (X12) node[pos=0.5,sloped,above] {};
		\draw[-arcsq] (L2) -- (L5) node[pos=0.5,sloped,above] {};
		\draw[-arcsq] (L2) -- (L6) node[pos=0.5,sloped,above] {};
		\draw[-arcsq] (L3) -- (L5) node[pos=0.5,sloped,above] {};
		\draw[-arcsq] (L3) -- (L6) node[pos=0.5,sloped,above] {};
		\draw[-arcsq] (L3) -- (L7) node[pos=0.5,sloped,above] {};
		\draw[-arcsq] (L4) -- (L8) node[pos=0.5,sloped,above] {};
		\draw[-arcsq] (L4) -- (L9) node[pos=0.5,sloped,above] {};
		\draw[-arcsq] (L4) -- (X13) node[pos=0.5,sloped,above] {};
		\draw[-arcsq] (L5) -- (X1) node[pos=0.5,sloped,above] {};
		\draw[-arcsq] (L5) -- (X2) node[pos=0.5,sloped,above] {};
		\draw[-arcsq] (L5) -- (X3) node[pos=0.5,sloped,above] {};
		\draw[-arcsq] (L6) -- (X1) node[pos=0.5,sloped,above] {};
		\draw[-arcsq] (L6) -- (X2) node[pos=0.5,sloped,above] {};
		\draw[-arcsq] (L6) -- (X3) node[pos=0.5,sloped,above] {};
		\draw[-arcsq] (L6) -- (X4) node[pos=0.5,sloped,above] {};
		\draw[-arcsq] (L7) -- (X5) node[pos=0.5,sloped,above] {};
		\draw[-arcsq] (L7) -- (X6) node[pos=0.5,sloped,above] {};
		\draw[-arcsq] (L8) -- (X7) node[pos=0.5,sloped,above] {};
		\draw[-arcsq] (L8) -- (X8) node[pos=0.5,sloped,above] {};
		\draw[-arcsq] (L9) -- (X9) node[pos=0.5,sloped,above] {};
		\draw[-arcsq] (L9) -- (X10) node[pos=0.5,sloped,above] {};
		\draw[-arcsq] (L9) -- (X11) node[pos=0.5,sloped,above] {};
		\end{tikzpicture}
        \vspace{-3mm}
		\caption{A hierarchical causal structure involving $9$ latent variables (shaded nodes) and 13 observed variables (unshaded nodes).}
		\vspace{-0.7cm}
		\label{fig:simple-main-example} 
	\end{center}
\vspace{-4mm}
\end{figure}
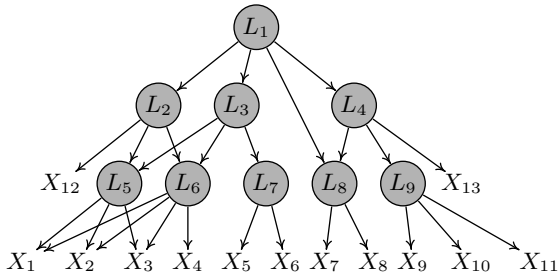

To incorporate higher-order information, one approach is to apply over-complete independent component analysis \citep{hoyer2008estimation, shimizu2009estimation}. However, this method does not consider the causal structure among latent variables and the size of the equivalence class of the identified structure may be large \citep{entner2010discovering, tashiro2014parcelingam}. Another interesting work by \cite{anandkumar2013learning} extracts second-order statistics to identify latent factors, while using non-Gaussianity when estimating causal relations among latent variables. More recently, \cite{cai2019triad} proposed the LSTC algorithm to discover the structure among latent variables with non-Gaussian distributions by making use of the proposed Triad condition. However, the above methods assume that each set of latent variables has a much larger number of observed variables as children and cannot handle the situation with a latent hierarchical structure (i.e., the children of latent variables may still be latent). For instance, consider a latent hierarchical causal structure illustrated in Figure  \ref{fig:simple-main-example}, where the variables $L_i$ ($i=1,...,9$) are unobserved and $X_j$ ($j=1,...,13$) are observed. These methods would fail to discover the latent variables $L_1, L_2, L_3$, and $L_4$.

Several contributions have been made to learning the latent hierarchical structure other than the traditional measurement model.\footnote{Each latent variable has specific measured variables as children in the measurement model.} For instance, \cite{zhang2004hierarchical} generalized the classic latent cluster models and proposed hierarchical latent class models (also known as latent tree models) for discrete variables. \cite{poon2010variable} extended this model and introduced Pouch Latent Tree Models, which allow each leaf node to consist of one or more continuous observed variables. \cite{choi2011learning} further investigated more general latent tree models applicable to both discrete and Gaussian random variables and provided efficient estimation algorithms.
Other interesting developments along this line include~\citet{harmeling2010greedy,mourad2013survey,zhang2017latent,etesami2016learning,drton2017marginal,zhou2020learning}. Although these methods have been used in various fields, they typically assume a tree-structured graph, i.e., there is only one path between any pair of variables in the system. However, in many settings, e.g., with the structure in Figure \ref{fig:simple-main-example}, this assumption is violated.

\revision{Recently, \cite{adams2021identification} established necessary and sufficient conditions for structure identifiability in linear non-Gaussian setting, which is exciting. However, their work does not provide a practical estimation procedure, which we aim to achieve in this work. Besides, \Revitwo{their work requires the number of latent factors a priori} while our work can determine it automatically.}

It is well known that the \emph{Independent Noise (IN)} condition can be used to recover the causal structure from a linear non-Gaussian causal model without latent variables~\citep{shimizu2011directlingam}. This leads to a natural question: is it possible to solve the latent-variable problem, by incorporating higher-order information (i.e., non-Gaussianity) and developing a condition similar to the independent noise condition? With this motivation in mind, the goal of this work is to develop a general independent noise condition and establish the corresponding theorems for recovering the causal structure from a linear non-Gaussian causal model with latent variables.
\revision{
Interestingly, we find that a more general independent noise condition can be achieved by testing the independence between $\omega^{\intercal} \mathbf{Y}$ and $\mathbf{Z}$, where $\mathbf{Y}$ and $\mathbf{Z}$ are two observed random vectors, and $\omega$ is a parameter vector determined by the cross-covariance between $\mathbf{Y}$ and $\mathbf{Z}$; we term it  \textit{Generalized Independent Noise (GIN) condition}.
We say $(\mathbf{Z}, \mathbf{Y})$ follows the GIN condition, if and only if $\omega^{\intercal} \mathbf{Y}$ and $\mathbf{Z}$ are statistically independent. By leveraging GIN, we further develop a practical algorithm to estimate Linear, Non-Gaussian Latent Hierarchical Models (LiNGLaHs), including where the latent variables are, the number of latent variables behind any two observed variables, and the causal order of the latent variables.
Specifically, we make the following contributions:
\begin{itemize}[leftmargin=30pt,align=left,itemsep=1pt,topsep=0pt]
    \item [1.] \Revitwo{We define the GIN condition for an ordered pair of variable sets and show that the well-known IN condition} presented in \cite{shimizu2011directlingam} can be seen as a special case of GIN.\footnote{Preliminary results of GIN condition were presented at \cite{xie2020GIN}.} 
    \item [2.] We then further provide necessary and sufficient graphical criteria of the GIN condition in linear non-Gaussian acyclic causal models. Compared to the original graphical criteria presented in~\cite{xie2020GIN}, \Revitwo{this is a more general graphical criterion and is not restricted by the \textit{Purity assumption}} (i.e., direct edges between observed variables are permitted), and the \textit{Double-Pure Child Variable assumption} (i.e., latent variable sets can have any number of measurement variables as children).
    \item [3.] We exploit GIN to estimate the LiNGLaHs, in which latent confounders may also be causally related and some latent variables may not have observed variables as children (i.e., beyond a measurement model). 
    We demonstrate that, under mild assumptions, the whole structure of a LiNGLaH is identifiable by leveraging GIN conditions under mild assumptions. 
    Compared to the algorithm presented in \citet{xie2022identification}, our proposed algorithm is capable of identifying more complex latent graphs. This suggests that the method outlined in \citet{xie2022identification} can be viewed as a particular case of our proposed algorithm, specifically when the size of the set of latent variables is limited to one.
    \item [4.] We address several practical challenges associated with the proposed method and offer a more reliable and statistically efficient approach for estimating the structure of LiNGLaHs with limited samples in practical scenarios.
    \item [5.] We demonstrate the efficacy of our algorithm on both synthetic and real-world datasets across three different domains.
\end{itemize}
}

The rest of this paper is organized as follows. In Section \ref{Sec-notation-problem-def}, we introduce notations, terminologies, and definitions used in this paper. In Section \ref{Sec-GIN}, we define the GIN condition for an ordered pair of variable sets, provide mathematical conditions that are sufficient for it, and show that the independent noise condition can be seen as its special case. We then further give necessary and sufficient graphical conditions under which the GIN condition holds. In Section \ref{Sec-Application}, we exploit GIN, together with an efficient search algorithm, to estimate the LiNGLaH, which allows causal relationships between latent variables, multiple latent confounders behind any two variables, and certain causal edges among measured variables. We show that the structure of a LiNGLaH is (mostly) identifiable in terms of the GIN condition under mild assumptions.
We additionally provide comprehensive implementation details of our proposed algorithm to ensure a more robust estimation of LiNGLaH in practical scenarios. 
We demonstrate the efficacy of our algorithm on both synthetic and real-world datasets in Section \ref{Sec-Experiment}$\&$\ref{Section-Real-word-Study} and conclude in Section \ref{Sec-Conclusion}.

\section{Background} \label{Sec-notation-problem-def}

\subsection{Notation} \label{Subsec-notation}
The main symbols used in this paper are summarized in Table \ref{Table-symbols}, and commonly-used concepts in graphical models, such as path and d-separation, can be found in Appendix A or standard sources such as ~\citet{pearl2009causality,spirtes2000causation}. In this paper, we use ``variable" and ``vertex/node" interchangeably.

\subsection{Linear Non-Gaussian Acyclic Causal Model} \label{Subsec-problem-description}
In this paper, we consider linear non-Gaussian acyclic causal models of a set of variables $\mathbf{V}=\{V_1,\cdots,V_n\}$. Without loss of generality, we assume that each variable in $\mathbf{V}$ has a zero mean, and the causal process can be represented by the following linear structural equation model (SEM):
\begin{nospaceflalign}\label{Eq-linear-model}
{V_i} & = \sum_{{{V_j} \in \mathbf{Pa}({V_i})}}{{b_{ij}}{V_j} + {\varepsilon_{V_i}}},
\end{nospaceflalign} 
where $b_{ij}$ represents the causal strength from $V_j$ to $V_i$.
Each noise variable $\varepsilon_{V_i}$ is a continuous random variable with a non-Gaussian distribution and non-zero variance, and $\varepsilon_{V_i}$, for $i=1,\cdots,n$, are independent of each other. We assume that the generating process is recursive~\citep{bollen1989structural}; that is to say, the causal relationships among variables $\mathbf{V}$ can be represented by a DAG~\citep{pearl2009causality,spirtes2000causation}.
This model is known as the Linear, Non-Gaussian, Acyclic Model (LiNGAM) when all variables in $\mathbf{V}$ are observed~\citep{shimizu2006linear}.
It is important to note that, in this paper, we allow some variables in $\mathbf{V}$ to be not observed, where $\mathbf{X}$ denotes the set of observed variables and $\mathbf{L}$ denotes the set of latent variables.

\begin{table*}[t]
    \small
    \center \caption{The list of main symbols used in this paper}
    \label{Table-symbols}
    \begin{tabular}{p{2.4cm}p{11.6cm}}
    \toprule
    \textbf{Symbol} & \textbf{Description}\\ 
    \midrule
    $\mathcal{G}=(\mathbf{V},\mathbf{E})$       & A directed acyclic graph (DAG) with a set of nodes (or vertex) $\mathbf{V}$ and a set of directed edges $\mathbf{E}$  \\
    $\mathbf{V}=\mathbf{X}\cup\mathbf{L}$      & The set of all variables, where ${\bf{X}}$ denotes the set of observed variables and ${\bf{L}}$ denotes the set of latent variables. \\
    $\mathbf{V}_{1:k}$ & $\{{V}_{1}, \ldots,{V}_{k}\}$\\
    $\textrm{Adj}_{\mathbf{Y},\mathbf{Z}}$ & The sub-adjacency matrix of $\textrm{Adj}$ with row indices $\mathbf{Y}$ and column indices $\mathbf{Z}$, where a non-zero element $\textrm{Adj}_{V_i,V_j}$ indicates a directed edge from $V_i$ to $V_j$\\
    $\mathbf{Pa}(V_i)$, $\mathbf{Ch}(V_i)$  & The set of all parents and all children of $V_i$, respectively \\
    $\mathbf{An}(V_i)$, $\mathbf{De}(V_i)$  & The set of all ancestors and all descendants of $V_i$, respectively \\
    $\mathbf{De}_{\mathbf{O}}(V_i)$       & The set of all observed descendants of $V_i$ \\
    $\mathbf{Ne}(V_i)$   & The set of all neighbors of $V_i$ \\
    $|*|$ (e.g., $|\mathbf{Y}|$)       & The cardinality of a set \\
    $\mathcal{L}_i$ & A set of latent variables \\
    $L(\mathbf{C}_1)$  & The set that contains all the common latent parents of any two nodes in $\mathbf{C}_1$, excluding the variables in $\mathbf{C}_1$ \\
    $\textrm{APa}(\mathbf{Y})$ & The set of variables that are parents of any component of $\mathbf{Y}$ \\
    $\boldsymbol{\Sigma}_{\mathbf{A},\mathbf{B}}$  & The cross-covariance matrix of set $\mathbf{A}$ and $\mathbf{B}$ \\
    $\mathrm{rank}(\boldsymbol{\Sigma}_{\mathbf{A},\mathbf{B}})$ & The rank of cross-covariance matrix of set $\mathbf{A}$ and $\mathbf{B}$ \\
    $\CI$ & Symbol of independence, e.g., $Y\CI Z$ means ``$Y$ is statistically independent of $Z$"\\
    $\nCI$ & Symbol of dependence, e.g., $Y \nCI Z$ means ``$Y$ is not statistically independent of $Z$"\\
    $|$ & Given or conditional on, e.g., $Y | Z$ means ``$Y$ conditional on (or given) Z " \\
    $\neg_{\mathbf{C}_1}(*)$ & The maximal set of $\mathbf{C}_1$ such that for any variables $V_i \in *$ and $V_j \in \neg_{\mathbf{C}_1}(*)$,  $\{V_i,V_j\}$ is a pure set \\
    \bottomrule
    \end{tabular}
    \vspace{-3mm}
\end{table*}  

\section{GIN Condition and Its Graphical Representation}\label{Sec-GIN}

In this section, we first briefly review the Independent Noise (IN) condition in linear non-Gaussian acyclic causal models without latent variables in Section \ref{Subsec-IN-condition}. Then in Section \ref{Subsec-GIN-condition}, we formulate the Generalized Independent Noise (GIN) condition and show that it contains the Independent Noise (IN) condition as a special case. We further give mathematical characterizations of the GIN condition. In Section \ref{Subsec-graph-criteria}, we provide the implication of GIN in a linear non-Gaussian acyclic causal model and present its graphical representation, which connects the GIN condition and causal graphs. All proofs are provided in Appendix B.

\subsection{Independent Noise Condition}\label{Subsec-IN-condition}

Below, we formulate the independent noise condition, which has been used in causal discovery of linear, non-Gaussian acyclic causal models without latent confounders \citep{shimizu2011directlingam}.
\begin{Definition}[IN Condition]
Suppose all variables follow the linear non-Gaussian acyclic causal model where all variables are observed. Let $Y$ be a single variable and $\mathbf{Z}$ be a set of variables. Denote by $\tilde{E}_{Y||\mathbf{Z}}$ the residual of regressing $Y$ on $\mathbf{Z}$, that is,
\begin{nospaceflalign}\label{Eq-IN-E_XZ}
\tilde{E}_{Y||\mathbf{Z}} = Y - \tilde{\omega}^{\intercal}\mathbf{Z},
\end{nospaceflalign}
where $\tilde{\omega}\coloneqq \mathbb{E}[Y \mathbf{Z}^\intercal]\mathbb{E}^{-1}[\mathbf{Z} \mathbf{Z}^\intercal]$. We say that $(\mathbf{Z},Y)$ follows the IN condition if and only if  $\tilde{E}_{Y||\mathbf{Z}} \CI \mathbf{Z}$.
\end{Definition}

Lemma 1 in \cite{shimizu2011directlingam} considers the case where $\mathbf{Z}$ is a single variable and shows that $(\mathbf{Z}, Y)$ satisfies the IN condition if and only if $\mathbf{Z}$ is an exogenous variable relative to $Y$, based on which one can identify the causal relationship between $Y$ and $\mathbf{Z}$. As a direct extension of this result, we show that in the case where $\mathbf{Z}$ contains multiple variables, $(\mathbf{Z}, Y)$ satisfies the IN condition if and only if all variables in $\mathbf{Z}$ are causally earlier (according to causal ordering) than $Y$ and there is no common cause behind any variable in $\mathbf{Z}$ and $Y$. This result is given in the following proposition.
\begin{Proposition}[Graphical Criterion of IN Condition]\label{pro-extension-IN}
Suppose all considered variables follow a linear non-Gaussian acyclic causal model and all variables are observed. Let $\mathbf{Z}$ be a subset of those variables and $Y$ be a single variable. Then the following two statements are equivalent.
\begin{itemize}[leftmargin=30pt,align=left,itemsep=1pt,topsep=0pt]
    \item[(A)] 1) All variables in $\mathbf{Z}$ are causally earlier than $Y$,  and 2) there is no common cause for each variable in $\mathbf{Z}$ and $Y$ that is not in $\mathbf{Z}$.
    \item[(B)] $(\mathbf{Z},Y)$ satisfies the IN condition.
\end{itemize}
\end{Proposition}

\subsection{Generalized Independent Noise Condition}\label{Subsec-GIN-condition}
Below, we first define the GIN condition, followed by an illustrative example.
\begin{Definition}[GIN Condition]\label{Definition-GIN}
Suppose all variables follow a linear non-Gaussian acyclic causal model. Let $\mathbf{Y}$, $\mathbf{Z}$ be two sets of random variables.  Define the surrogate-variable of $\mathbf{Y}$ relative to $\mathbf{Z}$ as
\begin{nospaceflalign}\label{eq-E_XZ}
E_{\mathbf{Y}||\mathbf{Z}} \coloneqq \omega^\intercal \mathbf{Y},
\end{nospaceflalign}
where $\omega$ satisfies $\omega^\intercal \mathbb{E}[\mathbf{Y}\mathbf{Z}^\intercal] = \mathbf{0}$ and $\omega \neq \mathbf{0}$.  
We say that $(\mathbf{Z},\mathbf{Y})$ follows the GIN condition if and only if $E_{\mathbf{Y}||\mathbf{Z}} \CI \mathbf{Z}$.
\end{Definition}

\begin{figure}[htp]
	\begin{center}
		\begin{tikzpicture}[scale=1.5, line width=0.5pt, inner sep=0.2mm, shorten >=.1pt, shorten <=.1pt]
		\draw (1.5, 2.5) node(i-L1) [circle, fill=gray!60,draw] {{\footnotesize\,$L_1$\,}};
		\draw (1.5, 0.5) node(i-L2) [circle, fill=gray!60,draw] {{\footnotesize\,$L_2$\,}};
		\draw (3.5, 2.3) node(i-L3) [circle, fill=gray!60,draw] {{\footnotesize\,$L_3$\,}};
		\draw (3.5,0.7) node(i-L4) [circle, fill=gray!60,draw] {{\footnotesize\,$L_4$\,}};
		\draw (0.5, 1.5) node(i-X1) [] {{\footnotesize\,$X_1$\,}};
		\draw (1, 1.5) node(i-X2) [] {{\footnotesize\,$X_2$\,}};
		\draw (2, 1.5) node(i-X3) [] {{\footnotesize\,{$X_3$}\,}};
		\draw (2.5, 1.5) node(i-X4) [] {{\footnotesize\,{$X_4$}\,}};
		\draw (4.5, 2.8) node(i-Y1) [] {{\footnotesize\,{$X_5$}\,}};
		\draw (4.5, 1.8) node(i-Y3) [] {{\footnotesize\,{$X_6$}\,}};
		\draw (4.5, 1.2) node(i-Z1) [] {{\footnotesize\,{$X_7$}\,}};
		\draw (4.5, 0.2) node(i-Z3) [] {{\footnotesize\,{$X_{8}$}\,}};
		%
		\draw[-arcsq] (i-L1) -- (i-X1) node[pos=0.5,sloped,above] {\scriptsize{$a_{1}$}}; 
		\draw[-arcsq] (i-L1) -- (i-X2)node[pos=0.5,sloped,below] {\scriptsize{$a_{2}$}}; 
		\draw[-arcsq] (i-L1) -- (i-X3) node[pos=0.5,sloped,below] {\scriptsize{$a_{3}$}};
		\draw[-arcsq] (i-L1) -- (i-X4) node[pos=0.5,sloped,above] {\scriptsize{$a_{4}$}};
		\draw[-arcsq] (i-L2) -- (i-X1) node[pos=0.5,sloped,below] {\scriptsize{$b_{1}$}}; 
		\draw[-arcsq] (i-L2) -- (i-X2)node[pos=0.5,sloped,above] {\scriptsize{$b_{2}$}}; 
		\draw[-arcsq] (i-L2) -- (i-X3) node[pos=0.5,sloped,above] {\scriptsize{$b_{3}$}};
		\draw[-arcsq] (i-L2) -- (i-X4) node[pos=0.5,sloped,below] {\scriptsize{$b_{4}$}};
		\draw[-arcsq] (i-L3) -- (i-Y1) node[pos=0.5,sloped,above] {\scriptsize{$c_{1}$}}; 
		\draw[-arcsq] (i-L3) -- (i-Y3)node[pos=0.5,sloped,below] {\scriptsize{$c_{2}$}}; 
		\draw[-arcsq] (i-L4) -- (i-Z1) node[pos=0.5,sloped,above] {\scriptsize{$d_{1}$}}; 
		\draw[-arcsq] (i-L4) -- (i-Z3)node[pos=0.5,sloped,below] {\scriptsize{$d_{2}$}}; 
		\draw[-arcsq] (i-L1) -- (i-L2) node[pos=0.5,sloped,above] {\scriptsize{$\alpha$}};
		\draw [-arcsq] (i-L1) edge[bend right=-25] (i-L4);
		\draw (2.5,2.2) node(label-iii) [] {{\scriptsize\,$\gamma$\,}};
		\draw [-arcsq] (i-L1) edge[bend right=-25] (i-L3);
		\draw (2.5,2.8) node(label-iii) [] {{\scriptsize\,$\beta$\,}};
		\draw [-arcsq] (i-L2) edge[bend right=25] (i-L3);
		\draw (2.5,0.8) node(label-iii) [] {{\scriptsize\,$\sigma$\,}};
		\draw [-arcsq] (i-L2) edge[bend right=25] (i-L4);
		\draw (2.5,0.4) node(label-iii) [] {{\scriptsize\,$\eta$\,}};
		\draw[-arcsq] (i-L3) -- (i-L4) node[pos=0.5,sloped,above] {\scriptsize{$\theta$}};
		\end{tikzpicture}
		\caption{A causal structure involving $4$ latent variables and 8 observed variables, where each pair of observed variables in $\{X_1,X_2,X_3,X_4\}$ are affected by two latent variables.}
		\vspace{-0.4cm}
		\label{fig:simple-example-explain-GIN-constraint} 
	\end{center}
\end{figure}
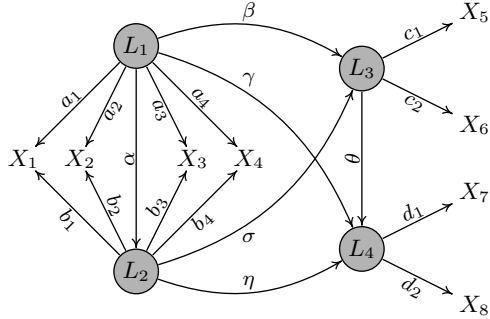

In other words, $(\mathbf{Z},\mathbf{Y})$ violates the GIN condition if and only if $E_{\mathbf{Y}||\mathbf{Z}}\nCI\mathbf{Z}$. Notice that the Triad condition~\citep{cai2019triad} can be seen as a restrictive, special case of the GIN condition, where $|\mathbf{Y}|=2$ and $|\mathbf{Z}|=1$.
We give an example to illustrate the connection between this condition and the causal structure. Considering the causal structure depicted in Figure \ref{fig:simple-example-explain-GIN-constraint} and assuming faithfulness,\revision{\footnote{\revision{The faithfulness condition states that if $\mathbf{A}$ and $\mathbf{B}$ are probabilistically independent conditional on $\mathbf{C}$ according to the true probability measure, then $\mathbf{A}$ and $\mathbf{B}$ are d-separated by $\mathbf{C}$ in true $\mathcal{G}$.}}} it can be deduced that $(\{X_4, X_5\},\{X_1, X_2, X_3\})$ satisfies the GIN condition. The explanation for this is provided below.
The causal models of latent variables are $L_1 =\varepsilon_{L_1}$, $L_2 =\alpha L_1+\varepsilon_{L_2}=\alpha \varepsilon_{L_1}+ \varepsilon_{L_2}$, and $L_3 =\beta L_1+\sigma L_2+ \varepsilon_{L_3}=(\beta+\alpha\sigma) \varepsilon_{L_1}+ \sigma\varepsilon_{L_2}+\varepsilon_{L_3}$, so 
$\{X_1,X_2,X_3\}$ and $\{X_4,X_5\}$ can then be represented as
\vspace{2mm}
\begin{nospaceflalign}
\underbrace{\left[\begin{matrix}
{{X_1}}\\
{{X_2}}\\
{{X_3}}
\end{matrix}\right]}_{\mathbf{Y}} & =  {\left[\begin{matrix}
a_1 & {b_1}\\
a_2 & {b_2}\\
a_3 & {b_3}
\end{matrix}\right]}\left[\begin{matrix}
{{{{L_1}}}}\\
{{{{L_2}}}}
\end{matrix}\right] + \underbrace{\left[\begin{matrix}
{{\varepsilon_{{X_1}}}}\\
{{\varepsilon_{{X_2}}}}\\
{{\varepsilon_{{X_3}}}}
\end{matrix}\right]}_{\mathbf{E_Y}}, \label{equ-y1-y2}\\
\underbrace{\left[\begin{matrix}
{{X_4}}\\
{{X_5}}
\end{matrix}\right]}_{\mathbf{Z}} & =  {\left[\begin{matrix}
a_4 &  b_4\\
\beta c_1 &  \sigma c_1
\end{matrix}\right]}\left[\begin{matrix}
{{L_{1}}}\\
{{L_{2}}}
\end{matrix}\right] + \underbrace{\left[\begin{matrix}
{{\varepsilon_{{X_4}}}}\\
{{\varepsilon_{{X^{'}_5}}}}
\end{matrix}\right]}_{\mathbf{E_Z}}, \label{equ-z1-z2}
\end{nospaceflalign}
where $\varepsilon_{X^{'}_5}=c_{2}\varepsilon_{L_3}+\varepsilon_{X_5}$.
According to the above equations,  $\omega^\intercal \mathbb{E}[\mathbf{Y}\mathbf{Z}^\intercal] = \mathbf{0} \Rightarrow \omega = [a_2b_3-b_2a_3,b_1a_3-a_1b_3, a_1b_2-b_1a_2]^{\intercal}$. Then we can see  $E_{\mathbf{Y}||\mathbf{Z}} = \omega^{\intercal}\mathbf{Y}= \omega^{\intercal}\mathbf{E_Y}$, and further because $\mathbf{E_Y} \CI \mathbf{Z}$, we have $E_{\mathbf{Y}||\mathbf{Z}} \CI \mathbf{Z}$. That is to say, $(\{X_4,X_5\},\{X_1,X_2,X_3\})$ satisfies the GIN condition. Intuitively, we have  $E_{\mathbf{Y}||\mathbf{Z}} \CI \mathbf{Z}$ because although $\{X_1,X_2,X_3\}$ were generated by $\{L_1,L_2\}$, which are not measurable, $E_{\mathbf{Y}||\mathbf{Z}}$, as a particular linear combination of $\mathbf{Y} = \{X_1,X_2,X_3\}$, successfully removes the influences of $\{L_1,L_2\}$ by properly making use of $\mathbf{Z} = \{X_4,X_5\}$ as a ``surrogate". 

Next, we discuss a situation where GIN is violated. Continue to consider the structure in Figure \ref{fig:simple-example-explain-GIN-constraint}, where $(\{X_3,X_6\},\{X_1,X_2,X_5\})$ violates GIN. Specifically, the corresponding variables satisfy the following equations:
\vspace{2mm}
\begin{nospaceflalign}
\underbrace{\left[\begin{matrix}
{{X_1}}\\
{{X_2}}\\
{{X_5}}
\end{matrix}\right]}_{\mathbf{Y}} & = \left[\begin{matrix}
{a_1} & {b_1} \\
{a_2} & {b_2} \\
\beta c_1 & \sigma c_2 
\end{matrix}\right]\left[\begin{matrix}
{{L_{1}}}\\
{{L_{2}}}
\end{matrix}\right] + \underbrace{\left[\begin{matrix}
{{\varepsilon _{{X_1}}}}\\
{{\varepsilon _{{X_2}}}}\\
{{\varepsilon _{{X^{'}_5}}}}
\end{matrix}\right]}_{\mathbf{E_Y}},\\
\underbrace{\left[\begin{matrix}
{{X_3}}\\
{{X_6}}
\end{matrix}\right]}_{\mathbf{Z}} & = \left[\begin{matrix}
{a_3} & {b_3}\\
\beta c_2 &  \sigma c_2
\end{matrix}\right]\left[\begin{matrix}
{{L_{1}}}\\
{{L_{2}}}
\end{matrix}\right] + \underbrace{\left[\begin{matrix}
{{\varepsilon _{{X_3}}}}\\
{\varepsilon _{X^{'}_6}}
\end{matrix}\right]}_{\mathbf{E_Z}},
\end{nospaceflalign}
where $\varepsilon_{X^{'}_6}=c_{2}\varepsilon_{L_3}+\varepsilon_{X_6}$.
Then under faithfulness assumption, we can see $\omega^{T}\mathbf{Y} \nCI \mathbf{Z}$ because $\mathbf{E_Y} \nCI \mathbf{E_Z}$ (there exists common component $\varepsilon_{L_3}$ for $\varepsilon_{X^{'}_5}$ and $\varepsilon_{X^{'}_6}$), no matter $\omega^\intercal \mathbb{E}[\mathbf{Y}\mathbf{Z}^\intercal] = 0$ or not. 

Consider the structure in Figure \ref{fig:simple-example-explain-GIN-constraint}. We have $(\{X_4,X_5\},\{X_1,X_2,X_3\})$ satisfies the GIN condition, and  $\{L_1,L_2\}$, the latent common causes 
for $\{X_1,X_2,X_3\}$, d-separate $\{X_1,X_2,X_3\}$ from $\{X_4,X_5\}$. In contrast, $(\{X_3,X_6\},\{X_1,X_2,X_5\})$ violates GIN, and $\{X_1,X_2,X_5\}$ and $\{X_3,X_6\}$ are {\it not} d-separated by $\{L_1,L_2\}$, the latent common causes of $\{X_1,X_2,X_5\}$.

We next give mathematical characterizations of the GIN condition in the following theorem, by providing sufficient conditions for when $(\mathbf{Z},\mathbf{Y})$ satisfies the GIN condition.

\begin{Theorem}[Mathematical Characterization of GIN]\label{Theorem-mathematical-GIN}
Suppose that random vectors $\mathcal{S}$, $\mathbf{Y}$, and $\mathbf{Z}$ are related in the following way:
\begin{nospaceflalign}
\mathbf{Y} &= A \mathcal{S} + \mathbf{E}_Y,  \label{eq-X} \\ 
\mathbf{Z} & = B \mathcal{S} + \mathbf{E}_Z. \label{eq-Z}
\end{nospaceflalign}
Denote by $s$ the dimensionality of $\mathcal{S}$.  
Assume $A$ is of full column rank.
Then, if 1) $\textrm{Dim}(\mathbf{Y}) > s$, 2) $\mathbf{E}_Y \CI \mathcal{S}$, 3)  $\mathbf{E}_Y \CI \mathbf{E}_Z$,\footnote{Note that we do not assume $\mathbf{E}_Z \CI \mathcal{S}$.} and 4) the cross-covariance matrix of $\mathcal{S}$ and $\mathbf{Z}$, $\boldsymbol{\Sigma}_{\mathcal{S},\mathbf{Z}} = \mathbb{E}[\mathcal{S}\mathbf{Z}^\intercal]$ has rank $s$, then $E_{\mathbf{Y}||\mathbf{Z}} \CI \mathbf{Z}$, i.e., $(\mathbf{Z},\mathbf{Y})$ satisfies the GIN condition.
\end{Theorem}

\begin{Example-set}
Continue the example in Figure~\ref{fig:simple-example-explain-GIN-constraint}. Let $\mathbf{Z}=\{X_{4},X_{5}\}$ and $\mathbf{Y}=\{X_1,X_2,X_3\}$, and thus $\mathcal{S}=\{L_1,L_2\}$. We can observe the following facts: $\textrm{Dim}(\mathbf{Y})=2 > s$, $\mathbf{E}_Y \CI \mathcal{S}$ and $\mathbf{E}_Y \CI \mathbf{E}_Z$ according to Equations \ref{equ-y1-y2} and \ref{equ-z1-z2}, and $\boldsymbol{\Sigma}_{L,Z} = \mathbb{E}[\mathcal{S}\mathbf{Z}^\intercal]$ has full row rank, i.e., $2$. Therefore, $(\mathbf{Z},\mathbf{Y})$ satisfies the GIN condition.
\end{Example-set}

The following proposition shows that the IN condition can be seen as a special case of the GIN condition with $\mathbf{E}_Z = 0$ (i.e., $\mathbf{Z}$ and $\mathcal{S}$ are linearly and deterministically related).

\begin{Proposition}[Connection between IN and GIN]\label{Propo-IN-Special_Case}
Let $\ddot{{Y}} \coloneqq ({Y}, \mathbf{Z})$.  
Then the following statements hold:
\begin{itemize}[itemsep=0.2pt,topsep=0.2pt]
    \item[1.] If $({\mathbf{Z}, \ddot{{Y}}})$ follows the GIN condition, then $(\mathbf{Z}, {Y})$ follows IN condition.
    \item[2.] If $(\mathbf{Z}, {Y})$ follows the IN condition, then $(\mathbf{Z}, \ddot{{Y}})$ follows the GIN condition.
\end{itemize}
\end{Proposition}

Proposition \ref{Propo-IN-Special_Case} inspires a unified framework to handle causal relations with or without latent variables.

\subsection{Graphical Criteria of GIN Condition}
\label{Subsec-graph-criteria}
In this section, we investigate the graphical implication of the GIN condition in linear non-Gaussian acyclic causal models. In particular, we provide necessary and sufficient graphical conditions under which the GIN condition holds. These conditions allow GIN to be employed as a testable implication of linear non-Gaussian causal models. 

We next give the rank-faithfulness assumption, which will allow us to use rank constraints to find out (unobservable) structural constraints among latent variables.

\begin{Definition}[Rank Faithfulness~\citep{spirtes2013calculation-t-separation}]\label{Ass_rankfaithful}
Let a distribution $P$ be (linearly) rank-faithful to a directed acyclic graph $\mathcal{G}$ if every rank-constraint on a sub-covariance matrix that holds in $P$ is entailed by every free-parameter linear structural model with path diagram equal to $\mathcal{G}$.
\end{Definition}
\revision{
Note that the widely-used (linear) faithfulness, which assumes that conditional independence statements hold in the distribution if and only if a corresponding d-separation holds in $\mathcal{G}$ \citep{spirtes2000causation}, is included in rank faithfulness \citep{silva2017learning}.
This is due to the following reasons. First, the conditional independence statement $\mathbf{A} \CI \mathbf{B} | \mathbf{C}$ holds in the graph $\mathcal{G}$ if and only if $\mathrm{rank}(\boldsymbol{\Sigma}_{\mathbf{A}\cup \mathbf{C}, \mathbf{B}\cup \mathbf{C}}) $ equals $|\mathbf{C}|$.
Moreover, additional rank-deficiency constraints may imply further structural constraints beyond those inferred by conditional independence statements, as discussed in Example 2.5 in \cite{Sullivant-T-separation}.
The practicality of rank-faithfulness has been demonstrated through simulation results and applications in \citet{Kummerfeld2016,huang2022latent}, as well as in our paper.
For a further discussion of rank-faithfulness assumptions, please refer to Section 4 in \cite{spirtes2013calculation-t-separation} or Section 3.1.1 in \cite{silva2017learning}.
} 

We next show the connection between GIN and graphical properties of variables in terms of the linear non-Gaussian acyclic causal model in the following theorem, which then inspires us to exploit the GIN condition to discover the graph containing latent variables. 
\revision{
Before that, we first introduce a key concept, termed  \textit{side choke-point set}, which builds upon the notion of choke points introduced in \cite{shaferageneralization1993}.

\begin{Definition}[Side Choke-Point Set]\label{Def-Choke-point}
    Let $\mathbf{S}$, $\mathbf{Y}$, and $\mathbf{Z}$ be three vertices subsets of $\mathbf{V}$ which need not be disjoint. We say that a set $\mathcal{S}$ is a $\mathbf{Y}$-side choke-point set between $\mathbf{Y}$ and $\mathbf{Z}$ if, for each active path $\pi$ between $\mathbf{Y}$ and $\mathbf{Z}$, the following three conditions are met:
	\begin{description}[itemsep=0.2pt,topsep=0.2pt]
		\item [(1)] path $\pi$ contains a node $S$ in $\mathcal{S}$,\footnote{An active path is defined as a path that does not contain any colliders.} and
		\item [(2)] $S$ is on the $\mathbf{Y}$ side of path $\pi$, that is, there is a directed subpath from $S$ to $\mathbf{Y}$ on $\pi$.
	\end{description}
\end{Definition}

\begin{Theorem}[Graphical Criteria of GIN]\label{Theorem:GIN graphical}
Let  $\mathbf{Y}$ and $\mathbf{Z}$ be two sets of observed variables of a linear non-Gaussian acyclic causal model which may not be disjoint. Assume the rank faithfulness holds. 
$(\mathbf{Z},\mathbf{Y})$ satisfies the GIN condition 
if and only if there exists a set $\mathcal{S}$ with $0\leq |\mathcal{S}| \leq \textrm{min}(|\mathbf{Y}|-1, |\mathbf{Z}|)$,
such that 1) for any variable $V$ in $\mathbf{Y}$ but not in $\mathcal{S}$, $V$ is not an ancestor of any variable in $\mathcal{S}$, 
that 2) $\mathcal{S}$ is a $\mathbf{Y}$-side choke-point set,
and that 3) the covariance matrix of $\mathcal{S}$ and $\mathbf{Z}$ has rank $|\mathcal{S}|$, and so does that of $\mathcal{S}$ and $\mathbf{Y}$.
\end{Theorem}
Roughly speaking, the conditions in this theorem can be interpreted as follows: i.) every active (collider-free) path between $\mathbf{Y}$ and $\mathbf{Z}$ must contain a node from $\mathcal{S}$, which is always causally earlier (according to the causal order) of the $\mathbf{Y}$-side,
and ii.) the linear transformation from $\mathcal{S}$ to $\mathbf{Z}$ has full column rank, and so does that of $\mathcal{S}$ and $\mathbf{Y}$.
This version offers a broader graphical criterion over the one in \cite{xie2020GIN}, removing the constraints of the Purity assumption (i.e., direct edges between observed variables are permitted) and the Double-Pure Child Variable assumption (i.e., latent variable sets can have any number of measurement variables as children).
}
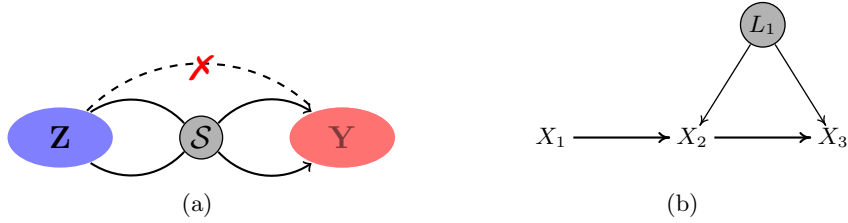
\begin{figure}[htp]
	\begin{center}
		\begin{tikzpicture}[scale=1.5, line width=0.5pt, inner sep=0.2mm, shorten >=.1pt, shorten <=.1pt]
			\draw (1.25, 0.7) node(S) [circle, fill=gray!60,draw] {{\large\,$\mathcal{S}$\,}};
			\draw (0.0, 0.7) node(Z) [minimum width=1.4cm, minimum height=0.8cm, shape=ellipse] {{\large\,$\mathbf{Z}$\,}};
            \begin{scope}[on background layer]
            \node [fill=blue!50, thick, minimum width=1.4cm, minimum height=0.8cm, shape=ellipse] at (Z) {};
            \end{scope}
			\draw (2.5, 0.7) node(Y) [fill=red!50,thick, fill opacity=0.5,minimum width=1.4cm, minimum height=0.8cm, shape=ellipse] {{\large\,$\mathbf{Y}$\,}};
            \begin{scope}[on background layer]
            \node [fill=red!60, thick, minimum width=1.4cm, minimum height=0.8cm, shape=ellipse] at (Y) {};
            \end{scope}
			\draw [->, thick] (S) edge[bend right=40] (Y);
			\draw [->, thick] (S) edge[bend right=-40] (Y);
			\draw [-, thick] (Z) edge[bend right=40] (S);
			\draw [-, thick] (Z) edge[bend right=-40] (S);
            \draw [-, thick, dashed] (Z) edge[bend right=-45] (Y);
            %
            \draw (1.25, 1.32) node(cross) [] {{\small\,\color{red}{\XSolidBrush}\,}};
            \draw (1.25, 0.1) node(con1) [] {{\footnotesize\,(a) \,}};
		\end{tikzpicture}~~~~~~~~~~~~~
		\begin{tikzpicture}[scale=1.5, line width=0.5pt, inner sep=0.2mm, shorten >=.1pt, shorten <=.1pt]
			%
			\draw (1.875, 1.0) node(L1) [circle, fill=gray!60,draw] {{\footnotesize\,$L_1$\,}};
			\draw (0.0, 0.0) node(X1) [] {{\footnotesize\,$X_{1}$\,}};
			\draw (1.25, 0.0) node(X2) [] {{\footnotesize\,$X_2$\,}};
			\draw (2.5, 0.0) node(X3) [] {{\footnotesize\,$X_3$\,}};
			%
			\draw[-arcsq] (L1) -- (X2) node[pos=0.5,sloped,above] {};
			\draw[-arcsq] (L1) -- (X3) node[pos=0.5,sloped,above] {};
			\draw[-arcsq, thick] (X1) -- (X2) node[pos=0.5,sloped,above] {};
			\draw[-arcsq, thick] (X2) -- (X3) node[pos=0.5,sloped,above] {};
			\draw (1.2, -0.6) node(con1) [] {{\footnotesize\,(b) \,}};
		\end{tikzpicture}
		\caption{(a) Illustrations for the graphical conditions in Theorem \ref{Theorem:GIN graphical}, where only the active paths between nodes are drawn, and dashed lines with {\color{red}{\XSolidBrush}} indicate the absence of edges. (b) An illustrative example of Theorem \ref{Theorem:GIN graphical}. }
		\vspace{-2mm}
		\label{fig:illustration-graphical-criteria}
	\end{center}
\end{figure}

\begin{Example-set}
Let's consider Figure \ref{fig:illustration-graphical-criteria}(b). Let $\mathbf{Y}=\{X_2,X_3\}$ and $\mathbf{Z} = \{X_1\}$.
We observe that there exist a set $\mathcal{S}=\{X_2\}$ such that the active paths between $\mathbf{Z}$ and $\mathbf{Y}$, i.e., $X_1\to X_2\to X_3$ and $X_1\to X_2$ are blocked by $\{X_2\}$ and $X_2$ is on the $\mathbf{Y}$ side of every such path, indicating that $(\{X_1\}, \{X_2, X_3\})$ satisfies GIN. It's important to note that this example does not apply to the graphical criteria of Xie et al. (2020), due to the presence of direct edges between observed variables, which violates the Purity assumption.
\end{Example-set}

We next describe the notion of trek-separation (t-separation), which is a more general separation criterion than d-separation in linear causal models~\citep{Sullivant-T-separation} and can be used to interpret the graphical criteria of the GIN condition. 

\begin{Definition}[\textbf{t-Separation}]
Let $\mathbf{A},\mathbf{B},\mathbf{C_{A}}$, and $\mathbf{C_{B}}$ be four subsets of $\mathbf{V}$. We say the ordered pair $(\mathbf{C_{A}}, \mathbf{C_{B}})$ {t-separates} $\mathbf{A}$ from $\mathbf{B}$ if, for every trek $(\tau_1;\tau_2)$ from a vertex in $\mathbf{A}$ to a vertex in $\mathbf{B}$, either $\tau_1$ contains a vertex in $\mathbf{C_{A}}$ or $\tau_2$ contains a vertex in $\mathbf{C_{B}}$.
\end{Definition}
\revision{
\begin{Theorem}[Graphical Criteria of GIN with t-Separation]\label{Theorem-GIN-representation-Trek-separation}
Let  $\mathbf{Y}$ and $\mathbf{Z}$ be two sets of observed variables of a linear non-Gaussian acyclic causal model which may not be disjoint.
Assume the rank-faithfulness holds. Then $(\mathbf{Z},\mathbf{Y})$ satisfies the GIN condition, if and only if there exists a set $\mathcal{S}$ with $0\leq |\mathcal{S}| \leq \textrm{min}(|\mathbf{Y}|-1, |\mathbf{Z}|)$,
such that 1) for any variable $V$ in $\mathbf{Y}$ but not in $\mathcal{S}$, $V$ is not an ancestor of any variable in $\mathcal{S}$, 
that 2) the ordered pair $(\emptyset, \mathcal{S})$ t-separates $\mathbf{Z}$ and $\mathbf{Y}$,
and that 3) the covariance matrix of $\mathcal{S}$ and $\mathbf{Z}$ has rank $|\mathcal{S}|$, and so does that of $\mathcal{S}$ and $\mathbf{Y}$.
\end{Theorem}
}
Theorem \ref{Theorem-GIN-representation-Trek-separation} shows that the satisfaction of the GIN condition imposes certain structural t-separation constraints.
\revision{
The main difference from the Graphical Criteria of GIN with choke-point set (Theorem \ref{Theorem:GIN graphical}) is that the second condition in Theorem \ref{Theorem:GIN graphical}, i.e., $\mathcal{S}$ is a $\mathbf{Y}$-side choke-point set, can be described by t-separation, i.e., the ordered pair $(\emptyset, \mathcal{S})$ t-separates $\mathbf{Z}$ and $\mathbf{Y}$ (condition 1)) in Theorem \ref{Theorem-GIN-representation-Trek-separation}. \cite{dai2022independence} discusses similar conclusions regarding the graphical criteria of GIN through the use of t-separation.
}
\begin{Example-set}
Continue to consider the structure in Figure \ref{fig:simple-example-explain-GIN-constraint}. We know that there exists a set $\mathcal{S}=\{L_1,L_2\}$ with $0\leq |\mathcal{S}| \leq \textrm{min}(|\mathbf{Y}|-1, |\mathbf{Z}|)$ such that $(\emptyset, \{L_1,L_2\})$ $t$-separates $\{X_3,X_4\}$ and $\{X_1,X_2,X_5\}$, which imply that $(\{X_3,X_4\}, \{X_1,X_2,X_5\})$ satisfies GIN. However, if we replace $X_4$ with $X_6$ in $\mathbf{Z}$, $(\emptyset, \{L_1,L_2\})$ does not $t$-separate $\{X_3,X_6\}$ and $\{X_1,X_2,X_5\}$; note that we here need to have $|\mathcal{S}|=|\{L_1,L_2,L_3\}|>|\mathbf{Y}|-1$. As a result, $(\{X_3,X_6\}, \{X_1,X_2,X_5\})$ violates GIN.
\end{Example-set}

\section{GIN for Estimating Causal Structure of Latent Hierarchical Models}\label{Sec-Application}
In this section, we utilize the GIN condition to estimate the causal structure of latent variable models and highlight its advantages. Specially, we focus on a particular latent variable model, i.e., Linear, Non-Gaussian Latent Hierarchical Model (LiNGLaH), in which latent confounders are also causally related and may even form a hierarchical structure (i.e., some latent variables may not have observed variables as children). The model is defined as follows.

\begin{Definition}[\textbf{Linear Non-Gaussian Latent Hierarchical Model (LiNGLaH)}]
A \\linear non-Gaussian acyclic causal model is called a LiNGLaH, with graph structure $\mathcal{G}=(\mathbf{V},\mathbf{E})$, if it further satisfies the following assumptions:
\begin{itemize}[itemsep=0.2pt,topsep=0.2pt]
    \item[A1.] [Causal Markov Condition] Each variable $V_i$ is independent of all its non-descendants, given its parent $\mathbf{Pa}(V_i)$ in $\mathcal{G}$. 
    \item[A2.] [Measurement Assumption] There is no observed variable in $\mathbf{X}$ being a parent of any latent variables in $\mathbf{L}$.
\end{itemize}
\end{Definition}

A key distinction from existing research on linear latent variable models, such as \cite{bollen1989structural,Silva-linearlvModel}, lies in our incorporation of the \emph{Non-Gaussianity} assumption. This assumption enables us to determine the causal direction among latent variables and imposes weaker structural constraints for identifiability. Furthermore, the prevalence of non-Gaussian data can be anticipated, as supported by the Cramér Decomposition Theorem \citep{Cramer62}, as mentioned in \cite{spirtes2016causal}. Figure \ref{fig:simple-main-example} gives an example of a LiNGLaH.

In the remainder of this section, we first provide sufficient structural conditions that render the causal structure of a LiNGLaH identifiable in light of GIN conditions in Section \ref{SubSection-Structural-Conditions-Identifiability}. Then, we give two essential identification criteria, including detecting latent variables and inferring causal direction among latent variables (Section \ref{Sub-Section-basic-identification-criterias}). We further propose a principled algorithm to estimate LiNGLaH by leveraging the above two criteria in Section \ref{Sub-Section-Recursive-Algorithm}. An example that illustrates the proposed approach is given in Section \ref{Sub-Section-Illustration-Algo}. Then, more practical implementation details of the proposed algorithm are given in Section \ref{Sub-Section-Practical-Algorithm}. We also show that the proposed method can be directly extended to cover causal relationships among observed variables as well in Appendix \ref{Appendix-Subsec-infer-observed}.

\subsection{Structural Conditions for Identifiability of LiNGLaH}\label{SubSection-Structural-Conditions-Identifiability}

It is noteworthy that one may not be able to uniquely identify locations and the number of latent nodes in LiNGLaH without additional assumptions. Several approaches have attempted to handle this issue under specific assumptions, e.g., the measurement model, where each latent variable $L_i$ has a certain number of pure measurement variables as children.\footnote{A set $\mathbf{C}$ is the set of pure children (measurement variables) of $L_i$ if each node in $\mathbf{C}$ has only one latent parent $L_i$, and each node in $\mathbf{C}$ is neither the cause nor the effect of other nodes in $\mathbf{C}$.} Representative methods along this line include  BPC~\citep{Silva-linearlvModel}, noisy ICA-based method~\citep{shimizu2009estimation}, FOFC~\citep{Kummerfeld2016}, CFPC algorithm~\citep{cui2018learning}, and LSTC~\citep{cai2019triad}. In this paper, we consider a clearly more general scenario where latent variables may form a hierarchical structure  (i.e., they may not have observed children), such as the latent variable $L_1$ in Figure \ref{fig:simple-main-example}.
\feng{We here mainly focus on discovering the presence of latent variables and learning causal relationships among latent variables and those between latent and observed variables. We further extend the framework to allow causal relations among measured variables, and this extension will be discussed in Appendix \ref{Appendix-Subsec-infer-observed}. 
}

In the following, we will give a sufficient structural condition that renders the causal structure of a latent hierarchical model identifiable. Specifically, with this condition, the structure among latent variables does not include any ``redundant" latent nodes, and we call such structure the \textit{Minimal Latent Hierarchical Structure}. 

Accordingly, we introduce two notions, \emph{Purity} and \emph{$p\mhyphen$Latent Atomic Structure}, which play a key role in establishing our identifiability results in terms of GIN conditions.
\begin{Definition}[\textbf{Purity}]
Let $\mathcal{L}_1$ be a set of latent variables, and $\mathbf{C}_1$ be a subset of descendant nodes of $\mathcal{L}_1$, i.e., $\mathbf{C}_1 \subset \mathbf{De}(\mathcal{L}_1)$.
We say $\mathbf{C}_1$ is a pure set relative to $\mathcal{L}_1$ iff i) $V_a \CI V_b | \mathcal{L}_1$ for any $V_a,V_b \in \mathbf{C}_1$, and ii) $\mathbf{C}_1 \CI \{\mathbf{V} \backslash \mathbf{De}(\mathcal{L}_1)\} | \mathcal{L}_1$.
In addition, we say a variable $V_c$ in $\mathbf{C}_1$ relative to $\mathcal{L}_1$ is a pure variable if $\mathbf{C}_1$ is a pure set relative to $\mathcal{L}_1$.
\end{Definition}

The notion of purity has also been used in measurement models ~\citep{spirtes2000causation}. Here, we modify and generalize it to latent hierarchical causal models.

\begin{Definition}[\textbf{$p$-Latent Atomic Structure}]
Let $\mathcal{L}_1$ be a latent variable set with $|\mathcal{L}_1|=p$. We say $\mathcal{L}_1$ follows the $p$-latent atomic structure if $\mathcal{L}_1$ has at least $2|p|+1$ neighbors in which $2|p|$ neighbors are pure children relative to $\mathcal{L}_1$, denoted by $\mathbf{S}$, such that (1) $\mathbf{S} \CI \{\mathbf{Ne}(\mathcal{L}_1)\backslash \mathbf{S}\}| \mathcal{L}_1$, and (2) The adjacency matrix $\textrm{Adj}_{\mathcal{L}_1,\mathbf{Ne}(\mathcal{L}_1)}$ has rank $p$.
\end{Definition}

\begin{Condition}[\textbf{Minimal Latent Hierarchical Structure}]\label{cond-1}
Let $\mathcal{G}$ be the DAG associated with a LiNGLaH. We say that the $\mathcal{G}$ is a minimal latent structure if for each latent variable $L_i$, there exists at least one latent set $\mathcal{L}_1$, such that 1) $L_i \in \mathcal{L}_1$, and that 2) $\mathcal{L}_1$ satisfies the $p$-latent atomic structure, where $|\mathcal{L}_1|=p$.
\end{Condition}

\begin{figure}[htp]
	\begin{center}
        \begin{tikzpicture}[scale=1.5, line width=0.5pt, inner sep=0.2mm, shorten >=.1pt, shorten <=.1pt]
		\draw (2, 2.4) node(L1) [circle, fill=gray!60,draw] {{\footnotesize\,$L_1$\,}};
		\draw (0.8, 1.6) node(L2) [circle, fill=gray!60,draw] {{\footnotesize\,$L_2$\,}};
		\draw (1.6, 1.6) node(L3) [circle, fill=gray!60,draw] {{\footnotesize\,$L_3$\,}};
		\draw (2.4, 1.6) node(L4) [circle, fill=gray!60,draw] {{\footnotesize\,$L_4$\,}};
		\draw (3.2, 1.6) node(L5) [circle, fill=gray!60,draw] {{\footnotesize\,$L_5$\,}};
		
		\draw (0.3, 0.8) node(X1) [] {{\footnotesize\,$X_{1}$\,}};
		\draw (0.8, 0.8) node(X2) [] {{\footnotesize\,$X_2$\,}};
		\draw (1.3, 0.8) node(X3) [] {{\footnotesize\,$X_3$\,}};
		\draw (1.8, 0.8) node(X4) [] {{\footnotesize\,$X_4$\,}};
		\draw (2.3, 0.8) node(X5) [] {{\footnotesize\,$X_5$\,}};
		\draw (2.8,0.8) node(X6) [] {{\footnotesize\,$X_6$\,}};
		\draw (3.3, 0.8) node(X7) [] {{\footnotesize\,$X_7$\,}};
		\draw (3.8, 0.8) node(X8) [] {{\footnotesize\,$X_8$\,}};
		\draw[-arcsq] (L1) -- (L2) node[pos=0.5,sloped,above] {};
		\draw[-arcsq] (L1) -- (L3) node[pos=0.5,sloped,above] {};
		\draw[-arcsq] (L1) -- (L4) node[pos=0.5,sloped,above] {};
		\draw[-arcsq] (L1) -- (L5) node[pos=0.5,sloped,above] {};

		\draw[-arcsq] (L2) -- (X1) node[pos=0.5,sloped,above] {};
		\draw[-arcsq] (L2) -- (X2) node[pos=0.5,sloped,above] {};
		\draw[-arcsq] (L3) -- (X3) node[pos=0.5,sloped,above] {};
		\draw[-arcsq] (L3) -- (X4) node[pos=0.5,sloped,above] {};
		\draw[-arcsq] (L4) -- (X5) node[pos=0.5,sloped,above] {};
		\draw[-arcsq] (L4) -- (X6) node[pos=0.5,sloped,above] {};
		\draw[-arcsq] (L4) -- (X7) node[pos=0.5,sloped,above] {};
		\draw[-arcsq] (L4) -- (X8) node[pos=0.5,sloped,above] {};
		\draw[-arcsq] (L5) -- (X5) node[pos=0.5,sloped,above] {};
		\draw[-arcsq] (L5) -- (X6) node[pos=0.5,sloped,above] {};
		\draw[-arcsq] (L5) -- (X7) node[pos=0.5,sloped,above] {};
		\draw[-arcsq] (L5) -- (X8) node[pos=0.5,sloped,above] {};
		\draw (2, 0.3) node(con1) [] {{\footnotesize\,(a) \,}};
		\end{tikzpicture}~~~~~~~
		\begin{tikzpicture}[scale=1.5, line width=0.5pt, inner sep=0.2mm, shorten >=.1pt, shorten <=.1pt]
		\draw [dashed, color=red, line width = 0.9pt] (1.2,2.4) ellipse [x radius=0.40cm, y radius=0.3cm];
		\draw [dashed, color=red, line width = 0.9pt] (2.8,2.4) ellipse [x radius=0.88cm, y radius=0.3cm];
		\draw (2, 3.1) node(L1) [circle, fill=gray!60,draw] {{\footnotesize\,$L_1$\,}};
		\draw (1.2, 2.4) node(L6) [circle, fill=gray!60,draw] {{\footnotesize\,$L_6$\,}};
		\draw (2.4, 2.4) node(L7) [circle, fill=gray!60,draw] {{\footnotesize\,$L_7$\,}};
		\draw (3.2, 2.4) node(L8) [circle, fill=gray!60,draw] {{\footnotesize\,$L_8$\,}};
		\draw (0.7, 1.7) node(L2) [circle, fill=gray!60,draw] {{\footnotesize\,$L_2$\,}};
		\draw (1.6, 1.7) node(L3) [circle, fill=gray!60,draw] {{\footnotesize\,$L_3$\,}};
		\draw (2.6, 1.7) node(L4) [circle, fill=gray!60,draw] {{\footnotesize\,$L_4$\,}};
		\draw (3.4, 1.7) node(L5) [circle, fill=gray!60,draw] {{\footnotesize\,$L_5$\,}};
		
		\draw (0.3, 1) node(X1) [] {{\footnotesize\,$X_{1}$\,}};
		\draw (0.8, 1) node(X2) [] {{\footnotesize\,$X_2$\,}};
		\draw (1.3, 1) node(X3) [] {{\footnotesize\,$X_3$\,}};
		\draw (1.8, 1) node(X4) [] {{\footnotesize\,$X_4$\,}};
		\draw (2.3, 1) node(X5) [] {{\footnotesize\,$X_5$\,}};
		\draw (2.8,1) node(X6) [] {{\footnotesize\,$X_6$\,}};
		\draw (3.3, 1) node(X7) [] {{\footnotesize\,$X_7$\,}};
		\draw (3.8, 1) node(X8) [] {{\footnotesize\,$X_8$\,}};
		\draw[-arcsq] (L1) -- (L6) node[pos=0.5,sloped,above] {};
		\draw[-arcsq] (L1) -- (L3) node[pos=0.5,sloped,above] {};
		\draw[-arcsq] (L6) -- (L2) node[pos=0.5,sloped,above] {};
		\draw[-arcsq] (L1) -- (L7) node[pos=0.5,sloped,above] {};
		\draw[-arcsq] (L1) -- (L8) node[pos=0.5,sloped,above] {};
		\draw[-arcsq] (L7) -- (L4) node[pos=0.5,sloped,above] {};
		\draw[-arcsq] (L7) -- (L5) node[pos=0.5,sloped,above] {};
		\draw[-arcsq] (L8) -- (L4) node[pos=0.5,sloped,above] {};
		\draw[-arcsq] (L8) -- (L5) node[pos=0.5,sloped,above] {};
		\draw[-arcsq] (L2) -- (X1) node[pos=0.5,sloped,above] {};
		\draw[-arcsq] (L2) -- (X2) node[pos=0.5,sloped,above] {};
		\draw[-arcsq] (L3) -- (X3) node[pos=0.5,sloped,above] {};
		\draw[-arcsq] (L3) -- (X4) node[pos=0.5,sloped,above] {};
		\draw[-arcsq] (L4) -- (X5) node[pos=0.5,sloped,above] {};
		\draw[-arcsq] (L4) -- (X6) node[pos=0.5,sloped,above] {};
		\draw[-arcsq] (L4) -- (X7) node[pos=0.5,sloped,above] {};
		\draw[-arcsq] (L4) -- (X8) node[pos=0.5,sloped,above] {};
		\draw[-arcsq] (L5) -- (X5) node[pos=0.5,sloped,above] {};
		\draw[-arcsq] (L5) -- (X6) node[pos=0.5,sloped,above] {};
		\draw[-arcsq] (L5) -- (X7) node[pos=0.5,sloped,above] {};
		\draw[-arcsq] (L5) -- (X8) node[pos=0.5,sloped,above] {};
		\draw (2, 0.5) node(con1) [] {{\footnotesize\,(b) \,}};
		\end{tikzpicture}\\
		\caption{Examples of minimal latent structure. (a) An identifiable latent structure. (b) A non-identifiable latent structure because $L_6$ has fewer neighbor nodes than 3 and $\{L_7,L_8\}$ has fewer neighbor nodes than 5.}
		\vspace{-0.4cm}
		\label{fig:simple-example-minimal-latent-hierarchical-structure} 
	\end{center}
\end{figure}
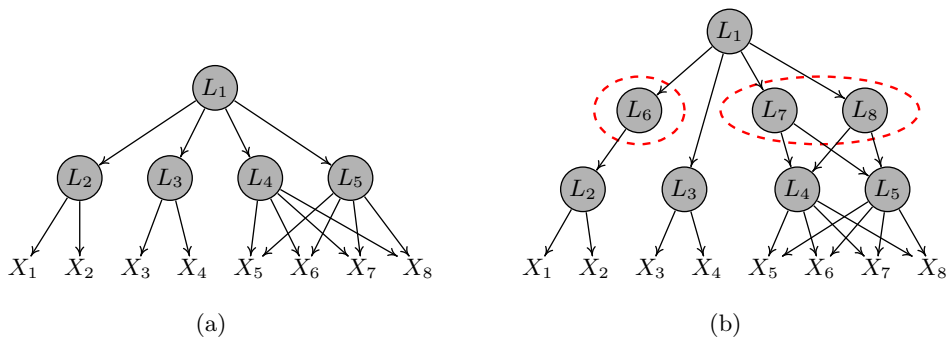

We emphasize that this condition is much milder than existing Tetrad-based methods that deal with latent variable models, such as BPC~\citep{Silva-linearlvModel}, FTFC~\citep{kummerfeld2014causal}, FOFC~\citep{Kummerfeld2016}. Specifically, for the latent variable set $\mathcal{L'}$, we need $2|\mathcal{L'}|$ pure variables as children (those variables may be latent variables), 
while Tetrad-based methods need $2|\mathcal{L'}|+1$ pure variables as children (those variables must be observed variables).
Furthermore, this condition is also much milder than existing Tree-based methods, such as minimal latent tree model~\citep{pearl1988probabilistic,choi2011learning}: we introduce multi-latent sets and allow causal relationships between them, while the latent tree model only considers one-latent set and allows only one path between each pair of nodes.
Figure \ref{fig:simple-example-minimal-latent-hierarchical-structure}(a) shows an example of a minimal latent structure satisfying condition 1. In contrast, Figure \ref{fig:simple-example-minimal-latent-hierarchical-structure}(b) does not satisfy condition 1 because the neighbor nodes of $L_6$ are fewer than  $2|\{L_6\}|+1 = 3$ and the neighbor nodes of $\{L_7,L_8\}$ are fewer than $2|\{L_7,L_8\}|+1 = 5$. Intuitively speaking, for $L_6$, all paths from $L_6$ to its observable descendants $\{X_1,X_2\}$ go through $L_2$ and there is no additional and unique observable descendant relative to $L_2$ to help us to determine $L_6$. Thus $L_6$ is a redundant variable.
A similar result holds for $\{L_7,L_8\}$. 

In the remainder of this section, under Condition 1, i.e., \emph{minimal latent hierarchical structure} condition, we will discuss how to estimate a LiNGLaH by making use of GIN conditions.

\subsection{Two Basic Identification Criteria of LiNGLaH}\label{Sub-Section-basic-identification-criterias}
In this section, we show that one can locate latent variables and infer causal structure among latent variables by making use of GIN conditions in LiNGLaH under Condition \ref{cond-1}. Before giving the technical details of the two identification criteria, we first give the main concepts used in our results.

\begin{Definition}[Causal Cluster / Pure (Impure)]
A set $\mathbf{C}_1$ is a \emph{causal cluster} if the variables in $\mathbf{C}_1$ partially share the same latent parents.\footnote{Here, ``partially" means that variables in $\mathbf{C}_1$ may not share exactly the same set of latent parents but only require some common latent parents} In addition, we say causal cluster $\mathbf{C}_1$ is \emph{pure (impure)} if $\mathbf{C}_1$ is a pure (impure) set relative to $L(\mathbf{C}_1)$.
\end{Definition}

\begin{Definition}[Global Causal Cluster]
Let $\mathbf{C}_1$ be a causal cluster. We say $\mathbf{C}_1$ is a {\textbf{\emph{global causal cluster}}} if (a) $\mathbf{C}_1=\mathbf{Ch}(L(\mathbf{C}_1))$, or (b) $L(\mathbf{C}_1)$ d-separates $\mathbf{C}_1$ and $\mathbf{Ch}(L(\mathbf{C}_1)) \backslash \mathbf{C}_1$. In addition, We say $\mathbf{C}_1$ is a {\textbf{\emph{global pure causal cluster}}} if $\mathbf{C}_1$ is a pure set relative to $L(\mathbf{C}_1)$.
\end{Definition}

\begin{Example-set}
Consider the causal graph in Figure \ref{fig:simple-main-example}, and let $\mathbf{C}_1=\{X_1,X_2,X_3\}$. $\mathbf{C}_1$ is a causal cluster because the variables in $\mathbf{C}_1$ share the same latent parents, i.e., $\{L_5,L_6\}$. $\mathbf{C}_1$ is pure because $\mathbf{C}_1$ is a pure set relative to $L(\mathbf{C}_1)$ ( $X_i \CI X_j | \{L_5,L_6\}$ for any $X_i,X_j \in \mathbf{C}_1$, and $\mathbf{C}_1 \CI \{\mathbf{V} \backslash \mathbf{C}_1\} | \{L_5,L_6\}$).
In addition, $\mathbf{C}_1$ is a global causal cluster because $\{L_5,L_6\}$ d-separates $\mathbf{C}_1$ and $\mathbf{Ch}(L(\mathbf{C})_1) \backslash \mathbf{C}_1=\{X_4\}$.
\end{Example-set}

Note that the global causal cluster is a special kind of causal cluster that will help us quickly locate the latent variables. The following theorem states a basic criterion for identifying global causal clusters from observed data $\mathbf{X}$. 

\begin{Theorem}[Identifying Global Causal Clusters]\label{Theorem-global-cluster}
Let $\mathcal{G}$ be the DAG associated with a LiNGLaH. Let $\mathbf{Y}$ be a set of observed nodes in $\mathcal{G}$. Suppose that rank-faithfulness and Condition \ref{cond-1} hold. Then $\mathbf{Y}$ is a global causal cluster with $|L(\mathbf{Y})|=\mathrm{Len}$  if and only if for any subset $\tilde{\mathbf{Y}}$ of $\mathbf{Y}$ with $|\tilde{\mathbf{Y}}|=\mathrm{Len}+1$, the following two conditions hold: (1) $(\mathbf{X} \backslash {\mathbf{Y}},\tilde{\mathbf{Y}})$ 
follows the GIN condition, and (2) there is no subset $\tilde{\mathbf{Y}'} \subseteq \tilde{\mathbf{Y}}$ such that $(\mathbf{X} \backslash \tilde{\mathbf{Y}'},\tilde{\mathbf{Y}}')$ follows the GIN condition.
\end{Theorem}

Intuitively, Condition 1 of Theorem \ref{Theorem-global-cluster} implies that there exists a latent set $\mathcal{L}_1$ that d-separates $\tilde{\mathbf{Y}}$ from $\mathbf{X} \backslash {\mathbf{Y}}$, and condition 2 of Theorem \ref{Theorem-global-cluster} implies that there does not exist a subset of $\mathcal{L'}_1$ d-separate $\tilde{\mathbf{Y}}'$ from $\mathbf{X} \backslash \tilde{\mathbf{Y}}'$, which ensures that this latent set $\mathcal{L}_1$ is the common latent parents $L(\mathbf{Y})$. 
According to Theorem \ref{Theorem-global-cluster}, one may not only identify whether a set of variables share the same latent parents, but also know the number of latent parents of this cluster.

\begin{Example-set}
Consider the causal graph in Figure \ref{fig:simple-main-example}, and let $\mathbf{Y}=\mathbf{X}_{1:4}$ and $\tilde{\mathbf{Y}}=\mathbf{X}_{1:3}$. We can verify that $(\mathbf{X}_{5:12},\mathbf{X}_{1:3})$ follows the GIN condition and there is no subset $\tilde{\mathbf{Y}'} \subseteq \mathbf{X}_{1:3}$ such that $(\mathbf{X} \backslash \tilde{\mathbf{Y}'},\tilde{\mathbf{Y}}')$ follows the GIN condition. It also holds for any other $\tilde{\mathbf{Y}} \subset \mathbf{Y}$ with $\tilde{\mathbf{Y}}=3$. This will imply that $\mathbf{Y}$ is a global causal cluster.  
In contrast, let $\mathbf{Y}=\mathbf{X}_{1:6}$, so there exist a subset $\tilde{\mathbf{Y}}'=\{X_5,X_6\}$ such that $(\mathbf{X} \backslash \tilde{\mathbf{Y}'},\tilde{\mathbf{Y}}')$ follows the GIN condition, i.e., $L(\tilde{\mathbf{Y}}')= L_7$ d-separates $\tilde{\mathbf{Y}}'$ from $\mathbf{X} \backslash \tilde{\mathbf{Y}}'$. This will imply that $\mathbf{Y}$ is not a global causal cluster.
\end{Example-set}

We now discuss how to identify the causal direction among latent variables given their corresponding children. The following theorem shows the asymmetry between the underlying latent variables in terms of the GIN condition.


\begin{Theorem}[Identifying Causal Directions among Latent Variables]\label{Theorem-causal-direction}
Let $\mathcal{G}$ be the DAG associated with a LiNGLaH, and let $\mathbf{C}_p$ and $\mathbf{C}_q$ be two global pure causal clusters in $\mathcal{G}$. Suppose there are no confounders behind $L(\mathbf{C}_p)$ and $L(\mathbf{C}_q)$, and $L(\mathbf{C}_p) \cap L(\mathbf{C}_q)=\emptyset$.
Further suppose that $\mathbf{C}_p$ contains $2|L(\mathbf{C}_p)|$ number of variables with $\mathbf{C}_p=P_{1:2|L(\mathbf{C}_p)|}$ and that $\mathbf{C}_q$ contains $2|L(\mathbf{C}_q)|$ number of variables with $\mathbf{C}_q=Q_{1:2|L(\mathbf{C}_q)|}$.
Then if $(\{P_{|L(\mathbf{C}_p)|+1},...,P_{2|L(\mathbf{C}_p)|}\},\{P_{1:|L(\mathbf{C}_p)|},Q_{1},...,Q_{|L(\mathbf{C}_q)|}\})$ follows the GIN condition, $L(\mathbf{C}_p) \to L(\mathbf{C}_q)$ holds.
\end{Theorem}

Theorem \ref{Theorem-causal-direction} tells us how to infer the causal direction between two latent sets given their corresponding children.

\begin{Example-set}
Consider the graph in Figure \ref{fig:simple-example-explain-GIN-constraint}.
Let $\mathcal{L}_p=\{L_1,L_2\}$ and $\mathcal{L}_q=\{L_3\}$. Then $\{X_1,X_2,X_3,X_4\}$ and $\{X_5,Y_6\}$ are their corresponding children respectively. There is no confounders between $\mathcal{L}_p$ and $\mathcal{L}_q$, we can verify and obtain $(\{X_3,X_4\},\{X_1,X_2,X_5\})$ follows the GIN condition. This will imply that $\mathcal{L}_p \to \mathcal{L}_q$, i.e., $\{L_1,L_2\} \to \{L_3\}$.
\end{Example-set}

\subsection{Structure Identification of LiNGLaH}\label{Sub-Section-Recursive-Algorithm}
In this section, we leverage the above two criteria and propose an algorithm, \emph{\textbf{La}tent \textbf{Hi}erarchical \textbf{Ca}ausal \textbf{S}tructure \textbf{L}earning (LaHiCaSl)},
for estimating the structure of LiNGLaH. In particular, we first briefly describe the LaHiCaSl algorithm in Section~\ref{Sub-section-framework}, with more details provided in Sections~\ref{subsection-phase-I} and \ref{subsection-phase-II}. Finally, we show the identifiability of causal structure with LiNGLaH in terms of GIN conditions in Section~\ref{Subsection-summary-identification-correctness}.

\subsubsection{Latent Hierarchical Causal Structure Learning (LaHiCaSl)}\label{Sub-section-framework}
The LaHiCaSl algorithm contains two main phases, including locating latent variables (Phase I) and inferring causal structure among the identified latent variables (Phase II). The algorithm is designed with the following rules, which will be proved later: (1) all latent variables can be correctly discovered given the observed variables alone and no redundant latent variables will be introduced (Section \ref{subsection-phase-I}), and (2) the causal structure of the identified latent variables can be uniquely determined, including the causal direction (Section \ref{subsection-phase-II}). The entire process is summarized in Algorithm \ref{algorithm-framework}.

\begin{algorithm}[htb]
	\caption{Latent Hierarchical Causal Structure Learning (LaHiCaSl)}
	\label{algorithm-framework}
		\begin{algorithmic}[1]
            \REQUIRE
	A set of observed variables $\mathbf{X}$\\
			\STATE // \emph{Phase I: Locate latent variables}
			\STATE Partially determined causal structure $\mathcal{G} \leftarrow$ LocateLatentVariables($\mathbf{X}$); 
			\STATE // \emph{Phase II: Infer causal structure among latent variables}
			\STATE  Causal structure $\mathcal{G} \leftarrow$ LocallyInferCausalStructure($\mathbf{X}$, $\mathcal{G}$); 
        \ENSURE
	Causal structure $\mathcal{G}$ over both observed and latent variables
		\end{algorithmic}
\end{algorithm}

\revision{
\begin{Remark}
    It should be noted that the LaHME (Latent Hierarchical Model Estimation) algorithm, as described in \cite{xie2022identification}, can be considered as a specific instance of the proposed LaHiCaSl algorithm when the size of the latent variable set is restricted to 1. This implies that the algorithm proposed here is more general compared to that of \cite{xie2022identification}, as it accommodates multiple latent confounders behind any two variables, exemplified by the latent variable set $\{L_4, L_5\}$ illustrated in Figure \ref{fig:simple-example-minimal-latent-hierarchical-structure}.
\end{Remark}
}

Below, we provide the technical details of the two phases of the framework.

\subsubsection{Phase I: Locating  Latent  Variables}\label{subsection-phase-I}
We leverage a recursive procedure to locate latent variables from observed variables. Specifically, at each iteration, it contains the following three steps: 
\begin{description}[itemsep=0.2pt,topsep=0.2pt]
    \item I-S1. Identifying the global causal clusters from the \textbf{\emph{active variable set}}.\footnote{We say a set $\mathcal{A}$ is active if it is selected in the current iteration. In the first iteration, the active set $\mathcal{A}$ is $\mathbf{X}$.}
    \item I-S2. Determining the number of new latent variables that need to be introduced for these identified clusters.
    \item I-S3. Updating the active variable set.
\end{description} 
These three steps are repeated iteratively until all latent variables of the system are discovered, with the complete procedure summarized in Algorithm \ref{algorithm-LocateLatentVariables}. An illustrative example of each step will be given immediately after introducing each step in the following subsections, and a complete example is given in Section \ref{Sub-Section-Illustration-Algo}.

\begin{algorithm}[htb]
	\caption{LocateLatentVariables}
	\label{algorithm-LocateLatentVariables}
		\begin{algorithmic}[1]
            \REQUIRE
	A set of observed variables $\mathbf{X}$\\
			\STATE Initialize active set $\mathcal{A} = \mathbf{X}$, and $\mathcal{G} = \emptyset$;
		    \WHILE{$\mathcal{A} \neq \emptyset$}
			\STATE $\mathrm{ClusterList} \leftarrow$ IdentifyGlobalCausalClusters($\mathcal{A}$); // I-S1
			\STATE $\mathcal{G} \leftarrow $ DetermineLatentVariables($\mathrm{ClusterList}$, $\mathcal{A}$, $\mathcal{G}$); // I-S2
			\STATE $\mathcal{A} \leftarrow$ UpdateActiveData($\mathbf{X}$, $\mathcal{A}$, $\mathcal{G}$) // I-S3
			\ENDWHILE
	\ENSURE
	Partial causal structure $\mathcal{G}$
		\end{algorithmic}
	\vspace{-0.1cm}
\end{algorithm}

\emph{I-S1: Identifying Global Causal Clusters}

In this section, we mainly deal with the identification of global causal clusters. According to Theorem \ref{Theorem-global-cluster}, the global causal clusters in the current active variable set $\mathcal{A}$ can be identified by the following proposition.

\begin{Proposition}
[Identifying Global Causal Clusters]\label{proposition-identify-global-cluster}
Let $\mathcal{A}$ be the active variable set and $\mathbf{Y}$ be a proper subset of $\mathcal{A}$. Suppose rank-faithfulness and Condition \ref{cond-1} hold. Then $\mathbf{Y}$ is a global causal cluster with $|L(\mathbf{Y})|=\mathrm{Len}$ if and only if for any subset $\tilde{\mathbf{Y}}$ of $\mathbf{Y}$ with $|\tilde{\mathbf{Y}}|=\mathrm{Len}+1$, the following two conditions hold: (1) $(\mathbf{X} \backslash {\mathbf{Y}},\tilde{\mathbf{Y}})$ follows the GIN condition, and (2) there is no subset $\tilde{\mathbf{Y}'} \subseteq \tilde{\mathbf{Y}}$ such that $(\mathbf{X} \backslash \tilde{\mathbf{Y}'},\tilde{\mathbf{Y}}')$ follows the GIN condition.
\end{Proposition}

To efficiently identify global causal clusters $\mathbf{Y}$, we start with finding clusters with a single latent variable, and then increase the number of considered latent variables until the conditions in Proposition \ref{proposition-identify-global-cluster} are satisfied or the length of the set of latent variables equals to $|\mathbf{Y}|-1$. The details of the above process are given in Algorithm \ref{algorithm-find-causal-cluster}, and an illustrative example is given accordingly.

\begin{algorithm}[htb]
	\caption{IdentifyGlobalCausalClusters}
	\label{algorithm-find-causal-cluster}
	\vspace{-0.1cm}
		\begin{algorithmic}[1]
        \REQUIRE An active variable set $\mathcal{A}$\\
			\STATE Initialize a cluster set $\mathrm{ClusterList}= \emptyset$ and the group size $\mathrm{GrLen}=2$;
			\WHILE{$|\mathcal{A}| \geq 2\times \mathrm{GrLen} - 1$}
            \REPEAT
            \STATE Select a subset $\mathbf{Y}$ from $\mathcal{A}$ such that $|\mathbf{Y}|=\mathrm{GrLen}$;
			\FOR {$\mathrm{LaLen}=1:\mathrm{GrLen}-1$}
			\IF {$(\mathcal{A} \backslash {\mathbf{Y}}, \tilde{\mathbf{Y}}) $ follows GIN condition for $\forall \tilde{\mathbf{Y}} \in \mathbf{Y}$ such that $|\tilde{\mathbf{Y}}|=\mathrm{LaLen}+1$}
			\STATE $L(\mathbf{Y})={\mathrm{LaLen}}$;
			\STATE Add $\mathbf{Y}$ into $\mathrm{ClusterList}$;
			\STATE Break the for loop of line 5;
			\ENDIF
			\ENDFOR
			\UNTIL{{\color{black}{all subsets with group length $\mathrm{GrLen}$ in $\mathcal{A}$ have been selected;}}}
			\STATE $\mathcal{A}=\mathcal{A} \backslash \mathrm{ClusterList}$, and $\mathrm{GrLen} \leftarrow \mathrm{GrLen}+1$;
			\ENDWHILE
	\ENSURE A cluster set $\mathrm{ClusterList}$
		\end{algorithmic}
\end{algorithm}
\begin{Example-set}
Consider the causal structure in Figure \ref{fig:simple-main-example}. Suppose the active variable set is $\mathcal{A} = X_{1:13}$. We first set $\mathrm{GrLen}=2$ and $\mathrm{LaLen}=1$, and can find five clusters, i.e., $\mathbf{C}_1=\{X_5,X_6\}$, $\mathbf{C}_2=\{X_7,X_8\}$, $\mathbf{C}_3=\{X_9,X_{10}\}$, $\mathbf{C}_4=\{X_9,X_{11}\}$, and $\mathbf{C}_5=\{X_{10},X_{11}\}$. Then, we set $\mathrm{GrLen}=3$ and $\mathrm{LaLen}=2$, and find four clusters, i.e., $\mathbf{C}_6=\{X_1,X_2,X_3\}$, $\mathbf{C}_7=\{X_1,X_2,X_4\}$, $\mathbf{C}_8=\{X_1,X_3,X_4\}$, and $\mathbf{C}_9=\{X_2,X_3,X_4\}$. 
\end{Example-set}

\emph{I-S2: Determining Latent Variables}

We then determine how many new latent variables need to be introduced for these clusters identified in Algorithm \ref{algorithm-find-causal-cluster}. To this end, we need to deal with the following two issues: 
\begin{itemize}[itemsep=0.2pt,topsep=0.2pt]
    \item which clusters of variables share the common (subset of) latent parents and should be merged, and
    \item which clusters of variables are the children of the latent variables that have been introduced in the previous iterations.
\end{itemize}
We now give the following two 
lemmas on identifying pure and impure (sub-) clusters, which will help us to address the above two issues.

\begin{Lemma}[Identifying Pure/Impure Cluster]\label{lemma-pure-impure-cluster}
Let $\mathcal{A}$ be the active variable set and $\mathbf{C}_1$ be a global causal cluster with $|\mathbf{C}_1|=k$. Then $\mathbf{C}_1$ is a pure causal cluster relative to $L(\mathbf{C}_1)$ if one of the following conditions are satisfied:
\begin{itemize}[itemsep=0.2pt,topsep=0.2pt]
    \item[(1)] $k>|L(\mathbf{C}_1)|+1$, and for any subset $\tilde{\mathbf{C}}_1$ of $\mathbf{C}_1$ such that $|\tilde{\mathbf{C}}_1|=|L(\mathbf{C}_1)|+1$, and that $(\mathcal{A} \cup \{\mathbf{C}_1 \backslash \tilde{\mathbf{C}}_1\},\tilde{\mathbf{C}}_1)$ follows the  GIN condition.
    \item[(2)] $k=|L(\mathbf{C}_1)|+1$, and for any ordered pair of variables $\{V_i,V_j\} \subset \mathbf{C}_1$, there does not exist $\{\mathbf{P},V_k\} \subset \{\mathcal{A} \backslash \mathbf{C}_1\}$, such that $|\mathbf{P}|=|L(\mathbf{C}_1)|$, and that  $(\{V_i,\mathbf{P}\},\{\mathbf{C}_1, V_k\})$ follows the GIN condition while $(\{V_j,\mathbf{P}\},\{\mathbf{C}_1, V_k\})$ violates the GIN condition.
\end{itemize}
Otherwise, $\mathbf{C}_1$ is an impure cluster.
\end{Lemma}

Intuitively, given a global cluster $\mathbf{C}_1$, Condition 1 of Lemma \ref{lemma-pure-impure-cluster} tests whether $L(\mathbf{C}_1)$ d-separates $\tilde{\mathbf{C}}_1$ from $(\mathcal{A} \cup \{\mathbf{C}_1 \backslash \{\tilde{\mathbf{C}}_1\}\}$ for any subset $\tilde{\mathbf{C}}_1$ of $\mathbf{C}_1$. Condition 2 of Lemma \ref{lemma-pure-impure-cluster} tests whether $L(\mathbf{C}_1)$ d-separates $V_i$ from $V_j$ for any ordered pair of variables $\{V_i,V_j\} \subset \mathbf{C}_1$. Note that one impure cluster $\mathbf{C}_1$ may contain a pure subset cluster. In Figure \ref{fig:simple-main-example}, for example, $\mathbf{C}_1=\{L_2,L_3,L_4,L_8\}$ is an impure cluster relative to $L_1$ but $\{L_2,L_3,L_8\}$ is a pure sub-cluster of $\mathbf{C}_1$ relative to $L_1$. Given an impure cluster $\mathbf{C}_1$, the next lemma gives the conditions for identifying whether a sub-cluster of $\mathbf{C}_1$ is pure relative to $L(\mathbf{C}_1)$.

\begin{Lemma}[Identifying Pure sub-Cluster] \label{lemma-pure-impure-variables}
Let $\mathbf{C}_1$ be an impure global causal cluster and $\mathcal{A}$ be the active variable set. Further let $\tilde{\mathbf{C}}_1$ ($|\tilde{\mathbf{C}}_1| \geq 2$) be a subset of $\mathbf{C}_1$. Then $\tilde{\mathbf{C}}_1$ is a pure cluster if one of the following conditions is satisfied:
\begin{itemize}[itemsep=0.2pt,topsep=0.2pt]
    \item[(1)] $|\tilde{\mathbf{C}}_1| \geq |L(\mathbf{C}_1)|+1$, and conditions (1) or (2) of Lemma \ref{lemma-pure-impure-cluster} holds.
    \item[(2)] $|\tilde{\mathbf{C}}_1| < |L(\mathbf{C}_1)|+1$, and for any ordered pair of variables $\{V_i,V_j\} \subset \tilde{\mathbf{C}}_1$, there does not exist $\{\mathbf{P},\mathbf{Q}\} \subset \{\mathcal{A} \backslash \mathbf{C}_1\}$, such that $|\mathbf{P}|=|L(\mathbf{C}_1)|$ and $|\mathbf{Q}|=|L(\mathbf{C}_1)|-|\tilde{\mathbf{C}}_1|+1$, and that  $(\{V_i,\mathbf{P}\},\{\tilde{\mathbf{C}}_1, \mathbf{Q}\})$ follows the GIN condition while $(\{V_j,\mathbf{P}\},\{\tilde{\mathbf{C}}_1, \mathbf{Q}\})$ violates the GIN condition.
\end{itemize}
Otherwise, $\tilde{\mathbf{C}}_1$ is an impure cluster.
\end{Lemma}

We next define the \textit{permissible set}, which will help us to quickly find the maximal pure set relatively a subset in a cluster. 

\begin{Definition}[Permissible Set]
Let $\tilde{\mathbf{C}}_1$ be a sub-cluster of $\mathbf{C}_1$. We use notation $\neg_{\mathbf{C}_1}(\tilde{\mathbf{C}}_1)$ to denote the \emph{permissible} set relative to $\tilde{\mathbf{C}}_1$ in $\mathbf{C}_1$ if $\neg_{\mathbf{C}_1}(\tilde{\mathbf{C}}_1)$ is the maximal set of $\mathbf{C}_1$ such that any variables $V_i \in \tilde{\mathbf{C}}_1$ and $V_j \in \neg_{\mathbf{C}_1}(\tilde{\mathbf{C}}_1)$,  $\{V_i,V_j\}$ is a pure set relative to $L(\mathbf{C}_1)$. 
\end{Definition}
For instance, in Figure \ref{fig:simple-main-example}, let $\tilde{\mathbf{C}}_1=\{L_2,L_8\}$ be a sub-cluster of $\mathbf{C}_1=\{L_2,L_3,L_4,L_8\}$. Then $\neg_{\mathbf{C}_1}(\tilde{\mathbf{C}}_1)=\{L_3\}$. Note that if $\mathbf{C}_1$ is a pure cluster, then $\neg_{\mathbf{C}_1}(\tilde{\mathbf{C}}_1)$ is the complementary set of $\tilde{\mathbf{C}}_1$. 

We now provide the conditions under which the clusters of variables share the common (subset of) latent parents and should be merged to address the first issue.

\begin{Proposition}[Merging Rules]\label{Proposition-merge-rules-cases}
Let $\mathcal{A}$ be the active variable set and $\mathbf{C}_1$ and $\mathbf{C}_2$ be two global causal clusters. Then the following rules hold.
\begin{description}[itemsep=0.2pt,topsep=0.2pt]
    \item  $\mathcal{R}1.$ If (a) $|L(\mathbf{C}_1)|=|L(\mathbf{C}_2)|$, and (b) 
    for any subset $\tilde{\mathbf{C}} \subseteq \{\mathbf{C}_1 \cup \mathbf{C}_2$\} with $|\tilde{\mathbf{C}}|=|L(\mathbf{C}_1)|$, $(\{\mathcal{A} \backslash \tilde{\mathbf{C}}\} \cup \neg_{\mathbf{C}_1}(\tilde{\mathbf{C}}_1),\tilde{\mathbf{C}})$ follows the GIN condition, then $\mathbf{C}_1$ and $\mathbf{C}_2$ share the same set of latent variables as parents, i.e., $L(\mathbf{C}_1) = L(\mathbf{C}_2)$.
    \item  $\mathcal{R}2.$ If (a) $|L(\mathbf{C}_1)| \neq |L(\mathbf{C}_2)|$ (suppose $|L(\mathbf{C}_1)| > |L(\mathbf{C}_2)|$), and (b) for any subset $\tilde{\mathbf{C}}_1$ of $\mathbf{C}_1$ such that $|\tilde{\mathbf{C}}_1|=|L(\mathbf{C}_1)|$ and any $V_i \in {\mathbf{C}}_2$ and $V_i \notin \tilde{\mathbf{C}}_1$, $(\mathcal{A} \backslash \{ {\mathbf{C}}_1 \cup \mathbf{C}_2\} \cup \neg_{\mathbf{C}_1}(\tilde{\mathbf{C}}_1) \cup \neg_{\mathbf{C}_2}(V_i),\{V_i, \tilde{\mathbf{C}}_1\})$ follows the GIN condition, then the common parents of $\mathbf{C}_1$ contains the common parents of $\mathbf{C}_2$, i.e., $L(\mathbf{C}_1) \subset L(\mathbf{C}_2)$.
\end{description}
Otherwise, $\mathbf{C}_1$ and $\mathbf{C}_2$ do not share the common (subset of) latent variables as parents.
\end{Proposition}

Next, we discuss the solution of the second issue. Due to the property of hierarchical structure, we can not guarantee that all children of a latent variable are identified at the same iteration. Thus, we need to identify whether a new cluster's parents have been introduced in previous iterations. 
Fortunately, for any latent variable set $\mathcal{L}_1$ that was introduced in previous iterations, we know that all nodes in the active variable set $\mathcal{A}$ in the current iteration are causally earlier than the children of $\mathcal{L}_1$ found in the previous iteration. That is to say, $\mathbf{Ch}(\mathcal{L}_1)$ are leaf nodes in the subgraph with variables $\mathcal{A} \cup \mathbf{Ch}(\mathcal{L}_1)$.
This yields the following corollary derived from Proposition \ref{Corollary-merge-rules-earlycurrent-cases}.

\begin{Corollary}\label{Corollary-merge-rules-earlycurrent-cases}
Let $\mathcal{L}_1$ be a latent variable set that has been introduced in the previous iterations, $\mathbf{C}_2$ be a new cluster, and $\mathcal{A}$ be the active variable set in the current iteration. Further, let $\mathbf{C}_1$ be the set of children of $\mathcal{L}_1$ that have been found. Then the following rules hold.
\begin{description}[itemsep=0.2pt,topsep=0.2pt]
    \item  $\mathcal{R}3.$ If (a) $|L(\mathbf{C}_2)|=|\mathcal{L}_1|$, and (b) for any subset $\tilde{\mathbf{C}}_1$ of $\mathbf{C}_1$ such that $|\tilde{\mathbf{C}}_1|=|\mathcal{L}_1|$ and any $V_i \in {\mathbf{C}}_2$ and $V_i \notin \tilde{\mathbf{C}}_1$, $(\mathcal{A} \backslash \{ \mathcal{L}_1 \cup \mathbf{C}_2\} \cup \neg_{\mathbf{C}_1}(\tilde{\mathbf{C}}_1) \cup \neg_{\mathbf{C}_2}(V_i),\{V_i, \tilde{\mathbf{C}}_1\})$ follows the GIN condition, then the common latent parents of $\mathbf{C}_2$ is $\mathcal{L}_1$, i.e., $L(\mathbf{C}_2) = \mathcal{L}_1$.
    \item  $\mathcal{R}4.$ If (a) $|L(\mathbf{C}_2)| \neq |\mathcal{L}_1|$ (suppose $|\mathcal{L}_1|>|L(\mathbf{C}_2)|$), and (b) for any subset $\tilde{\mathbf{C}}_1$ of $\mathbf{C}_1$ such that $|\tilde{\mathbf{C}}_1|=|\mathcal{L}_1|$ and any $V_i \in {\mathbf{C}}_2$ and $V_i \notin \tilde{\mathbf{C}}_1$, $(\mathcal{A} \backslash \{ \mathcal{L}_1 \cup \mathbf{C}_2\} \cup \neg_{\mathbf{C}_1}(\tilde{\mathbf{C}}_1) \cup \neg_{\mathbf{C}_2}(V_i),\{V_i, \tilde{\mathbf{C}}_1\})$ follows the GIN condition, then $\mathcal{L}_1$ contains the common parents of $\mathbf{C}_2$, i.e., $L(\mathbf{C}_2) \subset \mathcal{L}_1$.
\end{description}
\end{Corollary}

The complete procedure of determining latent variables for the current active variable set is summarized in Algorithm \ref{algorithm-merge-cluster}, and an illustrated example is given in Example \ref{Example-determinie-latent-variable}.
\begin{algorithm}[htb]
	\caption{DetermineLatentVariables}
	\label{algorithm-merge-cluster}
	\vspace{-0.1cm}
		\begin{algorithmic}[1]
        \REQUIRE A cluster set $\mathrm{ClusterList}$, active variable set $\mathcal{A}$, and partial graph $\mathcal{G}$\\
		    \STATE Initialize set $\mathbf{C}= \emptyset$ and $\mathcal{G}'=\mathcal{G}$;
			\STATE $\mathbf{C} \leftarrow $ Merge clusters from $\mathrm{ClusterList}$ according to the Rules $\mathcal{R}1$ and $\mathcal{R}2$ of Proposition \ref{Proposition-merge-rules-cases};
			\FOR {each $\mathbf{C}_{i} \in \mathbf{C}$}
			\STATE $TagVar=TRUE$;
			\FOR {each latent set $\mathcal{L}_{j}$ in $\mathcal{G}'$}
			\IF{$\mathcal{L}_j$ and $\mathbf{C}_{i}$ satisfy $\mathcal{R}3$ of Corollary \ref{Corollary-merge-rules-earlycurrent-cases}}
			\STATE $\mathcal{G}=\mathcal{G} \cup \{\mathcal{L}_{j} \to V_i | V_i \in \mathbf{C}_i\}$;
			\STATE $TagVar=FALSE$;
			\STATE Break the for loop of line 5;
			\ELSIF{$|\mathcal{L}_j|>L(\mathbf{C}_i)$ and $\mathcal{L}_j$ and $\mathbf{C}_{i}$ satisfy $\mathcal{R}4$ of Corollary \ref{Corollary-merge-rules-earlycurrent-cases}}
			\STATE $\mathcal{G}=\mathcal{G} \cup \{\mathcal{L}'_{j} \to V_i | V_i \in C_i\}$, where $\mathcal{L}'_{j} \subset \mathcal{L}_{j}$ and $|\mathcal{L}'_{j}|=L(\mathbf{C}_i)$ ;
			\STATE $TagVar=FALSE$;
			\STATE Break the for loop of line 5;
			\ELSIF{$|\mathcal{L}_j|< L(\mathbf{C}_i)$ and $\mathcal{L}_j$ and $\mathbf{C}_{i}$ satisfy $\mathcal{R}4$ of Corollary \ref{Corollary-merge-rules-earlycurrent-cases}}
			\STATE Introduce a new latent set $\mathcal{L}_{k}$ such that $|\mathcal{L}_{k}|=|L(\mathbf{C}_i)|-|\mathcal{L}_{j}|$;
			\STATE $\mathcal{G}=\mathcal{G} \cup \{\{\mathcal{L}_{j} \cup \mathcal{L}_{k}\} \to V_i | V_i \in C_i\}$;
			\STATE $TagVar=FALSE$;
			\STATE Break the for loop of line 5;
			\ENDIF
			\ENDFOR
			\IF{$TagVar=TRUE$}
			\STATE Introduce a new latent set $\mathcal{L}_k$ with length $|L(\mathbf{C}_i)|$ into $\mathcal{G}$;
			\STATE $\mathcal{G}=\mathcal{G} \cup \{\mathcal{L}_{k} \to V_i | V_i \in \mathbf{C}_i\}$;
			\ENDIF
			\ENDFOR
        \ENSURE Updated partial graph $\mathcal{G}$
		\end{algorithmic}
\end{algorithm}

\begin{Example-set}\label{Example-determinie-latent-variable}
Continue to consider the structure in Figure \ref{fig:simple-main-example}, we have found 9 clusters by Algorithm \ref{algorithm-find-causal-cluster}: $\mathbf{C}_1=\{X_5,X_6\}$, $\mathbf{C}_2=\{X_7,X_8\}$, $\mathbf{C}_3=\{X_9,X_{10}\}$, $\mathbf{C}_4=\{X_9,X_{11}\}$, $\mathbf{C}_5=\{X_{10},X_{11}\}$, $\mathbf{C}_6=\{X_1,X_2,X_3\}$, $\mathbf{C}_7=\{X_1,X_2,X_4\}$, $\mathbf{C}_8=\{X_1,X_3,X_4\}$, and $\mathbf{C}_9=\{X_2,X_3,X_4\}$. Now, according to $\mathcal{R}1$ of Proposition \ref{Proposition-merge-rules-cases}, we have that $\mathbf{C}_3$, $\mathbf{C}_4$ and $\mathbf{C}_5$ are merged, and that $\mathbf{C}_6$, $\mathbf{C}_7$, $\mathbf{C}_8$, and $\mathbf{C}_9$ are merged. For any other two clusters, we can not merge them by Proposition \ref{Proposition-merge-rules-cases}. Furthermore, because there exist no latent variables that have been introduced in the previous iterations, we do not need to verify the rules of Corollary \ref{Corollary-merge-rules-earlycurrent-cases}. Overall, we can determine there are four latent variable sets, including three 1-latent sets $L_7=L(\mathbf{C}_1)$, $L_8=L(\mathbf{C}_2)$, and $L_9=L(\mathbf{C}_3 \cup \mathbf{C}_4 \cup \mathbf{C}_5)$, and one 2-latent set $\{L_5,L_6\}=L(\mathbf{C}_6 \cup \mathbf{C}_7 \cup \mathbf{C}_8 \cup \mathbf{C}_9)$. 
\end{Example-set}

\emph{I-S3: Updating Active Data Set}

When the active variable set $\mathcal{A}$ is the observed variable set $\mathbf{X}$, one can identify some specific latent variables that are the parents of observed variables, with Algorithm \ref{algorithm-find-causal-cluster} and Algorithm \ref{algorithm-merge-cluster}. However, one may have the following concerns: for a hierarchical structure, how can we further find latent variables that are the parents of latent variables and how can we check the GIN conditions over latent variables without observing them? Thanks to the transitivity of linear causal relations, we can use their observed descendants to test for the GIN conditions.  For instance, consider the structure in Figure \ref{fig:simple-main-example}. Suppose $\mathbf{Y}=\{L_9,X_{13}\}$, $(\mathbf{X} \backslash \{X_{9:11},X_{13}\},\{L_9,X_{13}\})$ follows GIN condition if and only if 
$(\mathbf{X} \backslash \{X_{9:11},X_{13}\},\{X_9,X_{13}\})$ follows it, where the measured descendant $X_9$ acts as a surrogate of the latent variable $L_4$. Thus, we can check subsequent GIN conditions over latent variables by using their proper pure observed descendants. For simplicity, we say $(\mathbf{Z}',\mathbf{Y}')$ is the surrogate pair of $(\mathbf{Z},\mathbf{Y})$ if the GIN test for $(\mathbf{Z}',\mathbf{Y}')$ is equivalent to the GIN test for $(\mathbf{Z},\mathbf{Y})$. The definition of \emph{surrogate sets of $(\mathbf{Z},\mathbf{Y})$} is given in the following definition and the correctness of this testing is given in Proposition \ref{Propostion-testing-GIN-using-Surrogatevariables}.

\begin{Definition}[Surrogate Pair of $(\mathbf{Z},\mathbf{Y})$]
Let $\mathcal{G}$ be the DAG associated with a LiNGLaH, and $\mathbf{Y}$, $\mathbf{Z}$ be two sets of nodes in $\mathcal{G}$.
Then the surrogate  pair of $(\mathbf{Z},\mathbf{Y})$, denoted by $(\mathbf{Z}',\mathbf{Y}')$, is generated as follows:
\begin{itemize}[itemsep=0.2pt,topsep=0.2pt]
    \item [a.] add all observed nodes in $\mathbf{Z}$ and $\mathbf{Y}$ into $\mathbf{Z}'$ and $\mathbf{Y}'$, respectively;
    \item [b.] for each latent set $\mathcal{L}_1$ such that $\mathcal{L}_1 \subset \mathbf{Z}$ and $\mathcal{L}_1 \subset \mathbf{Y}$, add $2|\mathcal{L}_1|$ disjoint and pure observed descendants relative to $\mathcal{L}_1$ into $\mathbf{Z}'$ and $\mathbf{Y}'$, respectively;
    \item [c.] for each latent set $\mathcal{L}_1$ such that $\mathcal{L}_1 \subset \mathbf{Z}$ but $\mathcal{L}_1 \notin \mathbf{Y}$, add $2|\mathcal{L}_1|$ disjoint and pure observed descendants relative to $\mathcal{L}_1$ into $\mathbf{Z}'$;
     \item [d.] for each latent set $\mathcal{L}_1$ such that $\mathcal{L}_1 \subset \mathbf{Y}$ but $\mathcal{L}_1 \notin \mathbf{Z}$, add $2|\mathcal{L}_1|$ disjoint and pure observed descendants relative to $\mathcal{L}_1$ into $\mathbf{Y}'$.
\end{itemize}
\end{Definition}
Note that the surrogate pair of $(\mathbf{Z},\mathbf{Y})$ is $(\mathbf{Z}',\mathbf{Y}')$ when all node in $\mathbf{Z} \cup \mathbf{Y}$ are observed nodes.

\begin{Proposition}[Testing GIN over Latent Variables]\label{Propostion-testing-GIN-using-Surrogatevariables}
Let $\mathcal{G}$ be the DAG associated with a LiNGLaH. Let $\mathbf{Y}$, $\mathbf{Z}$ be two sets of nodes in $\mathcal{G}$, and $(\mathbf{Z}',\mathbf{Y}')$ be the surrogate pair of $(\mathbf{Z},\mathbf{Y})$ in $\mathcal{G}$. Then, the GIN test for $(\mathbf{Z},\mathbf{Y})$ is equivalent to the GIN test for $(\mathbf{Z}',\mathbf{Y}')$.
\end{Proposition}

The following proposition shows how to update the active variable set during the search procedure.

\begin{Proposition}[Active Variable Set Update]\label{Proposition-update-active-data}
Let $\mathcal{A}$ be the current active variable set and $\mathcal{L}$ be the latent variable sets discovered in the current iteration. Then the updated active variable set $\mathcal{A}'=\mathcal{A} \cup \mathcal{L} \backslash \mathbf{Ch}(\mathcal{L})$. Moreover, the GIN test of any pair $(\mathbf{Z},\mathbf{Y})$ in $\mathcal{A}'$ is equivalent to the GIN test for the surrogate pair of $(\mathbf{Z},\mathbf{Y})$, where the surrogate sets can be selected from the cluster identified in the previous iterations.
\end{Proposition}

The above proposition ensures that one can test the GIN condition over latent variables by testing its surrogate pair in each iteration, without recovering the distribution of latent variables. The complete update procedure based on Proposition \ref{Proposition-update-active-data} is summarized in Algorithm \ref{algorithm-update-activedata}, and an illustrative example is given below. 

\begin{algorithm}[htb]
	\caption{UpdateActiveData}
	\label{algorithm-update-activedata}
		\begin{algorithmic}[1]
        \REQUIRE
	A set of observed variables $\mathbf{X}$, current active variable set $\mathcal{A}$, and partial graph $\mathcal{G}$\\
		\IF{No new latent set in $\mathcal{G}$}
		\STATE $\mathcal{A} = \emptyset$;
		\ELSE
			\FOR {each new latent set $\mathcal{L}_{i} \in \mathcal{G}$}
			\STATE $\mathbf{P} \leftarrow$ Find a subset of $\mathbf{De}_{\mathbf{O}}(\mathcal{L}_{i})$ such that $|\mathbf{P}|=|\mathcal{L}_{i}|$ and that as many pure descendants as possible are added;
			\STATE Initialize each latent variable in $\mathcal{L}_{i}$ with the value of each variable of $\mathbf{P}$;
			\STATE Add $\mathcal{L}_{i}$ into $\mathcal{A}$ and delete $\mathbf{Ch}(\mathcal{L}_{i})$ from $\mathcal{A}$;
			\ENDFOR
		\ENDIF
	\ENSURE Updated active variable set $\mathcal{A}$
		\end{algorithmic}
\end{algorithm}
\begin{Example-set}
Continue to consider the structure in Figure \ref{fig:simple-main-example}. We have determined that there are four latent variable sets, including $L_7$, $L_8$, $L_9$ and $\{L_5,L_6\}$, so we update the active data set $\mathcal{A}= \{\mathbf{X} \cup \{L_5,...,L_9\} \backslash X_{1:11}\}=\{X_{12},L_5,...,L_9,X_{13}\}$.
\end{Example-set}

\subsubsection{Phase II: Inferring Causal Structure among Latent Variables}\label{subsection-phase-II}
With phase I, we can identify latent variables, as well as the causal structure among the latent parents of pure clusters (see Lemma \ref{Lemma-Phase1-identifiable} in Section \ref{Subsection-summary-identification-correctness}). 
In this section, we show how to further identify the causal structure among latent variable sets within an impure cluster, so that the latent hierarchical causal structure is fully identifiable. The basic idea is to first identify the causal order among latent variables and then remove redundant edges. 

Below, we first show how to identify the causal order between any two latent variables. According to Theorem \ref{Theorem-causal-direction}, one can directly infer the causal order between two latent variable sets by appropriately testing for GIN conditions when their latent confounders are given. The details are given in the following proposition.

\begin{Proposition}[Identifying Causal Order between Latent Variables]\label{Proposition-causa-direction-behind-confounders}
Let ${\mathcal{L}}_p$ and ${\mathcal{L}}_q$ be two latent variables of interest in an impure cluster. Suppose $\{\mathbf{P}_1,\mathbf{P}_2\}$ and $\{\mathbf{Q}_1,\mathbf{Q}_2\}$ are the subsets of pure children of $\mathcal{L}_p$ and $\mathcal{L}_q$ with $|\mathbf{P}_1|=|\mathbf{P}_2|=|\mathcal{L}_{p}|$ and $|\mathbf{Q}_1|=|\mathbf{Q}_2|=|\mathcal{L}_{q}|$, respectively. Further suppose $\mathcal{L}_t$ is the set of latent confounders of ${\mathcal{L}}_p$ and ${\mathcal{L}}_q$.
Let $\mathbf{T}_1$ and $\mathbf{T}_2$ be two sets that contain $|\mathcal{L}'_t|$ pure children of each latent variable set $\mathcal{L}'_t$ in $\mathcal{L}_t$, and $\mathbf{T}_1 \cap \mathbf{T}_2 =\emptyset$. Then if $(\{\mathbf{P}_2,\mathbf{T}_2\},\{\mathbf{P}_1,\mathbf{Q}_1,\mathbf{T}_1\})$ follows the GIN condition, $\mathcal{L}_p$ is causally earlier than $\mathcal{L}_q$ (denoted by $\mathcal{L}_p \succ \mathcal{L}_q$).
\end{Proposition}

\begin{figure}[htp]
	\begin{center}
		\begin{tikzpicture}[scale=1.5, line width=0.5pt, inner sep=0.2mm, shorten >=.1pt, shorten <=.1pt]
		\draw [fill=blue!50,thick, fill opacity=0.5, draw=none] (2.0,0.8) ellipse [x radius=1.8cm, y radius=0.35cm];
		\draw (2.0, 1.6) node(L1) [circle, fill=gray!60,draw] {{\footnotesize\,$L_1$\,}};
		\draw (0.5, 0.8) node(L2) [] {{\footnotesize\,$L_{2}$\,}};
		\draw (2.0, 0.8) node(L3) [] {{\footnotesize\,$L_3$\,}};
		\draw (3.5, 0.8) node(L4) [] {{\footnotesize\,$L_4$\,}};
		\draw (4.0, 1.1) node(X7) [] {{\footnotesize\,$X_7$\,}};
		\draw (4.5, 1.4) node(X8) [] {{\footnotesize\,$X_8$\,}};
		\draw (0, 0.0) node(X1) [] {{\footnotesize\,$X_1$\,}};
		\draw (1, 0.0) node(X2) [] {{\footnotesize\,$X_2$\,}};
		\draw (1.5, 0.0) node(X3) [] {{\footnotesize\,$X_3$\,}};
		\draw (2.5, 0.0) node(X4) [] {{\footnotesize\,$X_4$\,}};
		\draw (3.0, 0.0) node(X5) [] {{\footnotesize\,$X_5$\,}};
		\draw (4.0, 0.0) node(X6) [] {{\footnotesize\,$X_6$\,}};
		\draw[-arcsq] (L1) -- (L2) node[pos=0.5,sloped,above] {};
		\draw[-arcsq] (L1) -- (L3) node[pos=0.5,sloped,above] {};
		\draw[-arcsq] (L1) -- (L4) node[pos=0.5,sloped,above] {};
		\draw[-arcsq] (L1) -- (X7) node[pos=0.5,sloped,above] {};
		\draw[-arcsq] (L1) -- (X8) node[pos=0.5,sloped,above] {};
		\draw[-arcsq] (L2) -- (X1) node[pos=0.5,sloped,above] {};
		\draw[-arcsq] (L2) -- (X2) node[pos=0.5,sloped,above] {};
		\draw[-arcsq] (L3) -- (X3) node[pos=0.5,sloped,above] {};
		\draw[-arcsq] (L3) -- (X4) node[pos=0.5,sloped,above] {};
		\draw[-arcsq] (L4) -- (X5) node[pos=0.5,sloped,above] {};
		\draw[-arcsq] (L4) -- (X6) node[pos=0.5,sloped,above] {};
		\draw[-arcsq,color=red, thick] (L2) -- (L3) node[pos=0.5,sloped,above] {};
		\draw[-arcsq,color=red, thick] (L3) -- (L4) node[pos=0.5,sloped,above] {};
		
		\draw (1.5, 0.9) node(R21A) [] {{\,}};
		\draw (0.7, 1.4) node(R21B) [] {{\footnotesize\,$\mathbf{C}_{t}$\,}};
        \draw[-arcsq, color=blue!30, thick] (R21B) to [in = 150, out = -80] (R21A);
		\end{tikzpicture}
		\caption{An illustrative example of Proposition \ref{Proposition-causa-direction-behind-confounders}.}
            \vspace{-4mm}
		\label{fig:merge-rules-differentsize-cases} 
	\end{center}
\end{figure}
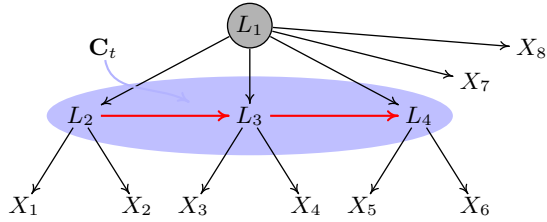

\begin{Example-set}
Consider the causal graphs in Figure \ref{fig:merge-rules-differentsize-cases}, where $\mathbf{C}_t=\{L_2,L_3,L_4\}$ is am impure cluster. Suppose $\mathcal{L}_p=L_2$ and $\mathcal{L}_q=L_3$. Then the set of latent confounders $\mathcal{L}_t=\{L_1\}$. Let $\mathbf{T}_1=\{X_7\}$, $\mathbf{T}_2=\{X_8\}$, $\{\mathbf{P}_1,\mathbf{P}_2\}=\{X_1,X_2\}$ and $\{\mathbf{Q}_1,\mathbf{Q}_2\}=\{X_3,X_4\}$.
According to Proposition \ref{Proposition-causa-direction-behind-confounders}, $(\{X_2,X_8\},\{X_1,X_3,X_7\})$ follows the GIN condition. This implies that $L_2 \succ L_3$.

\end{Example-set}

We next show how to use Proposition \ref{Proposition-causa-direction-behind-confounders} to learn the causal order between any pair of latent variable sets within an impure cluster, in a recursive way. Before that, we first introduce \textit{local root latent variable set}, which will be used in the learning procedure.

\begin{Definition}[Local Root Latent Variable Set] Let $\mathbf{C}_i=\{\mathcal{L}_1,...,\mathcal{L}_p\}$ be an impure cluster. We say  $\mathcal{L}_r \in \mathbf{C}_i$ is a local root variable set if there is no other latent variable set in $\mathbf{C}_i$ that causes it.
\end{Definition}

For any impure cluster $\mathbf{C}_i$, due to the acyclic assumption, given the identified latent parent set $\mathcal{L}_t$ and the identified pure children of any latent variable set in $\mathbf{C}_i$ from the previous step, there always exists a local root variable set $\mathcal{L}_r$ in $\mathbf{C}_i$ and it can be found with Proposition \ref{Proposition-causa-direction-behind-confounders} when $\mathcal{L}_t=\{\mathcal{L}_t\}$. After identifying the local root variable $\mathcal{L}_r$ in $\mathbf{C}_i$, we remove $\mathcal{L}_r$ from $\mathbf{C}_i$ and add it into the latent confounder set $\mathcal{L}_t$. 
We repeat the above procedure and recursively discover the local root variable until the causal order of the latent variables within the impure cluster is fully determined. The detailed procedure of identifying causal orders is given in Lines 2-8 of Algorithm \ref{algorithm-locally-infer-causalorder}.

\begin{Example-set}
	Continue to consider the example in Figure~\ref{fig:merge-rules-differentsize-cases}. We have known that $L_2$ is a local root variable. Now, we update $\mathbf{C}_i=\{L_3,L_4\}$ and the latent confounder $\mathbf{LC}=\{L_1,L_2\}$. According to
	Proposition \ref{Proposition-causa-direction-behind-confounders}, we further obtain that $L_3$ is the local root variable. Thus, we return the causal order: $L_2 \succ L_3 \succ L_4$.
\end{Example-set}

\begin{algorithm}[htb]
	\caption{LocallyInferCausalStructure}
	\label{algorithm-locally-infer-causalorder}
	\begin{algorithmic}[1]
	\REQUIRE
	A set of observed variables $\mathbf{X}$ and partial structure $\mathcal{G}$\\
	\REPEAT
			\STATE Select an impure cluster $\mathbf{C}_i$ from $\mathcal{G}$;
			\STATE Initialize latent confounder set: $\mathbf{LC}= \emptyset$;
			\STATE Add the common parent set $\mathcal{L}_t$ of $\mathbf{C}_i$ into $\mathbf{LC}$;
			\WHILE{$|\mathbf{C}_i| > 1$}
			\STATE Find a local root variable $\mathcal{L}_{r}$ according to Proposition \ref{Proposition-causa-direction-behind-confounders};
			\STATE $\mathbf{C}_i=\mathbf{C}_i \backslash \mathcal{L}_r$ and add $\mathcal{L}_r$ into $\mathbf{LC}$;
			\STATE ${\mathcal{G}}={\mathcal{G}} \cup \{\mathcal{L}_{r} \succ \mathcal{L}_{i} | \mathcal{L}_{i} \in \mathbf{C}_i\}$;
			\ENDWHILE
			\REPEAT
			\STATE Select an ordered pair of variables $\mathcal{L}_p$ and $\mathcal{L}_q$ in $\mathbf{C}_i$ with $\mathcal{L}_p \succ \mathcal{L}_q$;
			\IF{there exists set $\mathcal{L}_\mathbf{S} \subset \mathbf{C}_1$ such that each latent variable set is causally later than $\mathcal{L}_p$ and is causally earlier than $\mathcal{L}_q$, and the conditions in Proposition \ref{Proposition-removing-redundant-edges} hold.}
			\STATE Remove the directed edge between $\mathcal{L}_p$ and $\mathcal{L}_q$.
			\ENDIF
			\UNTIL{All ordered pairs of variables in $\mathbf{C}_i$ selected}
			\UNTIL{All impure clusters in $\mathcal{G}$ selected}
	\ENSURE
	Fully identified structure $\mathcal{G}$
	\end{algorithmic}
\end{algorithm}

After identifying the causal order over a set of latent variables within an impure cluster, we extend the rank-deficiency test stated in \cite{Silva-linearlvModel} to remove redundant edges between latent variable sets. 
\begin{Proposition}[Removing Redundant Edges]\label{Proposition-removing-redundant-edges} Let $\mathcal{L}_p$ and $\mathcal{L}_q$ be two latent variable sets in an impure cluster $\mathbf{C}_i$, and denote by $\mathbf{P}_1$ and $\mathbf{Q}_1$ the pure children sets of $\mathcal{L}_p$ and $\mathcal{L}_q$, respectively, with $|\mathbf{P}_1|=|\mathcal{L}_p|$ and $|\mathbf{Q}_1|=|\mathcal{L}_q|$. 
Suppose $\mathcal{L}_p$ is causally earlier than $\mathcal{L}_q$.
Let $\mathcal{L}_t$ be the common parents of $\mathbf{C}_i$ and $\mathcal{L}_\mathbf{S}=\{\mathcal{L}_{\mathbf{S}_1},...,\mathcal{L}_{\mathbf{S}_s}\}$ be the set of latent variable sets in $\mathbf{C}_i$ such that each latent variable set $\mathcal{L}_{\mathbf{S}_i}$ is causally later than $\mathcal{L}_p$ and is causally earlier than $\mathcal{L}_q$. Furthermore, let $\{\mathbf{T}_1,\mathbf{T}_2\}$ be pure children of $\mathcal{L}_t$ with $|\mathbf{T}_1|=|\mathbf{T}_2|=|\mathcal{L}_t|$, and $\mathbf{S}$ be a set that contains  $|\mathcal{L}_{\mathbf{S}_i}|$ pure children 
of each latent variable set $\mathcal{L}_{\mathbf{S}_i} \subset \mathcal{L}_{\mathbf{S}}$.
Then $\mathcal{L}_p$ and $\mathcal{L}_q$ are d-separated by $\mathcal{L}_\mathbf{S}\cup \mathcal{L}_t$, i.e., there is no directed edge between $\mathcal{L}_p$ and $\mathcal{L}_q$ iff the rank of the cross-covariance matrix of $\{\mathbf{P}_1,\mathbf{Q}_1\} \cup \{\mathbf{T}_1,\mathbf{T}_2\} \cup \mathbf{S}$ is less than or equal to $|\mathcal{L}_t \cup \mathcal{L}_\mathbf{S}|$.
\end{Proposition}
Proposition \ref{Proposition-removing-redundant-edges} helps us to identify whether there is a directed edge between two latent variable sets by searching the d-separation set from other latent variables in sequence. 
The detailed procedure for removing the redundant edges is given in Lines 10-15 of Algorithm  \ref{algorithm-locally-infer-causalorder}.
\begin{Example-set}
	Continue to consider the example in Figure~\ref{fig:merge-rules-differentsize-cases}. Now, we are going to verify the directed edge between $L_2$ and $L_4$. 
	According to Proposition \ref{Proposition-removing-redundant-edges}, we obtain that the rank of the cross-covariance matrix of $\{X_1,X_5\} \cup \{X_3,X_4,X_7,X_8\}$ is less than or equal to $|\{L_1,L_3\}|=2$ (the d-separation set is $\{L_1,L_3\}$). This implies that $L_2$ and $L_4$ are d-separated by $\{L_1,L_3\}$, and we will remove the directed edge between $L_2$ and $L_4$.
\end{Example-set}
The complete learning procedure for identifying the causal structure among latent variables, based on Proposition \ref{Proposition-causa-direction-behind-confounders} and Proposition \ref{Proposition-removing-redundant-edges}, is summarized in Algorithm \ref{algorithm-locally-infer-causalorder}.

\subsubsection{Summary of Identification of LiNGLaH}\label{Subsection-summary-identification-correctness}

In this section, we show that the LaHiCaSl algorithm (Algorithm \ref{algorithm-framework}) can identify the correct causal structure asymptotically, if the data satisfies LiNGLaH and the graph structure satisfies the minimal latent hierarchical structure (Condition 1).

Below, we first show that the latent variables, as well as the causal structure among the latent parents of pure clusters, are identifiable by Step 1 of the LaHiCaSl algorithm, which is stated in the following lemma.

\begin{Lemma}\label{Lemma-Phase1-identifiable}
Suppose that the input data $\mathbf{X}$ follow LiNGLaH with the minimal latent hierarchical structure. Then the underlying latent variables, as well as the causal structure among the latent parents of pure clusters, are identifiable by Step 1 of the LaHiCaSl algorithm.   
\end{Lemma}
Lemma \ref{Lemma-Phase1-identifiable} implies that if the underlying causal structure is a \emph{general} tree-based graph (each latent variable set only has pure children), then the underlying graph can be recovered with Step 1 of the LaHiCaSl algorithm alone.

We next show that the whole hierarchical structure is (mostly) identifiable with the LaHiCaSl algorithm, as stated in Theorem \ref{Theorem-Model-Identification}. An illustrative example of the entire procedure of LaHiCaSl is given in the next section \ref{Sub-Section-Illustration-Algo}.
\begin{Theorem}[Identifiability of Latent Hierarchical Structure]\label{Theorem-Model-Identification}
Suppose that the input data $\mathbf{X}$ follows LiNGLaH with the minimal latent hierarchical structure. Then the underlying causal graph $\mathcal{G}$ is (mostly) identifiable with the LaHiCaSl algorithm, including the causal relationships between the observed variables and their corresponding latent variable sets, and the causal relationships between the latent variable sets.
However, the causal ordering among the variables within the same latent variable set is unidentifiable.\footnote{For example, the causal order between $L_5$ and $L_6$ is unidentifiable in Figure \ref{fig:simple-main-example}.}
\end{Theorem}

\subsection{Illustration of the LaHiCaSl Algorithm}\label{Sub-Section-Illustration-Algo}

\begin{figure}[htp]
	\centering
		\includegraphics[width=1.0\textwidth, height=0.50\textwidth]{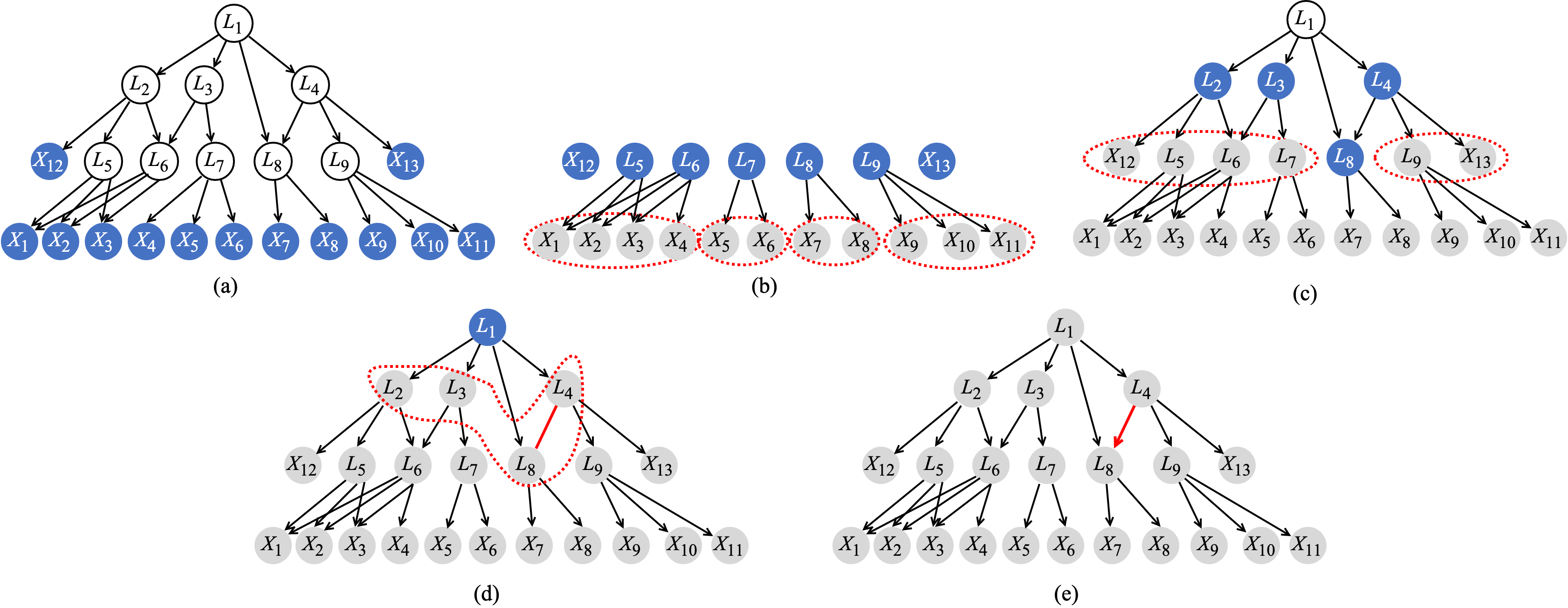}
		\vspace{-5mm}
	\caption{Illustration of the entire procedure of LaHiCaSl, where solid blue nodes indicate the active set $\mathcal{A}$ in each iteration. (a) Ground-truth latent structure. (b) Output after the first iteration of Phase I of LaHiCaSl. Red circles indicate the selected clusters. (c) Output after the second iteration of Phase I of LaHiCaSl. (d) Output after the third iteration of Phase I of LaHiCaSl. (e) Output after Phase II of LaHiCaSl, which is the same as the ground-truth causal structure.} 
        \vspace{-8mm}
	\label{fig:illustrate-algorithm}
\end{figure}

In this section, by assuming oracle tests for GIN conditions, we illustrate our LaHiCaSl algorithm with the ground-truth graph given in Figure \ref{fig:illustrate-algorithm}(a). In this structure, the variables $L_i$ ($i=1,...,9$) are unobserved and $X_j$ ($j=1,...,13$) are observed. The estimating process is as follows:
~\\

\noindent\textbf{Performing Phase I: locate latent variables}
\begin{enumerate}[itemsep=0.2pt,topsep=0.2pt]
    \item[1.1.] LaHiCaSl first initializes active variable set $\mathcal{A}=X_{1:13}$ and graph $\mathcal{G}=\emptyset$.
    \item[1.2.] It runs the first iteration of Phase I. Specifically,
    It runs \emph{IdentifyGlobalCausalClusters} (I-S1) and outputs nine clusters, i.e., $\mathbf{C}_1=\{X_5,X_6\}$, $\mathbf{C}_2=\{X_7,X_8\}$, $\mathbf{C}_3=\{X_9,X_{10}\}$, $\mathbf{C}_4=\{X_9,X_{11}\}$, $\mathbf{C}_5=\{X_{10},X_{11}\}$, $\mathbf{C}_6=\{X_1,X_2,X_3\}$, $\mathbf{C}_7=\{X_1,X_2,X_4\}$, $\mathbf{C}_8=\{X_1,X_3,X_4\}$, and $\mathbf{C}_9=\{X_2,X_3,X_4\}$.
    Next, it runs \emph{DetermineLatentVariables} (I-S2) and finds that $\mathbf{C}_3$, $\mathbf{C}_4$ and $\mathbf{C}_5$ are merged, and that $\mathbf{C}_6$, $\mathbf{C}_7$, $\mathbf{C}_8$, and $\mathbf{C}_9$ are merged, by using $\mathcal{R}1$ of Proposition \ref{Proposition-merge-rules-cases}. Thus, it obtains four clusters and introduces four latent variable sets for them: $\mathcal{L}(X_{1:4}) \coloneqq \{L_5, L_6\}$, $\mathcal{L}(\{X_5, X_6\}) \coloneqq L_7$, $\mathcal{L}(\{X_7, X_8\}) \coloneqq L_8$, and $\mathcal{L}(\{X_{9:11}\}) \coloneqq L_9$. Further, it runs \emph{UpdateActiveData} (I-S3) and updates the active variable set $\mathcal{A}= \{\mathbf{X} \cup \{L_5,...,L_9\} \backslash X_{1:11}\}=\{X_{12},L_5,...,L_9,X_{13}\}$. The outputs are shown in Figure \ref{fig:illustrate-algorithm}(b). 
    \item[1.3.] It runs the second iteration of Phase I. Specifically, it runs \emph{IdentifyGlobalCausalClusters} and finds five clusters, i.e., $\mathbf{C}_1=\{X_{12}, L_5, L_6\}$, $\mathbf{C}_2=\{X_{12}, L_5, L_7\}$, $\mathbf{C}_3=\{X_{12}, L_6, L_7\}$, $\mathbf{C}_4=\{L_5, L_6, L_7\}$, and $\mathbf{C}_5=\{L_9, X_{13}\}$. Next, it runs \emph{DetermineLatentVariables} and and merges $\mathbf{C}_1$, $\mathbf{C}_2$, $\mathbf{C}_3$ and $\mathbf{C}_4$ into one cluster by using $\mathcal{R}1$ of Proposition \ref{Proposition-merge-rules-cases}. It obtains two clusters and introduces two latent variable sets for them: $\mathcal{L}(\{X_{12}, L_5, L_6, L_7\}) \coloneqq \{L_2, L_3\}$, and $\mathcal{L}(\{L_9, X_{13}\}) \coloneqq \{L_4\}$. Further, it runs \emph{UpdateActiveData} and updates the active variable set $\mathcal{A}= \{L_2,L_3,L_8,L_4\}$. The outputs are shown in Figure \ref{fig:illustrate-algorithm}(c). 
    \item[1.4.] Analogously, in the third iteration, it runs \emph{DetermineLatentVariables} and identifies two clusters $\{L_2, L_3\}$ and$\{L_4, L_8\}$. Next, it runs \emph{DetermineLatentVariables} and finds that the introduced $L_1$ is the parent of $\{L_2,L_3\}$ and$\{L_4,L_8\}$.
    Further, it runs \emph{UpdateActiveData} and updates the active variable set $\mathcal{A}= \{L_1\}$.
    The outputs are shown in Figure \ref{fig:illustrate-algorithm}(d). 
    \item[1.5.] Since $|\mathcal{A}|<3$, there is no newly introduced latent variable. Thus, Phase I of LaHiCaSl stops.
\end{enumerate}

\noindent\textbf{Performing Phase II: infer causal structure among latent variables}

\begin{enumerate}[itemsep=0.2pt,topsep=0.2pt]
    \item[2.1.] LaHiCaSl performs Phase II as follows: for impure cluster $\{L_4, L_8\}$, it runs \emph{LocallyInferCausalStructure} and finds that $\{L_4\}$ is a local root set, i.e., $L_4 \succ L_5$ and there exists the directed edge between $L_4$ and $L_5$.
    \item[2.2.] Since there is no impure cluster, Phase II of the LaHiCaSl algorithm stops. The unknown latent structure is fully reconstructed, as given in Figure \ref{fig:illustrate-algorithm}(e).
\end{enumerate}

\subsection{Practical Implementation of LaHiCaSl Algorithm with Finite Data}\label{Sub-Section-Practical-Algorithm}
In this section, we give practical implementation details of the LaHiCaSl algorithm. We found that the originally proposed algorithm may not perform well in practical tests with limited sample sizes.
The main reasons are given below.

\begin{itemize}[itemsep=0.2pt,topsep=0.2pt]
    \item  For any two sets of variables $\mathbf{Y}$ and $\mathbf{Z}$, to test the GIN condition, we need to test the independence between $E_{\mathbf{Y}||\mathbf{Z}}$ and $\mathbf{Z}$. Such independence tests usually get less accurate with the dimensionality of $\mathbf{Z}$.
    \item The independent tests used in the GIN condition highly rely on higher-order statistics. If the variables are Gaussian, then the GIN condition cannot be used to identify the latent causal structure, since the independence holds all the time (see Proposition \ref{observation-GIN-Cov-equal-zero} below). However, reliable estimation of higher-order statistics requires much more samples than that of second-order statistics~\citep{hyvarinen2004independent}. 
\end{itemize}

\begin{Proposition}\label{observation-GIN-Cov-equal-zero}
Let $\mathbf{Y}$ and $\mathbf{Z}$ be two observed random vectors. Suppose the variables follow a linear Gaussian acyclic causal model. Then  $E_{\mathbf{Y}||\mathbf{Z}}$ is always statistically independent of $\mathbf{Z}$, i.e., $(\mathbf{Z},\mathbf{Y})$ always follows GIN condition.
\end{Proposition}

To mitigate the first issue, we check for the pairwise independence with Fisher’s method \citep{fisher1992statistical} instead of testing for the independence between $E_{\mathbf{Y}||\mathbf{Z}}$ and $\mathbf{Z}$ directly. In particular, denote by $p_{k}$, with $k=1,2,...,|\mathbf{Z}|$, all resulting $p$-values from pairwise independence between variables using the Hilbert-Schmidt Independence Criterion (HSIC)-based test ~\citep{zhang2018large}. The test statistic is $-2\sum_{k=1}^{|\mathbf{Z}|}\log{p_k}$, which follows a chi-square distribution with $2|\mathbf{Z}|$ degrees of freedom when all the pairs are independent. The complete procedure is given in Algorithm \ref{algorithm-test-GIN}.

\begin{algorithm}[htb]
	\caption{TestGIN}
	\label{algorithm-test-GIN}
	\begin{algorithmic}[1]
        \REQUIRE
    Variable sets $\mathbf{Z}$, $\mathbf{Y}$ and significance  level  $\alpha$ \\
	       \STATE Initialize test statistic ${sta}=0$;
            \STATE $\omega \leftarrow \omega^\intercal \mathbb{E}[\mathbf{Y}\mathbf{Z}^\intercal] = \mathbf{0}$ and $\omega \neq \mathbf{0}$;
            \STATE $E_{\mathbf{Y}||\mathbf{Z}} = \omega^\intercal \mathbf{Y}$;
            \FOR{each variable $Z_k$ in $\mathbf{Z}$}
            \STATE $p_k \leftarrow \mathrm{HSIC}(E_{\mathbf{Y}||\mathbf{Z}},Z_{k})$;
            \STATE ${sta} = {sta} + \mathrm{log}(p_k)$;
            \ENDFOR
            \STATE ${sta} = -2 * {sta}$;
	        \IF {At the significance  level  $\alpha$, the test statistic ${sta}$ follows the chi-square distribution with $2|\mathbf{Z}|$ degrees of freedom}
	        \STATE Boolean=True;
	        \ELSE
	        \STATE Boolean=False;
	        \ENDIF
        \ENSURE Boolean (that denotes whether to accept the null distribution or not) 
	\end{algorithmic}
\end{algorithm}

To mitigate the second issue, we try to explore whether second-order statistics can be complementarily used in the LaHiCaSl algorithm to help determine some structural information. Interestingly, we find that one can check rank constraints that use second-order statistics, instead of testing GIN conditions, to achieve the goal of Proposition \ref{proposition-identify-global-cluster} when $\mathbf{Y}$ and $\mathbf{Z}$ are two disjoint sets with $|\mathbf{Y}| \leq |\mathbf{Z}|$.
The theoretical guarantee is given in the following proposition.

\begin{Proposition}\label{Proposition-GIN-equal-Rank}
Let $\mathcal{G}$ be a DAG associated with a LiNGLaH. Let $\mathcal{A}$ be an active variable set in $\mathcal{G}$ and $\mathbf{Y}$ be a proper subset of $\mathcal{A}$. Suppose that rank-faithfulness and Condition 1 hold. Then $\mathbf{Y}$ is a global causal cluster with $|L(\mathbf{Y})|=\mathrm{Len}$ if and only if for any subset $\tilde{\mathbf{Y}}$ of $\mathbf{Y}$ with $|\tilde{\mathbf{Y}}|=\mathrm{Len}+1$, the following two conditions hold: (1) the cross-covariance matrix $(\boldsymbol{\Sigma}_{\{\mathcal{A} \backslash {\mathbf{Y}}\},\tilde{\mathbf{Y}}})$  has rank $|\tilde{\mathbf{Y}}|-1$, and (2) there is no subset $\tilde{\mathbf{Y}'} \subseteq \tilde{\mathbf{Y}}$ such that the cross-covariance matrix $(\boldsymbol{\Sigma}_{\{\mathcal{A} \backslash \tilde{\mathbf{Y}'}\},\tilde{\mathbf{Y}'}})$ has rank $|\tilde{\mathbf{Y}'}|-1$.
\end{Proposition}

Thus, in practical implementations, we combine Proposition \ref{proposition-identify-global-cluster} and \ref{Proposition-GIN-equal-Rank} to identify the global causal clusters, which can achieve the same goal as Algorithm \ref{algorithm-find-causal-cluster} but is statistically more efficient, with the complete procedure given in Algorithm \ref{algorithm-find-causal-cluster-practical}.

\begin{algorithm}[htb]
	\caption{IdentifyGlobalCausalClusters$^{+}$}
	\label{algorithm-find-causal-cluster-practical}
		\begin{algorithmic}[1]
        \REQUIRE An active variable set $\mathcal{A}$\\
			\STATE Initialize a cluster set $\mathrm{ClusterList}= \emptyset$ and the group size $\mathrm{GrLen}=2$;
			\WHILE{$|\mathcal{A}| \geq 2\times \mathrm{GrLen} - 1$}
            \REPEAT
            \STATE Select a subset $\mathbf{Y}$ from $\mathcal{A}$ such that $|\mathbf{Y}|=\mathrm{GrLen}$;
			\FOR {$\mathrm{LaLen}=1:\mathrm{GrLen}-1$}
			\IF {$|\tilde{\mathbf{Y}}| \leq |\mathcal{A} \backslash {\mathbf{Y}}| $ and $\boldsymbol{\Sigma}_{\{\mathcal{A} \backslash {\mathbf{Y}}\},\tilde{\mathbf{Y}}}$  has rank $|\tilde{\mathbf{Y}}|-1$ for $\forall \tilde{\mathbf{Y}} \in \mathbf{Y}$ such that $|\tilde{\mathbf{Y}}|=\mathrm{LaLen}+1$}
			\STATE $L(\mathbf{Y})={\mathrm{LaLen}}$;
			\STATE Add $\mathbf{Y}$ into $\mathrm{ClusterList}$;
			\STATE Break the for loop of line 5;
			\ENDIF
                \IF {$|\tilde{\mathbf{Y}}| > |\mathcal{A} \backslash {\mathbf{Y}}| $ and $(\mathcal{A} \backslash {\mathbf{Y}}, \tilde{\mathbf{Y}}) $ follows GIN condition for $\forall \tilde{\mathbf{Y}} \in \mathbf{Y}$ such that $|\tilde{\mathbf{Y}}|=\mathrm{LaLen}+1$}
			\STATE $L(\mathbf{Y})={\mathrm{LaLen}}$;
			\STATE Add $\mathbf{Y}$ into $\mathrm{ClusterList}$;
			\STATE Break the for loop of line 5;
			\ENDIF
			\ENDFOR
			\UNTIL{{\color{black}{all subsets with group length $\mathrm{GrLen}$ in $\mathcal{A}$ have been selected;}}}
			\STATE $\mathcal{A}=\mathcal{A} \backslash \mathrm{ClusterList}$, and $\mathrm{GrLen} \leftarrow \mathrm{GrLen}+1$;
			\ENDWHILE
	\ENSURE A cluster set $\mathrm{ClusterList}$
		\end{algorithmic}
\end{algorithm}

In addition, it is worth noting that although rank conditions help obtain some cluster information, it is not enough to identify the whole structure, such as the number of latent variables and the causal direction among the latent variables (see the example below).

\begin{figure}[htp]
	\begin{center}
        \begin{tikzpicture}[scale=1.5, line width=0.5pt, inner sep=0.2mm, shorten >=.1pt, shorten <=.1pt]
		\draw (1, 0) node(L1) [circle, fill=gray!60,draw] {{\footnotesize\,$L_1$\,}};
		\draw (0, -0.8) node(X1) [] {{\footnotesize\,${X}_{1}$\,}};
		\draw (0.6, -0.8) node(X2) [] {{\footnotesize\,$X_2$\,}};
		\draw (1.4, -0.8) node(X3) [] {{\footnotesize\,$X_3$\,}};
		\draw (2.0, -0.8) node(X4) [] {{\footnotesize\,${X}_{4}$\,}};
		\draw[-arcsq] (L1) -- (X1) node[pos=0.5,sloped,above] {};
		\draw[-arcsq] (L1) -- (X2) node[pos=0.5,sloped,above] {};
		\draw[-arcsq] (L1) -- (X3) node[pos=0.5,sloped,above] {};
		\draw[-arcsq] (L1) -- (X4) node[pos=0.5,sloped,above] {}; 
		\draw [-arcsq] (X3) edge[bend right=60] (X4);
            \draw (1.0, -1.2) node(con1) [] {{\footnotesize\,(a) \,}};
		\end{tikzpicture}~~~~~~~~~~~~~
		\begin{tikzpicture}[scale=1.5, line width=0.5pt, inner sep=0.2mm, shorten >=.1pt, shorten <=.1pt]
		\draw (0.0, 0) node(L1) [circle, fill=gray!60,draw] {{\footnotesize\,$L_1$\,}};
		\draw (1.5, 0) node(L2) [circle, fill=gray!60,draw] {{\footnotesize\,$L_2$\,}};
		\draw (-0.4, -0.8) node(X1) [] {{\footnotesize\,${X}_{1}$\,}};
		\draw (0.4, -0.8) node(X2) [] {{\footnotesize\,$X_2$\,}};
		\draw (1.1, -0.8) node(X3) [] {{\footnotesize\,$X_3$\,}};
		\draw (1.9, -0.8) node(X4) [] {{\footnotesize\,${X}_{4}$\,}};
		\draw[-arcsq] (L1) -- (X1) node[pos=0.5,sloped,above] {};
		\draw[-arcsq] (L1) -- (X2) node[pos=0.5,sloped,above] {};
		\draw[-arcsq] (L2) -- (X3) node[pos=0.5,sloped,above] {};
		\draw[-arcsq] (L2) -- (X4) node[pos=0.5,sloped,above] {}; 
		\draw[-arcsq] (L1) -- (L2)node[pos=0.5,sloped,above]{};
            \draw (0.75, -1.2) node(con1) [] {{\footnotesize\,(b) \,}};
		\end{tikzpicture}
		\vspace{-0.3cm}
		\caption{Two structures that are distinguishable by the GIN condition, but not by rank constraints, where (a) has an edge between observed variables $X_3$ and $X_4$. }
		\label{fig:discussion} 
	\end{center} 
 \vspace{-15mm}
\end{figure}
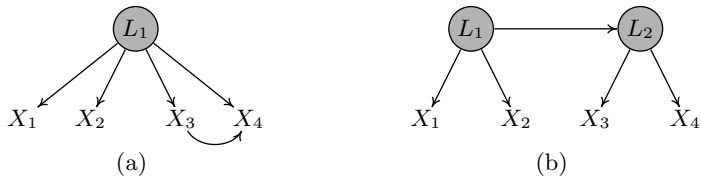

\begin{Example-set}
Consider the structures in Figure \ref{fig:discussion}. The two graphs (a) and (b) entail the same rank conditions, which means that only using rank constraints can not distinguish between the two graphs. However, they entail different GIN conditions: for (a), $(\{X_{1},X_{3}\},X_{2:4})$ satisfies the GIN condition while $(\{X_{1},X_{4}\},X_{2:4})$ violates the 
GIN condition; for (b) $(\{X_{1},X_{3}\},X_{2:4})$ and $(\{X_{1},X_{4}\},X_{2:4})$ both satisfies the GIN condition.
Based on the above analyses, the GIN condition on pairs of disjoint subsets of variables contains more causal information than the rank condition.
\end{Example-set}

\section{Experimental Results}\label{Sec-Experiment}
In this section, we show the simulation results on synthetic data to demonstrate the correctness of our proposed method. 

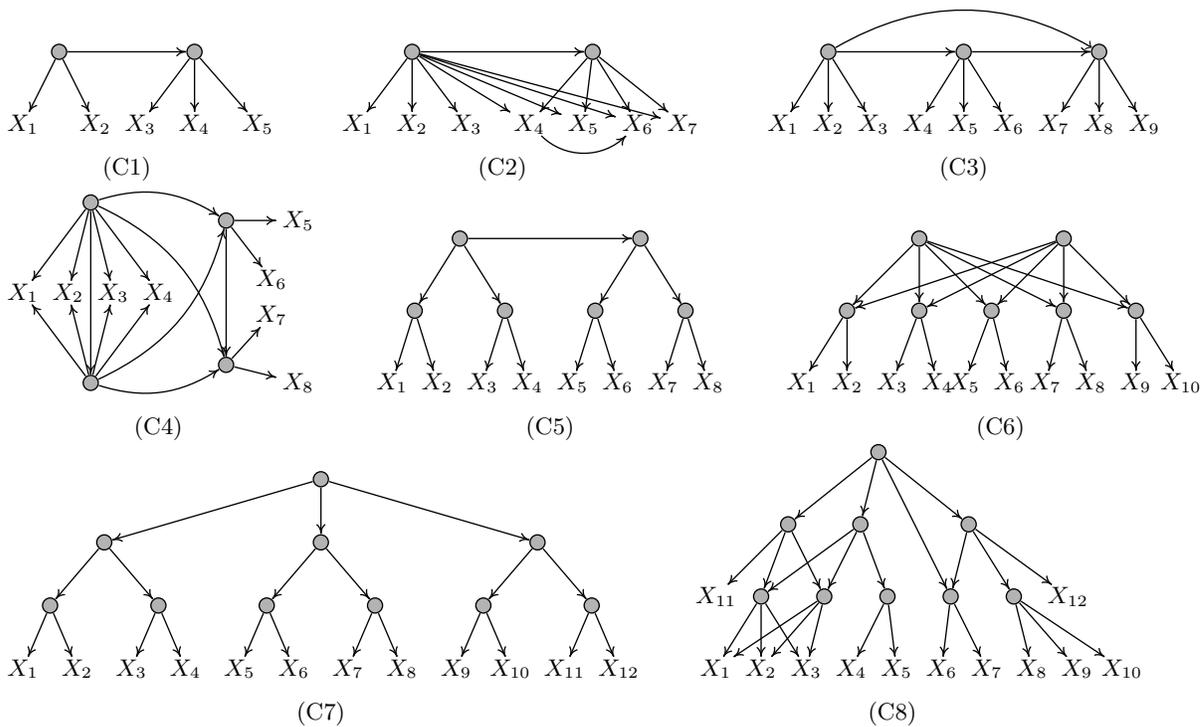
\begin{figure}[htp]
  \setlength{\abovecaptionskip}{0pt}
	\setlength{\belowcaptionskip}{-6pt}
	\vspace{-0.1cm}
	\begin{center}
    	\begin{tikzpicture}[scale=1.2, line width=0.5pt, inner sep=0.2mm, shorten >=.1pt, shorten <=.1pt]
		\draw (0, 1.6) node(L1) [circle,inner sep=0pt, minimum size=0.2cm,  fill=gray!60,draw] {};
		\draw (1.5, 1.6) node(L2) [circle,inner sep=0pt, minimum size=0.2cm, fill=gray!60,draw] {};
		\draw[-arcsq] (L1) -- (L2) node[pos=0.5,sloped,above] {};
		\draw (-0.4, 0.8) node(X1) [] {{\footnotesize\,$X_{1}$\,}};
		\draw (0.4, 0.8) node(X2) [] {{\footnotesize\,$X_2$\,}};
		\draw (0.9, 0.8) node(X3) [] {{\footnotesize\,$X_3$\,}};
		\draw (1.5, 0.8) node(X4) [] {{\footnotesize\,$X_4$\,}};
            \draw (2.2, 0.8) node(X5) [] {{\footnotesize\,$X_5$\,}};
		\draw[-arcsq] (L1) -- (X1) node[pos=0.5,sloped,above] {};
		\draw[-arcsq] (L1) -- (X2) node[pos=0.5,sloped,above] {};
		\draw[-arcsq] (L2) -- (X3) node[pos=0.5,sloped,above] {};
		\draw[-arcsq] (L2) -- (X4) node[pos=0.5,sloped,above] {};
            \draw[-arcsq] (L2) -- (X5) node[pos=0.5,sloped,above] {};
		%
            %
		\draw (0.75, 0.3) node(con3) [] {{\footnotesize\,(C1)\,}};
		\end{tikzpicture}~~~~~~~~
	    \begin{tikzpicture}[scale=1.2, line width=0.5pt, inner sep=0.2mm, shorten >=.1pt, shorten <=.1pt]
		\draw (0.0, 1.6) node(L1) [circle,inner sep=0pt, minimum size=0.2cm,  fill=gray!60,draw] {};
		\draw (2.0, 1.6) node(L2) [circle,inner sep=0pt, minimum size=0.2cm, fill=gray!60,draw] {};
		\draw[-arcsq] (L1) -- (L2) node[pos=0.5,sloped,above] {};
		\draw (-0.6, 0.8) node(X1) [] {{\footnotesize\,$X_{1}$\,}};
		\draw (0.0, 0.8) node(X2) [] {{\footnotesize\,$X_2$\,}};
		\draw (0.6, 0.8) node(X3) [] {{\footnotesize\,$X_3$\,}};
		\draw (1.3, 0.8) node(X4) [] {{\footnotesize\,$X_4$\,}};
		\draw (1.9, 0.8) node(X5) [] {{\footnotesize\,$X_5$\,}};
		\draw (2.5, 0.8) node(X6) [] {{\footnotesize\,$X_6$\,}};
            \draw (3.0, 0.8) node(X7) [] {{\footnotesize\,$X_7$\,}};
		\draw[-arcsq] (L1) -- (X1) node[pos=0.5,sloped,above] {};
		\draw[-arcsq] (L1) -- (X2) node[pos=0.5,sloped,above] {};
		\draw[-arcsq] (L1) -- (X3) node[pos=0.5,sloped,above] {};
		\draw[-arcsq] (L1) -- (X4) node[pos=0.5,sloped,above] {};
		\draw[-arcsq] (L1) -- (X5) node[pos=0.5,sloped,above] {};
		\draw[-arcsq] (L1) -- (X6) node[pos=0.5,sloped,above] {};
            \draw[-arcsq] (L1) -- (X7) node[pos=0.5,sloped,above] {};
		\draw[-arcsq] (L2) -- (X4) node[pos=0.5,sloped,above] {};
		\draw[-arcsq] (L2) -- (X5) node[pos=0.5,sloped,above] {};
		\draw[-arcsq] (L2) -- (X6) node[pos=0.5,sloped,above] {};
		\draw[-arcsq] (L2) -- (X7) node[pos=0.5,sloped,above] {};
            \draw [-arcsq] (X4) edge[bend right=45] (X6);
		\draw (1.0, 0.3) node(con3) [] {{\footnotesize\,(C2)\,}};
		\end{tikzpicture}~~~~~~~~
        \begin{tikzpicture}[scale=1.2, line width=0.5pt, inner sep=0.2mm, shorten >=.1pt, shorten <=.1pt]
		\draw (0, 1.6) node(L1) [circle,inner sep=0pt, minimum size=0.2cm,  fill=gray!60,draw] {};
		\draw (1.5, 1.6) node(L2) [circle,inner sep=0pt, minimum size=0.2cm, fill=gray!60,draw] {};
		\draw (3.0, 1.6) node(L3) [circle,inner sep=0pt, minimum size=0.2cm, fill=gray!60,draw] {};
		\draw[-arcsq] (L1) -- (L2) node[pos=0.5,sloped,above] {};
		\draw[-arcsq] (L2) -- (L3) node[pos=0.5,sloped,above] {};
		\draw [->] (L1) edge[bend right=-30] (L3);
		
		\draw (-0.5, 0.8) node(X1) [] {{\footnotesize\,$X_{1}$\,}};
		\draw (0, 0.8) node(X2) [] {{\footnotesize\,$X_2$\,}};
		\draw (0.5, 0.8) node(X3) [] {{\footnotesize\,$X_3$\,}};
		\draw (1.0, 0.8) node(X4) [] {{\footnotesize\,$X_4$\,}};
		\draw (1.5, 0.8) node(X5) [] {{\footnotesize\,$X_5$\,}};
		\draw (2.0,0.8) node(X6) [] {{\footnotesize\,$X_6$\,}};
		\draw (2.5, 0.8) node(X7) [] {{\footnotesize\,$X_{7}$\,}};
		\draw (3.0, 0.8) node(X8) [] {{\footnotesize\,$X_{8}$\,}};
		\draw (3.5, 0.8) node(X9) [] {{\footnotesize\,$X_{9}$\,}};
		%
            %
		\draw[-arcsq] (L1) -- (X1) node[pos=0.5,sloped,above] {};
		\draw[-arcsq] (L1) -- (X2) node[pos=0.5,sloped,above] {};
		\draw[-arcsq] (L1) -- (X3) node[pos=0.5,sloped,above] {};
		\draw[-arcsq] (L2) -- (X4) node[pos=0.5,sloped,above] {};
		\draw[-arcsq] (L2) -- (X5) node[pos=0.5,sloped,above] {};
		\draw[-arcsq] (L2) -- (X6) node[pos=0.5,sloped,above] {};
		\draw[-arcsq] (L3) -- (X7) node[pos=0.5,sloped,above] {};
		\draw[-arcsq] (L3) -- (X8) node[pos=0.5,sloped,above] {};
		\draw[-arcsq] (L3) -- (X9) node[pos=0.5,sloped,above] {};
		\draw (1.5, 0.3) node(con3) [] {{\footnotesize\,(C3)\,}};
		\end{tikzpicture}\\
		\begin{tikzpicture}[scale=1.2, line width=0.5pt, inner sep=0.2mm, shorten >=.1pt, shorten <=.1pt]
		\draw (1.5, 2.0) node(i-L1) [circle,inner sep=0pt, minimum size=0.2cm, fill=gray!60,draw] {};
		\draw (1.5, 0.0) node(i-L2) [circle,inner sep=0pt, minimum size=0.2cm, fill=gray!60,draw] {};
		\draw (3.0, 1.8) node(i-L3) [circle,inner sep=0pt, minimum size=0.2cm, fill=gray!60,draw] {};
		\draw (3.0,0.2) node(i-L4) [circle,inner sep=0pt, minimum size=0.2cm, fill=gray!60,draw] {};
		\draw (0.75, 1) node(i-X1) [] {{\footnotesize\,$X_1$\,}};
		\draw (1.25, 1) node(i-X2) [] {{\footnotesize\,$X_2$\,}};
		\draw (1.75, 1) node(i-X3) [] {{\footnotesize\,{$X_3$}\,}};
		\draw (2.25, 1) node(i-X4) [] {{\footnotesize\,{$X_4$}\,}};
		\draw (3.8, 1.8) node(i-Y1) [] {{\footnotesize\,{$X_5$}\,}};
		\draw (3.5, 1.15) node(i-Y3) [] {{\footnotesize\,{$X_6$}\,}};
		\draw (3.5, 0.75) node(i-Z1) [] {{\footnotesize\,{$X_7$}\,}};
		\draw (3.8, -0.0) node(i-Z3) [] {{\footnotesize\,{$X_{8}$}\,}};
		\draw[-arcsq] (i-L1) -- (i-X1) node[pos=0.5,sloped,above] {}; 
		\draw[-arcsq] (i-L1) -- (i-X2)node[pos=0.5,sloped,below] {}; 
		\draw[-arcsq] (i-L1) -- (i-X3) node[pos=0.5,sloped,below] {};
		\draw[-arcsq] (i-L1) -- (i-X4) node[pos=0.5,sloped,above] {};
		\draw[-arcsq] (i-L2) -- (i-X1) node[pos=0.5,sloped,below] {};
		\draw[-arcsq] (i-L2) -- (i-X2)node[pos=0.5,sloped,above] {};
		\draw[-arcsq] (i-L2) -- (i-X3) node[pos=0.5,sloped,above] {};
		\draw[-arcsq] (i-L2) -- (i-X4) node[pos=0.5,sloped,below] {};
		\draw[-arcsq] (i-L3) -- (i-Y1) node[pos=0.5,sloped,above] {};
		\draw[-arcsq] (i-L3) -- (i-Y3)node[pos=0.5,sloped,below] {}; 
		\draw[-arcsq] (i-L4) -- (i-Z1) node[pos=0.5,sloped,above] {}; 
		\draw[-arcsq] (i-L4) -- (i-Z3)node[pos=0.5,sloped,below] {};
		\draw[-arcsq] (i-L1) -- (i-L2) node[pos=0.5,sloped,above] {};
		\draw[-arcsq] (i-L3) -- (i-L4) node[pos=0.5,sloped,above] {};
		\draw [-arcsq] (i-L1) edge[bend right=-25] (i-L4);
		\draw [-arcsq] (i-L1) edge[bend right=-25] (i-L3);
		\draw [-arcsq] (i-L2) edge[bend right=25] (i-L3);
		\draw [-arcsq] (i-L2) edge[bend right=25] (i-L4);
		\draw (2.25, -0.5) node(con3) [] {{\footnotesize\,(C4)\,}};
		\end{tikzpicture}~~~~~~~
		\begin{tikzpicture}[scale=1.2, line width=0.5pt, inner sep=0.2mm, shorten >=.1pt, shorten <=.1pt]
		\draw (0.75, 2.4) node(L0) [circle,inner sep=0pt, minimum size=0.2cm,  fill=gray!60,draw] {};
		\draw (2.75, 2.4) node(L00) [circle,inner sep=0pt, minimum size=0.2cm,  fill=gray!60,draw] {};
		\draw (0.25, 1.6) node(L1) [circle,inner sep=0pt, minimum size=0.2cm,  fill=gray!60,draw] {};
		\draw (1.25, 1.6) node(L2) [circle,inner sep=0pt, minimum size=0.2cm, fill=gray!60,draw] {};
		\draw (2.25, 1.6) node(L3) [circle,inner sep=0pt, minimum size=0.2cm,  fill=gray!60,draw] {};
		\draw (3.25, 1.6) node(L4) [circle,inner sep=0pt, minimum size=0.2cm, fill=gray!60,draw] {};
		\draw[-arcsq] (L0) -- (L00) node[pos=0.5,sloped,above] {};
		\draw[-arcsq] (L0) -- (L1) node[pos=0.5,sloped,above] {};
		\draw[-arcsq] (L0) -- (L2) node[pos=0.5,sloped,above] {};
		\draw[-arcsq] (L00) -- (L3) node[pos=0.5,sloped,above] {};
		\draw[-arcsq] (L00) -- (L4) node[pos=0.5,sloped,above] {};
		\draw (0.0, 0.8) node(X1) [] {{\footnotesize\,$X_{1}$\,}};
		\draw (0.5, 0.8) node(X2) [] {{\footnotesize\,$X_2$\,}};
		\draw (1.0, 0.8) node(X3) [] {{\footnotesize\,$X_3$\,}};
		\draw (1.5, 0.8) node(X4) [] {{\footnotesize\,$X_4$\,}};
		\draw (2.0, 0.8) node(X5) [] {{\footnotesize\,$X_5$\,}};
		\draw (2.5,0.8) node(X6) [] {{\footnotesize\,$X_{6}$\,}};
		\draw (3.0,0.8) node(X7) [] {{\footnotesize\,$X_{7}$\,}};
		\draw (3.5,0.8) node(X8) [] {{\footnotesize\,$X_{8}$\,}};
		\draw[-arcsq] (L1) -- (X1) node[pos=0.5,sloped,above] {};
		\draw[-arcsq] (L1) -- (X2) node[pos=0.5,sloped,above] {};
		\draw[-arcsq] (L2) -- (X3) node[pos=0.5,sloped,above] {};
		\draw[-arcsq] (L2) -- (X4) node[pos=0.5,sloped,above] {};
		\draw[-arcsq] (L3) -- (X5) node[pos=0.5,sloped,above] {};
		\draw[-arcsq] (L3) -- (X6) node[pos=0.5,sloped,above] {};
		\draw[-arcsq] (L4) -- (X7) node[pos=0.5,sloped,above] {};
		\draw[-arcsq] (L4) -- (X8) node[pos=0.5,sloped,above] {};
		\draw (1.75, 0.3) node(con3) [] {{\footnotesize\,(C5)\,}};
		\end{tikzpicture}~~~~~~~
		\begin{tikzpicture}[scale=1.2, line width=0.5pt, inner sep=0.2mm, shorten >=.1pt, shorten <=.1pt]
		\draw (1.6, 2.4) node(L0) [circle,inner sep=0pt, minimum size=0.2cm,  fill=gray!60,draw] {};
		\draw (3.2, 2.4) node(L00) [circle,inner sep=0pt, minimum size=0.2cm,  fill=gray!60,draw] {};
		\draw (0.8, 1.6) node(L1) [circle,inner sep=0pt, minimum size=0.2cm,  fill=gray!60,draw] {};
		\draw (1.6, 1.6) node(L2) [circle,inner sep=0pt, minimum size=0.2cm, fill=gray!60,draw] {};
		\draw (2.4, 1.6) node(L3) [circle,inner sep=0pt, minimum size=0.2cm, fill=gray!60,draw] {};
		\draw (3.2, 1.6) node(L4) [circle,inner sep=0pt, minimum size=0.2cm, fill=gray!60,draw] {};
		\draw (4.0, 1.6) node(L5) [circle,inner sep=0pt, minimum size=0.2cm, fill=gray!60,draw] {};
		\draw[-arcsq] (L0) -- (L1) node[pos=0.5,sloped,above] {};
		\draw[-arcsq] (L0) -- (L2) node[pos=0.5,sloped,above] {};
		\draw[-arcsq] (L0) -- (L3) node[pos=0.5,sloped,above] {};
		\draw[-arcsq] (L0) -- (L4) node[pos=0.5,sloped,above] {};
		\draw[-arcsq] (L0) -- (L5) node[pos=0.5,sloped,above] {};
		\draw[-arcsq] (L00) -- (L1) node[pos=0.5,sloped,above] {};
		\draw[-arcsq] (L00) -- (L2) node[pos=0.5,sloped,above] {};
		\draw[-arcsq] (L00) -- (L3) node[pos=0.5,sloped,above] {};
		\draw[-arcsq] (L00) -- (L4) node[pos=0.5,sloped,above] {};
		\draw[-arcsq] (L00) -- (L5) node[pos=0.5,sloped,above] {};
		\draw (0.3, 0.8) node(X1) [] {{\footnotesize\,$X_{1}$\,}};
		\draw (0.8, 0.8) node(X2) [] {{\footnotesize\,$X_2$\,}};
		\draw (1.3, 0.8) node(X3) [] {{\footnotesize\,$X_3$\,}};
		\draw (1.8, 0.8) node(X4) [] {{\footnotesize\,$X_4$\,}};
		\draw (2.1, 0.8) node(X5) [] {{\footnotesize\,$X_5$\,}};
		\draw (2.6,0.8) node(X6) [] {{\footnotesize\,$X_{6}$\,}};
		\draw (3.0,0.8) node(X7) [] {{\footnotesize\,$X_{7}$\,}};
		\draw (3.5,0.8) node(X8) [] {{\footnotesize\,$X_{8}$\,}};
		\draw (4.0,0.8) node(X9) [] {{\footnotesize\,$X_{9}$\,}};
		\draw (4.5,0.8) node(X10) [] {{\footnotesize\,$X_{10}$\,}};
		\draw[-arcsq] (L1) -- (X1) node[pos=0.5,sloped,above] {};
		\draw[-arcsq] (L1) -- (X2) node[pos=0.5,sloped,above] {};
		\draw[-arcsq] (L2) -- (X3) node[pos=0.5,sloped,above] {};
		\draw[-arcsq] (L2) -- (X4) node[pos=0.5,sloped,above] {};
		\draw[-arcsq] (L3) -- (X5) node[pos=0.5,sloped,above] {};
		\draw[-arcsq] (L3) -- (X6) node[pos=0.5,sloped,above] {};
		\draw[-arcsq] (L4) -- (X7) node[pos=0.5,sloped,above] {};
		\draw[-arcsq] (L4) -- (X8) node[pos=0.5,sloped,above] {};
		\draw[-arcsq] (L5) -- (X9) node[pos=0.5,sloped,above] {};
		\draw[-arcsq] (L5) -- (X10) node[pos=0.5,sloped,above] {};
		\draw (2.5, 0.3) node(con3) [] {{\footnotesize\,(C6)\,}};
		\end{tikzpicture}\\
		\begin{tikzpicture}[scale=1.2, line width=0.5pt, inner sep=0.2mm, shorten >=.1pt, shorten <=.1pt]
		\draw (3.3, 2.8) node(L1) [circle,inner sep=0pt, minimum size=0.2cm,  fill=gray!60,draw] {};
		\draw (0.9, 2.1) node(L2) [circle,inner sep=0pt, minimum size=0.2cm,  fill=gray!60,draw] {};
		\draw (3.3, 2.1) node(L3) [circle,inner sep=0pt, minimum size=0.2cm,  fill=gray!60,draw] {};
		\draw (5.7, 2.1) node(L4) [circle,inner sep=0pt, minimum size=0.2cm,  fill=gray!60,draw] {};
		\draw (0.3, 1.4) node(L5) [circle,inner sep=0pt, minimum size=0.2cm, fill=gray!60,draw] {};
		\draw (1.5, 1.4) node(L6) [circle,inner sep=0pt, minimum size=0.2cm, fill=gray!60,draw] {};
		\draw (2.7, 1.4) node(L7) [circle,inner sep=0pt, minimum size=0.2cm, fill=gray!60,draw] {};
		\draw (3.9, 1.4) node(L8) [circle,inner sep=0pt, minimum size=0.2cm, fill=gray!60,draw] {};
		\draw (5.1, 1.4) node(L9) [circle,inner sep=0pt, minimum size=0.2cm, fill=gray!60,draw] {};
		\draw (6.3, 1.4) node(L10) [circle,inner sep=0pt, minimum size=0.2cm, fill=gray!60,draw] {};
		\draw (0, 0.7) node(X1) [] {{\footnotesize\,$X_{1}$\,}};
		\draw (0.6, 0.7) node(X2) [] {{\footnotesize\,$X_2$\,}};
		\draw (1.2, 0.7) node(X3) [] {{\footnotesize\,$X_3$\,}};
		\draw (1.8, 0.7) node(X4) [] {{\footnotesize\,$X_4$\,}};
		\draw (2.4, 0.7) node(X5) [] {{\footnotesize\,$X_5$\,}};
		\draw (3.0,0.7) node(X6) [] {{\footnotesize\,$X_6$\,}};
		\draw (3.6, 0.7) node(X7) [] {{\footnotesize\,$X_7$\,}};
		\draw (4.2, 0.7) node(X8) [] {{\footnotesize\,$X_8$\,}};
		\draw (4.8,0.7) node(X9) [] {{\footnotesize\,$X_9$\,}};
		\draw (5.4, 0.7) node(X10) [] {{\footnotesize\,$X_{10}$\,}};
		\draw (6.0, 0.7) node(X11) [] {{\footnotesize\,$X_{11}$\,}};
		\draw (6.6, 0.7) node(X12) [] {{\footnotesize\,$X_{12}$\,}};

		\draw[-arcsq] (L1) -- (L2) node[pos=0.5,sloped,above] {};
		\draw[-arcsq] (L1) -- (L3) node[pos=0.5,sloped,above] {};
		\draw[-arcsq] (L1) -- (L4) node[pos=0.5,sloped,above] {};
		\draw[-arcsq] (L2) -- (L5) node[pos=0.5,sloped,above] {};
		\draw[-arcsq] (L2) -- (L6) node[pos=0.5,sloped,above] {};
		\draw[-arcsq] (L3) -- (L7) node[pos=0.5,sloped,above] {};
		\draw[-arcsq] (L3) -- (L8) node[pos=0.5,sloped,above] {};
		\draw[-arcsq] (L4) -- (L9) node[pos=0.5,sloped,above] {};
		\draw[-arcsq] (L4) -- (L10) node[pos=0.5,sloped,above] {};
		\draw[-arcsq] (L5) -- (X1) node[pos=0.5,sloped,above] {};
		\draw[-arcsq] (L5) -- (X2) node[pos=0.5,sloped,above] {};
		\draw[-arcsq] (L6) -- (X3) node[pos=0.5,sloped,above] {};
		\draw[-arcsq] (L6) -- (X4) node[pos=0.5,sloped,above] {};
		\draw[-arcsq] (L7) -- (X5) node[pos=0.5,sloped,above] {};
		\draw[-arcsq] (L7) -- (X6) node[pos=0.5,sloped,above] {};
		\draw[-arcsq] (L8) -- (X7) node[pos=0.5,sloped,above] {};
		\draw[-arcsq] (L8) -- (X8) node[pos=0.5,sloped,above] {};
		\draw[-arcsq] (L9) -- (X9) node[pos=0.5,sloped,above] {};
		\draw[-arcsq] (L9) -- (X10) node[pos=0.5,sloped,above] {};
		\draw[-arcsq] (L10) -- (X11) node[pos=0.5,sloped,above] {};
		\draw[-arcsq] (L10) -- (X12) node[pos=0.5,sloped,above] {};
		\draw (3.3, 0.2) node(con3) [] {{\footnotesize\,(C7)\,}};
		\end{tikzpicture}~~~~~
		\begin{tikzpicture}[scale=1.2, line width=0.5pt, inner sep=0.2mm, shorten >=.1pt, shorten <=.1pt]
		\draw (2, 2.4) node(L1) [circle,inner sep=0pt, minimum size=0.2cm, fill=gray!60,draw] {};
		\draw (1, 1.6) node(L2) [circle,inner sep=0pt, minimum size=0.2cm, fill=gray!60,draw] {};
		\draw (1.8, 1.6) node(L3) [circle,inner sep=0pt, minimum size=0.2cm, fill=gray!60,draw] {};
		\draw (3.0, 1.6) node(L4) [circle,inner sep=0pt, minimum size=0.2cm, fill=gray!60,draw] {};
		
		\draw (0.2, 0.8) node(X11) [] {{\footnotesize\,$X_{11}$\,}};
		\draw (0.7, 0.8) node(L5) [circle,inner sep=0pt, minimum size=0.2cm, fill=gray!60,draw] {};
		\draw (1.4, 0.8) node(L6) [circle,inner sep=0pt, minimum size=0.2cm, fill=gray!60,draw] {};
		\draw (2.1, 0.8) node(L7) [circle,inner sep=0pt, minimum size=0.2cm, fill=gray!60,draw] {};
		\draw (2.8, 0.8) node(L8) [circle,inner sep=0pt, minimum size=0.2cm, fill=gray!60,draw] {};
		\draw (3.5, 0.8) node(L9) [circle,inner sep=0pt, minimum size=0.2cm, fill=gray!60,draw] {};
		\draw (4.1, 0.8) node(X12) [] {{\footnotesize\,$X_{12}$\,}};
		\draw (0.2, 0) node(X1) [] {{\footnotesize\,$X_1$\,}};
		\draw (0.7, 0) node(X2) [] {{\footnotesize\,$X_2$\,}};
		\draw (1.2, 0) node(X3) [] {{\footnotesize\,$X_3$\,}};
		\draw (1.7,0) node(X4) [] {{\footnotesize\,$X_4$\,}};
		\draw (2.2, 0) node(X5) [] {{\footnotesize\,$X_5$\,}};
		\draw (2.7, 0) node(X6) [] {{\footnotesize\,$X_6$\,}};
		\draw (3.2, 0) node(X7) [] {{\footnotesize\,$X_7$\,}};
		\draw (3.7, 0) node(X8) [] {{\footnotesize\,$X_8$\,}};
		\draw (4.2, 0) node(X9) [] {{\footnotesize\,$X_9$\,}};
		\draw (4.7, 0) node(X10) [] {{\footnotesize\,$X_{10}$\,}};
		\draw[-arcsq] (L1) -- (L2) node[pos=0.5,sloped,above] {};
		\draw[-arcsq] (L1) -- (L3) node[pos=0.5,sloped,above] {};
		\draw[-arcsq] (L1) -- (L4) node[pos=0.5,sloped,above] {};

		\draw[-arcsq] (L1) -- (L8) node[pos=0.5,sloped,above] {};
		\draw[-arcsq] (L2) -- (X11) node[pos=0.5,sloped,above] {};
		\draw[-arcsq] (L2) -- (L5) node[pos=0.5,sloped,above] {};
		\draw[-arcsq] (L2) -- (L6) node[pos=0.5,sloped,above] {};
		\draw[-arcsq] (L3) -- (L5) node[pos=0.5,sloped,above] {};
		\draw[-arcsq] (L3) -- (L6) node[pos=0.5,sloped,above] {};
		\draw[-arcsq] (L3) -- (L7) node[pos=0.5,sloped,above] {};
		\draw[-arcsq] (L4) -- (L8) node[pos=0.5,sloped,above] {};
		\draw[-arcsq] (L4) -- (L9) node[pos=0.5,sloped,above] {};
		\draw[-arcsq] (L4) -- (X12) node[pos=0.5,sloped,above] {};
		\draw[-arcsq] (L5) -- (X1) node[pos=0.5,sloped,above] {};
		\draw[-arcsq] (L5) -- (X2) node[pos=0.5,sloped,above] {};
		\draw[-arcsq] (L5) -- (X3) node[pos=0.5,sloped,above] {};
		\draw[-arcsq] (L6) -- (X1) node[pos=0.5,sloped,above] {};
		\draw[-arcsq] (L6) -- (X2) node[pos=0.5,sloped,above] {};
		\draw[-arcsq] (L6) -- (X3) node[pos=0.5,sloped,above] {};
		\draw[-arcsq] (L7) -- (X4) node[pos=0.5,sloped,above] {};
		\draw[-arcsq] (L7) -- (X5) node[pos=0.5,sloped,above] {};
		\draw[-arcsq] (L8) -- (X6) node[pos=0.5,sloped,above] {};
		\draw[-arcsq] (L8) -- (X7) node[pos=0.5,sloped,above] {};
		\draw[-arcsq] (L9) -- (X8) node[pos=0.5,sloped,above] {};
		\draw[-arcsq] (L9) -- (X9) node[pos=0.5,sloped,above] {};
		\draw[-arcsq] (L9) -- (X10) node[pos=0.5,sloped,above] {};
		\draw (2.2, -0.5) node(con1) [] {{\footnotesize\,(C8) \,}};
		\end{tikzpicture}
		\caption{The causal structures used in our simulation studies.}
		\vspace{-0.3cm}
		\label{fig:simulations-structures} 
	\end{center}
\end{figure}

\textbf{\emph{Experimental setup:}} We generated data from eight typical graph structures that satisfy minimal latent hierarchical structure, including tree-based and measurement-based structures (see Figure \ref{fig:simulations-structures}). 
We considered different sample sizes $N = 3k, 5k, 10k$. The causal strengths $b_{ij}$ were generated uniformly from $[-2,-0.5]\cup[0.5,2]$, and the non-Gaussian noise terms were generated from the square of exponential distributions.

\textbf{\emph{Comparisons:}} We compared the proposed LaHiCaSl algorithm with measurement-based methods, such as BPC~\citep{Silva-linearlvModel}, FOFC~\citep{Kummerfeld2016}, LSTC~\citep{cai2019triad}.\footnote{For BPC and FOFC algorithms, we used the implementations in the TETRAD package, which can be downloaded at \url{http://www.phil.cmu.edu/tetrad/}.} We also compared LaHiCaSl with tree-based methods, such as Chow-Liu Recursive Grouping (CLRG) and Chow-Liu Neighbor Joining (CLNJ)~\citep{choi2011learning}. 
Each experiment was repeated 50 times with randomly generated data, and the reported results were averaged.

\textbf{\emph{Metrics:}} To evaluate the accuracy of the estimated graph, we consider the following two learning tasks:
\begin{itemize}[itemsep=0.2pt,topsep=0.2pt]
    \item [T1.] Identification of the number of latent variables.
    \item [T2.] Identification of the whole structure, including causal directions. 
\end{itemize}

For a fair comparison, specifically, for measurement-based structures (cases $1\sim4$), we followed the evaluation metrics from \cite{Silva-linearlvModel,cai2019triad} to evaluate the accuracy of the estimated causal cluster. Specifically,
we used the following three metrics:
\begin{itemize}[itemsep=0.2pt,topsep=0.2pt]
    \item \emph{Latent omission}: the number of omitted latent variables divided by the total number of latent variables in the ground truth graph.
    \item \emph{Latent commission}: the number of falsely detected latent variables divided by the total number of latent variables in the ground truth graph.
    \item \emph{Mismeasurement}: the number of falsely observed variables that have at least one incorrectly measured latent divided by the number of observed variables in the ground truth graph.
\end{itemize}

Moreover, for tree-based structures (cases $5\sim8$), we modify the evaluation metrics from \cite{choi2011learning} to evaluate LiNGLaH. Specifically, we used the following two metrics:
\begin{itemize}[itemsep=0.2pt,topsep=0.2pt]
    \item \emph{Structure recovery error rate}: 
    the percentage that the proposed algorithm fails to recover the ground-truth structure. Note that this is a strict measure because even a wrong latent variable or a wrong direction results in an error.
    \item \emph{Error in the number of latent variable sets}: the average absolute difference between the number of latent variables estimated and the number of latent variables in the ground-truth structure.  
\end{itemize}

We used the \textit{correct-ordering rate} as a metric to further evaluate the estimated causal order in all cases.
\begin{itemize}
    \item \emph{Correct-ordering rate}:
    the number of correctly inferred causal ordering divided by the total number of causal ordering in the true structure.
\end{itemize}


\noindent\textbf{Cases $1\sim4$ Results}: As shown in Table \ref{tab:compare-case1-4}, our algorithm, LaHiCaSl, achieves the best performance (the lowest errors) in almost all four cases. We noticed that although the Mismeasurement of LaHiCaSl is a bit higher than LSTC in Case 4 when the sample size is small (N=3k), the Latent commission of LaHiCaSl is lower than LSTC. 
The BPC and FOFC algorithms (with distribution-free tests) do not perform well except for Case 3, which implies that the rank constraints over the covariance matrix are not enough to recover general latent structures.
Moreover, LSTC fails to recover Cases 2 and 4 because of the existence of multiple latent variables.
The above results demonstrate a clear advantage of our method over the comparisons. 

\noindent\textbf{Cases $5\sim8$ Results}: As shown in Table \ref{tab:compare-case5-8}, our algorithm, LaHiCaSl, is superior (with the lowest error) to other comparisons with both metrics in  all cases,
indicating that it can not only identify the tree-based and measurement-based structures, but also more general latent hierarchical structures (including the causal directions).  
The LSTC algorithm does not perform well, especially in cases 7 and 8, as it requires latent variables having enough observed variables as children. We also notice that CLRG and CLNJ algorithms do not perform well in case 7 though the structure is a tree. One possible reason is that these algorithms were designed for Gaussian and discrete variables only. These findings show a clear advantage of our method over the comparisons.

\begin{center}
\begin{table}[htp!]
	\small
	\center \caption{Performance of LaHiCSL, LSTC, FOFC, and BPC on learning measured-based latent structure (the lower the better).}
 \vspace{3mm}
	\label{tab:compare-case1-4}
	\resizebox{\textwidth}{!}{
	\begin{tabular}{|c|c|c|c|c|c|c|c|c|c|c|c|c|c|}
		\hline  \multicolumn{2}{|c|}{} &\multicolumn{4}{|c|}{\textbf{Latent omission}} & \multicolumn{4}{|c|}{\textbf{Latent commission}} & \multicolumn{4}{|c|}{\textbf{Mismeasurements}}\\
		\hline 
		\multicolumn{2}{|c|}{Algorithm} & LaHiCSL & LSTC & FOFC & BPC & LaHiCSL & LSTC & FOFC & BPC & LaHiCSL & LSTC & FOFC & BPC \\
		\hline 
		 & 3k & 0.03(03) & 0.18(18) & 0.82(50) & 0.44(46) 
		 & 0.00(00) & 0.16(16) & 0.03(00) & 0.00(00) 
		 & 0.02(05) & 0.15(15) & 0.44(50) & 0.50(42) \\
		\cline{2-14}
		{\emph{Case 1}} &5k & 0.00(0)  &0.08(07)& 0.79(50) & 0.48(50) 
		& 0.00(00) & 0.05(05) & 0.00(0) & 0.00(0) 
		& 0.00(00) & 0.03(03)& 0.50(50) & 0.45(50) \\
		\cline{2-14}
		& 10k & 0.00(00) & 0.00(0) & 0.83(50) & 0.49(50) 
		& 0.00(00) & 0.00(0)& 0.00(0) & 0.00(0) 
		& 0.00(00) & 0.00(0) & 0.50(50) & 0.36(50) \\
		\hline 
		 & 3k 
		 & 0.08(08) & 0.43(42) & 0.51(48) & 0.50(50) 
		 & 0.02(02) & 0.12(08) & 0.00(0) & 0.00(0)
		 & 0.08(11) & 0.06(02) & 0.00(0) & 0.00(0)  \\
		\cline{2-14}
		{\emph{Case 2}} &5k 
		& 0.02(02) & 0.42(41) & 0.50(49) & 0.50(50) 
		& 0.00(00) & 0.08(08) & 0.00(0) & 0.00(0) 
		& 0.01(02) & 0.05(03) & 0.00(0) & 0.00(0)\\
		\cline{2-14} &10k
		& 0.01(01) & 0.37(39) & 0.50(49) & 0.50(50)
		& 0.00(00) & 0.09(08) & 0.00(0) & 0.00(0)
		& 0.01(02) & 0.04(03) & 0.00(0)  & 0.00(0)\\
		\hline 
		 & 3k 
		 & 0.04(06) & 0.12(11) & 0.05(06) & 0.08(08) 
		 & 0.00(00) & 0.05(05) & 0.03(03) & 0.05(05) 
		 & 0.03(06) & 0.04(04) & 0.00(0) & 0.00(0)\\
		\cline{2-14}
		{\emph{Case 3}} &5k 
		& 0.01(01) & 0.08(07) & 0.03(02) & 0.04(02) 
		 & 0.00(00) & 0.03(3) & 0.01(01) & 0.00(0) 
		 & 0.01(01) & 0.03(3) & 0.00(0) & 0.00(0)\\
		\cline{2-14}&10k 
		& 0.00(00) & 0.02(2) & 0.02(01) & 0.02(01) 
		 & 0.00(00) & 0.00(0) & 0.01(01) & 0.01(01) 
		 & 0.00(00) & 0.00(0) & 0.00(0) & 0.00(0)\\
		\hline 
		 & 3k 
		 &0.05(10) &0.71(50) & 0.85(50) & 0.59(48)
		 &0.03(06) &0.25(49) & 0.12(20) & 0.18(40)
		 &0.07(10) &0.00(0) & 0.02(11) & 0.05(19)\\
		\cline{2-14}
		{\emph{Case 4}} &5k
	     &0.02(04) &0.71(50) & 0.89(50) & 0.60(50)
		 &0.01(02) &0.22(50) & 0.07(13) & 0.09(11)
		 &0.03(04) &0.00(0) & 0.01(04) & 0.02(04)\\
		\cline{2-14}
		&10k 
		 &0.02(04) &0.70(50) & 0.91(50) & 0.88(50)
		 &0.01(01) &0.20(50) & 0.06(14) & 0.09(10)
		 &0.01(02) &0.00(0) & 0.00(0) & 0.00(0)\\   
		\hline 
	\end{tabular}}
	\begin{tablenotes}
		\item Note: The number in parentheses indicates the number of occurrences that the current algorithm {\it cannot} correctly solve the problem.
	\end{tablenotes}
	\vspace{-5mm}
\end{table}
\end{center}

\begin{center}
\begin{table}[htp!]
	\small
	\center \caption{Performance of LaHiCSL, LSTC, CLRG, CLNJ, FOFC, and BPC on learning latent hierarchical structure (the lower the better).}
 \vspace{3mm}
	\label{tab:compare-case5-8}
	\resizebox{15cm}{!}{
	\begin{tabular}{|c|c|c|c|c|c|c|c|c|c|c|c|c|c|}
		\hline  \multicolumn{2}{|c|}{} &\multicolumn{6}{|c|}{\textbf{Structure Recovery Error Rate} $\downarrow$} & \multicolumn{6}{|c|}{\textbf{Error in Hidden Variables} $\downarrow$}\\
		\hline 
		\multicolumn{2}{|c|}{Algorithm} & LaHiCSL & LSTC & CLRG & CLNJ & FOFC & BPC & LaHiCSL & LSTC & CLRG & CLNJ & FOFC & BPC \\
		\hline 
		 & 3k & 0.14 & 1.0 & 1.0 & 1.0 & 1.0 & 1.0 
		 & 0.60 & 2.8 & 5.0 & 5.0 & 5.7 & 5.5\\
		\cline{2-14}
		{\emph{Case 5}} & 5k & 0.06 & 1.0 & 1.0 & 1.0 & 1.0 & 1.0 
		 & 0.20 & 2.2 & 5.0 & 5.0 & 5.7 & 5.6\\
		\cline{2-14}
		& 10k & 0.00 & 1.0 & 1.0 & 1.0 & 1.0 & 1.0 
		 & 0.00 & 2.0 & 5.0 & 5.0 & 5.9 & 5.9\\
		\hline 
		 & 3k & 0.26 & 1.0 & 1.0 & 1.0 & 1.0 & 1.0 
		 & 0.80 & 4.6 & 6.0 & 6.0 & 6.8 & 6.7\\
		\cline{2-14}
		{\emph{Case 6}} & 5k & 0.12 & 1.0 & 1.0 & 1.0 & 1.0 & 1.0 
		 & 0.42 & 3.2 & 6.0 & 6.0 & 6.8 & 6.8\\
		\cline{2-14} & 10k & 0.06 & 1.0 & 1.0 & 1.0 & 1.0 & 1.0 
		 & 0.18 & 3.0 & 6.0 & 6.0 & 6.9 & 6.9\\
		\hline 
		 & 3k & 0.16 & 1.0 & 1.0 & 1.0 & 1.0 & 1.0 
		 & 0.32 & 6.4 & 9.0 & 9.0 & 9.6 & 9.4\\
		\cline{2-14}
		{\emph{Case 7}} & 5k & 0.14 & 1.0 & 1.0 & 1.0 & 1.0 & 1.0 
		 & 0.20 & 5.6 & 9.0 & 9.0 & 9.7 & 9.6\\
		\cline{2-14} & 10k & 0.04 & 1.0 & 1.0 & 1.0 & 1.0 & 1.0 
		 & 0.12 & 4.5 & 9.0 & 9.0 & 9.9 & 9.9\\
		\hline 
		 & 3k & 0.44 & 1.0 & 1.0 & 1.0 & 1.0 & 1.0 
		 & 1.42 & 7.3 & 8.0 & 8.0 & 7.6 & 7.4\\
		\cline{2-14}
		{\emph{Case 8}} & 5k & 0.28 & 1.0 & 1.0 & 1.0 & 1.0 & 1.0 
		 & 0.74 & 6.5 & 8.0 & 8.0 & 7.9 & 7.7\\
		\cline{2-14}
		& 10k & 0.16 & 1.0 & 1.0 & 1.0 & 1.0 & 1.0 
		 & 0.22 & 6.3 & 8.0 & 8.0 & 7.9 & 7.8\\   
		\hline 
	\end{tabular}}
\end{table}
\end{center}

\begin{figure}[htp]
 \vspace{1mm}
	\label{fig:exp}
	\begin{center}
		\begin{tikzpicture}[scale=0.6]
		\begin{axis}[
		title=(a),
		title style={at={(0.5,-0.6)}},
		width=5.5cm,
		cycle list name=mark list,
		ybar,
		enlargelimits=0.15,
        ymin=0.6,ymax=1,
		legend style={at={(0.65,0.2)}, anchor=north,legend columns=-1},
		ylabel={Correct ordering rate},
		xlabel={Sample size},
		symbolic x coords={3k,5k,10k},
		xtick=data,
		nodes near coords,
		nodes near coords align={vertical},
		]
		\addplot [ fill=purple,postaction={pattern=north east lines}] coordinates {(3k,0.95) (5k,0.98) (10k,1)};
		\addplot [fill=green,postaction={pattern=dots}] coordinates {(3k,0.88) (5k,0.96) (10k,1)};
		\legend{LaHiCaSl,LSTC}
		\end{axis}
		\end{tikzpicture}~~
		\begin{tikzpicture}[scale=0.6]
		\begin{axis}[
		title=(b),
		title style={at={(0.5,-0.6)}},
		width=5.5cm,
		cycle list name=mark list,
		ybar,
		enlargelimits=0.15,
        ymin=0.0,ymax=1,
		legend style={at={(0.65,0.2)}, anchor=north,legend columns=-1},
		ylabel={Correct ordering rate},
		xlabel={Sample size},
		symbolic x coords={3k,5k,10k},
		xtick=data,
		nodes near coords,
		nodes near coords align={vertical},
		]
		\addplot [ fill=purple,postaction={pattern=north east lines}] coordinates {(3k,0.90) (5k,0.95) (10k,1.0)};
		\addplot [fill=green,postaction={pattern=dots}] coordinates {(3k,0.0) (5k,0.0) (10k,0.0)};
		\legend{LaHiCaSl,LSTC}
		\end{axis}
		\end{tikzpicture}
		\begin{tikzpicture}[scale=0.6]
		\begin{axis}[
		title=(c),
		title style={at={(0.5,-0.6)}},
		width=5.5cm,
		cycle list name=mark list,
		ybar,
		enlargelimits=0.15,
        ymin=0.5,ymax=1,
		legend style={at={(0.65,0.2)}, anchor=north,legend columns=-1},
		ylabel={Correct ordering rate},
		xlabel={Sample size},
		symbolic x coords={3k,5k,10k},
		xtick=data,
		nodes near coords,
		nodes near coords align={vertical},
		]
		\addplot [ fill=purple,postaction={pattern=north east lines}] coordinates {(3k,0.92) (5k,0.95) (10k,1)};
		\addplot [fill=green,postaction={pattern=dots}] coordinates {(3k,0.88) (5k,0.93) (10k,0.96)};
		\legend{LaHiCaSl,LSTC}
		\end{axis}
		\end{tikzpicture}~~
		\begin{tikzpicture}[scale=0.6]
		\begin{axis}[
		title=(d),
		title style={at={(0.5,-0.6)}},
		width=5.5cm,
		cycle list name=mark list,
		ybar,
		enlargelimits=0.15,
        ymin=0.2,ymax=1,
		legend style={at={(0.65,0.2)}, anchor=north,legend columns=-1},
		ylabel={Correct ordering rate},
		xlabel={Sample size},
		symbolic x coords={3k,5k,10k},
		xtick=data,
		nodes near coords,
		nodes near coords align={vertical},
		]
		\addplot [ fill=purple,postaction={pattern=north east lines}] coordinates {(3k,0.90) (5k,0.95) (10k,0.98)};
		\addplot [fill=green,postaction={pattern=dots}] coordinates {(3k,0.36) (5k,0.39) (10k,0.44)};
		\legend{LaHiCaSl,LSTC}
		\end{axis}
		\end{tikzpicture}\\
		\begin{tikzpicture}[scale=0.6]
		\begin{axis}[
		title=(e),
		title style={at={(0.5,-0.6)}},
		width=5.5cm,
		cycle list name=mark list,
		ybar,
		enlargelimits=0.15,
        ymin=0.0,ymax=1,
		legend style={at={(0.65,0.2)}, anchor=north,legend columns=-1},
		ylabel={Correct ordering rate},
		xlabel={Sample size},
		symbolic x coords={3k,5k,10k},
		xtick=data,
		nodes near coords,
		nodes near coords align={vertical},
		]
		\addplot [ fill=purple,postaction={pattern=north east lines}] coordinates {(3k,0.85) (5k,0.91) (10k,1.0)};
		\addplot [fill=green,postaction={pattern=dots}] coordinates {(3k,0.0) (5k,0.0) (10k,0.0)};
		\legend{LaHiCaSl,LSTC}
		\end{axis}
		\end{tikzpicture}~~
		\begin{tikzpicture}[scale=0.6]
		\begin{axis}[
		title=(f),
		title style={at={(0.5,-0.6)}},
		width=5.5cm,
		cycle list name=mark list,
		ybar,
		enlargelimits=0.15,
        ymin=0.0,ymax=1,
		legend style={at={(0.65,0.2)}, anchor=north,legend columns=-1},
		ylabel={Correct ordering rate},
		xlabel={Sample size},
		symbolic x coords={3k,5k,10k},
		xtick=data,
		nodes near coords,
		nodes near coords align={vertical},
		]
		\addplot [ fill=purple,postaction={pattern=north east lines}] coordinates {(3k,0.81) (5k,0.92) (10k,0.97)};
		\addplot [fill=green,postaction={pattern=dots}] coordinates {(3k,0.0) (5k,0.0) (10k,0.0)};
		\legend{LaHiCaSl,LSTC}
		\end{axis}
		\end{tikzpicture}~~
		\begin{tikzpicture}[scale=0.6]
		\begin{axis}[
		title=(g),
		title style={at={(0.5,-0.6)}},
		width=5.5cm,
		cycle list name=mark list,
		ybar,
		enlargelimits=0.15,
        ymin=0.0,ymax=1,
		legend style={at={(0.65,0.2)}, anchor=north,legend columns=-1},
		ylabel={Correct ordering rate},
		xlabel={Sample size},
		symbolic x coords={3k,5k,10k},
		xtick=data,
		nodes near coords,
		nodes near coords align={vertical},
		]
		\addplot [ fill=purple,postaction={pattern=north east lines}] coordinates {(3k,0.90) (5k,0.94) (10k,0.97)};
		\addplot [fill=green,postaction={pattern=dots}] coordinates {(3k,0.0) (5k,0.0) (10k,0.0)};
		\legend{LaHiCaSl,LSTC}
		\end{axis}
		\end{tikzpicture}~~
		\begin{tikzpicture}[scale=0.6]
		\begin{axis}[
		title=(h),
		title style={at={(0.5,-0.6)}},
		width=5.5cm,
		cycle list name=mark list,
		ybar,
		enlargelimits=0.15,
        ymin=0.0,ymax=1,
		legend style={at={(0.65,0.2)}, anchor=north,legend columns=-1},
		ylabel={Correct ordering rate},
		xlabel={Sample size},
		symbolic x coords={3k,5k,10k},
		xtick=data,
		nodes near coords,
		nodes near coords align={vertical},
		]
		\addplot [ fill=purple,postaction={pattern=north east lines}] coordinates {(3k,0.79) (5k,0.88) (10k,0.92)};
		\addplot [fill=green,postaction={pattern=dots}] coordinates {(3k,0.0) (5k,0.0) (10k,0.0)};
		\legend{LaHiCaSl,LSTC}
		\end{axis}
		\end{tikzpicture}
		\caption{ (a-h) Accuracy of the estimated causal order with LaHiCaSl (purple) and LSTC (green) for Cases 1-8 (the higher the better).}
		\vspace{-0.4cm}
		\label{fig:F1-scores}
	\end{center}
\end{figure}
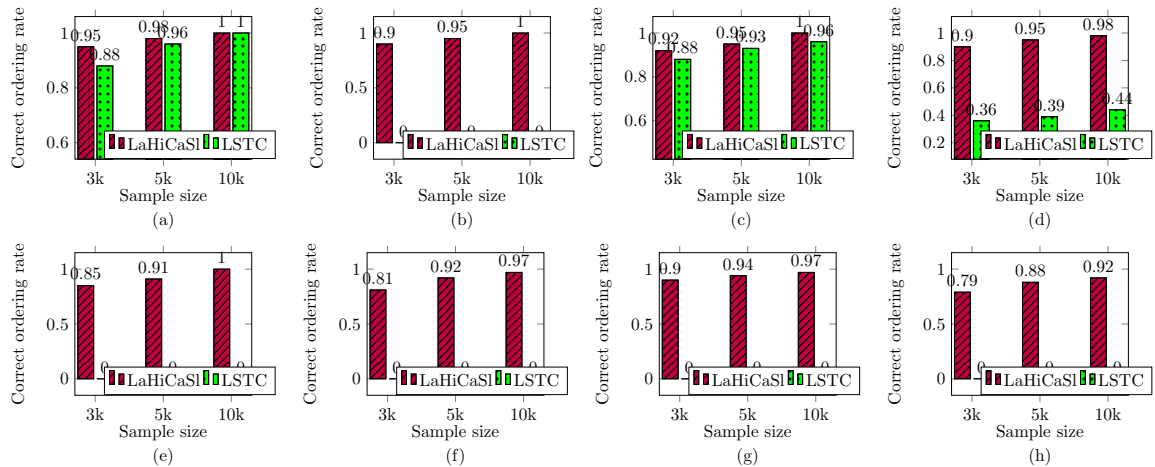

\noindent\textbf{Causal Ordering Results}: Since CLRG, CLNJ, BPC, and FOFC algorithms cannot recover the causal direction between latent variables, we only reported the results from the LSTC algorithm and our algorithm on causal order learning in Figure~\ref{fig:F1-scores}.
As shown in Figure~\ref{fig:F1-scores}, the accuracy of the identified causal ordering of our method gradually increases to 1 with the sample size in all eight cases. Notice that the correct-ordering rate of LSTC is always equal to zero in Case 2 and $5 \sim 8$. This is because LSTC can not handle multi-latent and hierarchical settings. 
As a consequence, these findings illustrate that our algorithm can handle hierarchical structure and discover the correct causal order.

\section{Real-word Study}\label{Section-Real-word-Study}
In this section, we apply our algorithm to three real-world data sets to show the efficacy of the proposed method.

\subsection{Teacher's Burnout Study}
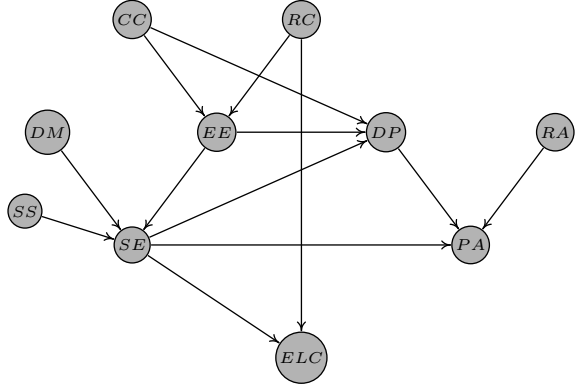
\begin{figure}
	\begin{center}
	    \begin{tikzpicture}[scale=1.5, line width=0.5pt, inner sep=0.2mm, shorten >=.1pt, shorten <=.1pt]
		\draw (0.75, 3.0) node(CC) [circle,inner sep=0pt, minimum size=0.2cm, fill=gray!60,draw] {{\tiny\,$CC$\,}};
		\draw (2.25, 3.0) node(RC) [circle,inner sep=0pt, minimum size=0.2cm, fill=gray!60,draw] {{\tiny\,$RC$\,}};
		\draw (0, 2.0) node(DM) [circle,inner sep=0pt, minimum size=0.2cm,  fill=gray!60,draw] {{\tiny\,$DM$\,}};
		\draw (1.5, 2.0) node(EE) [circle,inner sep=0pt, minimum size=0.2cm, fill=gray!60,draw] {{\tiny\,$EE$\,}};
		\draw (3.0, 2.0) node(DP) [circle,inner sep=0pt, minimum size=0.2cm, fill=gray!60,draw] {{\tiny\,$DP$\,}};
		\draw (4.5, 2.0) node(RA) [circle,inner sep=0pt, minimum size=0.2cm, fill=gray!60,draw] {{\tiny\,$RA$\,}};
		\draw (-0.2, 1.3) node(SS) [circle,inner sep=0pt, minimum size=0.2cm, fill=gray!60,draw] {{\tiny\,$SS$\,}};
		\draw (0.75, 1.0) node(SE) [circle,inner sep=0pt, minimum size=0.2cm, fill=gray!60,draw] {{\tiny\,$SE$\,}};
		\draw (3.75, 1.0) node(PA) [circle,inner sep=0pt, minimum size=0.2cm, fill=gray!60,draw] {{\tiny\,$PA$\,}};
		%
		\draw (2.25, 0.0) node(ELC) [circle,inner sep=0pt, minimum size=0.2cm, fill=gray!60,draw] {{\tiny\,$ELC$\,}};
		\draw[-arcsq] (CC) -- (EE) node[pos=0.5,sloped,above] {};
		\draw[-arcsq] (CC) -- (DP) node[pos=0.5,sloped,above] {};
		\draw[-arcsq] (RC) -- (EE) node[pos=0.5,sloped,above] {};
		\draw[-arcsq] (RC) -- (ELC) node[pos=0.5,sloped,above] {};
		\draw[-arcsq] (EE) -- (DP) node[pos=0.5,sloped,above] {};
		\draw[-arcsq] (EE) -- (SE) node[pos=0.5,sloped,above] {};
		\draw[-arcsq] (DM) -- (SE) node[pos=0.5,sloped,above] {};
		\draw[-arcsq] (SS) -- (SE) node[pos=0.5,sloped,above] {};
		%
		\draw[-arcsq] (SE) -- (DP) node[pos=0.5,sloped,above] {};
		\draw[-arcsq] (SE) -- (PA) node[pos=0.5,sloped,above] {};
		\draw[-arcsq] (SE) -- (ELC) node[pos=0.5,sloped,above] {};
		\draw[-arcsq] (RA) -- (PA) node[pos=0.5,sloped,above] {};
		\draw[-arcsq] (DP) -- (PA) node[pos=0.5,sloped,above] {};
		%
		\end{tikzpicture}\\
        \vspace{-2mm}
		\caption{The hypothesized model of latent factors given in \cite{byrne2010structural}. Note that latent factor \emph{PS} is not included in this model.} \label{fig:realdata_structure.}
		\vspace{-0.3cm}
		\label{fig:teacher-burnout-structures} 
	\end{center}
\end{figure}

Barbara Byrne conducted a study to investigate the impact of organizational (role ambiguity, role conflict, classroom climate, and superior support, etc.) and personality (self-esteem, external locus of control) on three facets of burnout in full-time elementary teachers \citep{byrne2010structural}.
The data set consists of 32 observed variables with 599 samples in total. The details of latent factors and their indicators are shown in Table \ref{Table-teacher-burnout-Data-Description} (See Chapter 6 in \cite{byrne2010structural} for more details). It is noteworthy that the ground-truth latent structure is usually hard to know in practice. Therefore, we here use the hypothesized model given in \cite{byrne2010structural} as a baseline. The structure is shown in Figure \ref{fig:teacher-burnout-structures}. 
Though latent factor \emph{PS} was removed in the hypothesized model given in \cite{byrne2010structural}, we here still input the complete dataset including its corresponding measurement variables (i.e., $PS_1$ and $PS_2$) to analyze.

\begin{center}
  \begin{table}[htp]
    \small
    \center
     \vspace{-5mm}
    \caption{The details of latent factors and their indicators.}
    \vspace{3mm}
    \label{Table-teacher-burnout-Data-Description}
    \begin{tabular}{p{6.0cm}p{6.5cm}}
    \toprule
    \textbf{Latent Factors}  & \textbf{Children (Indicators)}\\ 
    \midrule
    Role Ambiguity (RA)       & $RA_1$, $RA_2$\\\hline 
    Emotional Exhaustion (EE) & $EE_1$, $EE_2$, $EE_3$ \\ \hline
    Depersonalization (DP) & $DP_1$, $DP_2$ \\ \hline
    Role Conflict (RC)& $RC_1$, $RC_2$, $WO_1$, $WO_2$ \\ \hline
    Self-Esteem (SE) & $SE_1$, $SE_2$, $SE_3$ \\ \hline
    Personal Accomplishment (PA) & $PA_1$, $PA_2$, $PA_3$ \\ \hline
    Peer Support (PS) & $PS_1$, $PS_2$ \\ \hline
    Classroom (CC) & $CC_1$, $CC_2$, $CC_3$, $CC_4$ \\ \hline
    Decision Making (DM) & $DM_1$, $DM_2$ \\ \hline
    Superior Support (SS) & $SS_1$, $SS_2$ \\ \hline
    External Locus of Control (ELC) & $ELC_1$, $ELC_2$, $ELC_3$, $ELC_4$, $ELC_5$ \\
    \bottomrule
    \end{tabular}
    \vspace{-3mm}
    \end{table}  
\end{center}

We compared the proposed LaHiCaSl algorithm with LSTC, FOFC, and BPC algorithms.
We run the LaHiCaSl algorithm with the prior knowledge that the underlying graph contains only the 1-latent set, and the kernel width in the HSIC test was set to 0.05. This prior knowledge was also applied to other comparisons. The significance levels of LaHiCaSl, BPC, and FOFC algorithms were all set to $0.00001$. For the convenience of comparison, we here verify the abilities to locate latent variables and infer causal orders separately.

\noindent\textbf{Locating latent variables}: Table \ref{Table-teacher-burnout-latent-variables} gives the results of different algorithms on locating latent variables. Our algorithm achieves the best performance and can locate ten latent variables, while the other three algorithms can only discover partial latent variables. Compared to the hypothesized model given in \cite{byrne2010structural}, the difference is that our method merges \emph{DM} and \emph{SS} into one latent factor $L_{9}$ and discovers latent factor \emph{PS} ($L_7$).
Notice that FOFC and BPC can not discover \emph{RA}, \emph{DP}, \emph{PS}, \emph{DM}, and \emph{SS}. The reason is that those latent factors only have two measurement variables and are incapable of locating by these methods. These results further verify the efficacy of our algorithm.

\begin{center}
  \begin{table}[htp]
    \small
    \center
     \vspace{-5mm}
    \caption{Performance of LaHiCaSl, LSTC, FOFC, and BPC on locating latent variables for teacher's burnout data}
    \vspace{3mm}
    \label{Table-teacher-burnout-latent-variables}
    \begin{tabular}{p{2.5cm}p{11cm}}
    \toprule
    \textbf{Algorithms} & \textbf{Latent Factors and Their Indicators}\\
    \midrule
    LaHiCaSl       & \footnotesize{$L_1 \sim \{RA_1, RA_2\}$; $L_2 \sim \{EE_1, EE_2, EE_3\}$; $L_{3}\sim\{DP_1, DP_2\}$; $L_4\sim\{RC_1, RC_2, WO_1, WO_2\}$; $L_5\sim\{SE_1, SE_2, SE_3\}$; $L_6\sim\{PA_1, PA_2, PA_3\}$; $L_7\sim\{PS_1, PS_2\}$; $L_8\sim\{CC_1, CC_2, CC_3, CC_4\}$; {$L_9\sim \{DM_1, DM_2, SS_1,SS_2\}$; $L_{10} \sim 
    \{ELC_1, ELC_2, ELC_3, ELC_4, ELC_5\}$.}}\\\hline 
    LSTC     & \footnotesize{{$L_1 \sim \{RA_1, RA_2, DM_1, DM_2, SS_1,SS_2, RC_1, RC_2$, $WO_1\}$;  $L_2 \sim \{EE_1, EE_2\}$; $L_3\sim\{PS_1, PS_2\}$; $L_4\sim\{CC_1, CC_2, CC_3, CC_4\}$; $L_{5} \sim \{ELC_1, ELC_2, ELC_3, ELC_4, ELC_5\}$.}} \\\hline 
    FOFC   & \footnotesize{$L_1 \sim \{EE_1, EE_2, EE_3\}$; $L_2\sim\{SE_1, SE_2, SE_3\}$; $L_3\sim\{PA_1, PA_2, PA_3\}$; $L_4\sim\{RC_1, RC_2, RA_1\}$; {$L_5\sim\{CC_1, CC_2, CC_3, CC_4\}$; $L_{6}\sim \{ELC_1, ELC_3, ELC_4, ELC_5\}$.} }\\\hline 
    BPC       & \footnotesize{$L_1 \sim \{EE_1, EE_2, EE_3\}$; $L_2\sim\{SE_1, SE_2, SE_3\}$; $L_3\sim\{PA_1, PA_2, PA_3\}$; $L_4\sim\{DM_2, SS_1,SS_2\}$; {$L_5\sim\{CC_1, CC_2, CC_3, CC_4\}$; $L_{6}\sim \{ELC_1, ELC_2, ELC_3, ELC_4, ELC_5\}$.}}\\
    \bottomrule
    \end{tabular}
    \end{table}  
\end{center}

\noindent\textbf{Inferring causal orders}: Since BPC and FOFC can not discover causal directions between latent variables, we only reported the results from LSTC and our algorithm (LaHiCaSl). The results are given in Table \ref{Table-teacher-burnout-latent-variables-order}. The learned causal orders of our algorithm are almost consistent with Byrne's conclusion, e.g., \emph{role conflict} ($L_4$), \emph{classroom climate} ($L_8$), and \emph{Self-Esteem} ($L_5$) cause \emph{burnout} (including \emph{emotional exhaustion} ($L_2$), \emph{depersonalization} ($L_3$), and \emph{personal accomplishment} ($L_6$)), while some orderings of the LSTC algorithm are not correct. The possible reason is that some latent factors can not be discovered correctly, which further causes some unobserved confounding between latent variables, e.g., $L_5 \to L_2$ (\emph{External Locus of Control} $\to$ $\emph{Emotional Exhaustion}$ due to the latent confounding \emph{Self-Esteem}).

\begin{center}
  \begin{table}[htp]
    \small
    \center
    \caption{Performance of LaHiCaSl and LSTC on inferring causal order among latent variables for teacher's burnout data}
 \vspace{3mm}
    \begin{tabular}{p{3cm}p{8cm}}
    \toprule
    \textbf{Algorithms} & \textbf{Causal Order among Latent Factors}\\ 
    \midrule
    LaHiCaSl       & \footnotesize{$L_1 \succ L_6$; $L_9,L_7 \succ L_5$; $L_4,L_5,L_8 \succ L_2 \succ L_{10} \succ L_{3} \succ L_{6}$ }.\\\hline 
    LSTC     & \footnotesize{$L_1 \succ L_4 \succ L_3 \succ L_5 \succ L_2$ }.\\
    \bottomrule
    \end{tabular}
    \item Note: The subscript of the latent variable in each algorithm corresponds to the subscript of the latent variable of the corresponding algorithm in Table \ref{Table-teacher-burnout-latent-variables}. 
    \label{Table-teacher-burnout-latent-variables-order}
    \end{table}  
\end{center}

\subsection{Multitasking Behaviour Study}\label{Subsec-Multi-Bea-Stud}
We applied our LaHiCaSl algorithm to a multitasking behavior model, represented by a hierarchical SEM \citep{himi2019multitasking}. In particular, the multitasking behavior model contains four latent factors: \emph{Multitasking behavior (Mb)}, \emph{Speed (S)}, \emph{Error (E)}, and \emph{Question (Q)}, where factor \emph{Mb} has no observed variables as children, and \emph{Speed (S)}, \emph{Error (E)}, and \emph{Question (Q)} each has three measured variables. A detailed explanation of the data set is given in Table \ref{Table-Multitasking-Data-Description}. The data set consists of 202 samples.

\begin{center}
  \begin{table}[htp]
    \small
    \center
    \caption{The details of the hypothesized multitasking behavior model \cite{himi2019multitasking} }
 \vspace{3mm}
    \label{Table-Multitasking-Data-Description}
    \begin{tabular}{p{3cm}p{10.5cm}}
    \toprule
    \textbf{Latent Factors} & \textbf{Children (Indicators)}\\ 
    \midrule
    Speed ($S$)       & Correctly marked Numbers ($S_1$), Correctly marked Latters ($S_2$), and Correctly marked Figures ($S_3$)\\\hline 
    Error ($E$)      & Errors marking Numbers ($E_1$), Errors marking Latters ($E_2$), and Errors marking Figures ($E_3$) \\\hline 
    Question ($Q$)       & Correctly answered Questions Par.1 ($Q_1$), Correctly answered Questions Par.2 ($Q_2$), and  Correctly answered Questions Par.3 ($Q_3$)\\\hline
    Multitasking behavior ($Mb$)   & Speed, Error, and  Question\\
    \bottomrule
    \end{tabular}
    \end{table}  
\end{center}

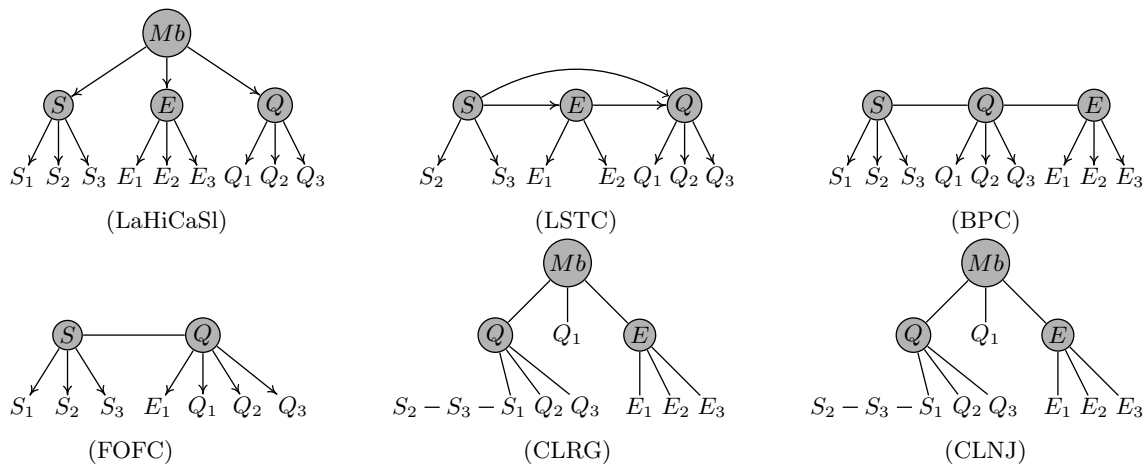
\begin{figure}[htp]
  \setlength{\abovecaptionskip}{0pt}
	\setlength{\belowcaptionskip}{-6pt}
	\vspace{-0.1cm}
	\begin{center}
	    \begin{tikzpicture}[scale=1.2, line width=0.5pt, inner sep=0.2mm, shorten >=.1pt, shorten <=.1pt]
		\draw (1.2, 2.4) node(Mb) [circle,inner sep=0pt, minimum size=0.2cm, fill=gray!60,draw] {{\footnotesize\,$Mb$\,}};
		\draw (0, 1.6) node(S) [circle,inner sep=0pt, minimum size=0.2cm,  fill=gray!60,draw] {{\footnotesize\,$S$\,}};
		\draw (1.2, 1.6) node(E) [circle,inner sep=0pt, minimum size=0.2cm, fill=gray!60,draw] {{\footnotesize\,$E$\,}};
		\draw (2.4, 1.6) node(Q) [circle,inner sep=0pt, minimum size=0.2cm, fill=gray!60,draw] {{\footnotesize\,$Q$\,}};
		\draw[-arcsq] (Mb) -- (S) node[pos=0.5,sloped,above] {};
		\draw[-arcsq] (Mb) -- (E) node[pos=0.5,sloped,above] {};
		\draw[-arcsq] (Mb) -- (Q) node[pos=0.5,sloped,above] {};
		\draw (-0.4, 0.8) node(S1) [] {{\footnotesize\,$S_{1}$\,}};
		\draw (0, 0.8) node(S2) [] {{\footnotesize\,$S_2$\,}};
		\draw (0.4, 0.8) node(S3) [] {{\footnotesize\,$S_3$\,}};
		\draw (0.8, 0.8) node(E1) [] {{\footnotesize\,$E_1$\,}};
		\draw (1.2, 0.8) node(E2) [] {{\footnotesize\,$E_2$\,}};
		\draw (1.6,0.8) node(E3) [] {{\footnotesize\,$E_3$\,}};
		\draw (2.0, 0.8) node(Q1) [] {{\footnotesize\,$Q_{1}$\,}};
		\draw (2.4, 0.8) node(Q2) [] {{\footnotesize\,$Q_{2}$\,}};
		\draw (2.8, 0.8) node(Q3) [] {{\footnotesize\,$Q_{3}$\,}};
		\draw[-arcsq] (S) -- (S1) node[pos=0.5,sloped,above] {};
		\draw[-arcsq] (S) -- (S2) node[pos=0.5,sloped,above] {};
		\draw[-arcsq] (S) -- (S3) node[pos=0.5,sloped,above] {};
		\draw[-arcsq] (E) -- (E1) node[pos=0.5,sloped,above] {};
		\draw[-arcsq] (E) -- (E2) node[pos=0.5,sloped,above] {};
		\draw[-arcsq] (E) -- (E3) node[pos=0.5,sloped,above] {};
		\draw[-arcsq] (Q) -- (Q1) node[pos=0.5,sloped,above] {};
		\draw[-arcsq] (Q) -- (Q2) node[pos=0.5,sloped,above] {};
		\draw[-arcsq] (Q) -- (Q3) node[pos=0.5,sloped,above] {};
		\draw (1.2, 0.3) node(con3) [] {{\footnotesize\,(LaHiCaSl)\,}};
		\end{tikzpicture}~~~~~~~~
	    \begin{tikzpicture}[scale=1.2, line width=0.5pt, inner sep=0.2mm, shorten >=.1pt, shorten <=.1pt]
		%
		\draw (0, 1.6) node(S) [circle,inner sep=0pt, minimum size=0.2cm,  fill=gray!60,draw] {{\footnotesize\,$S$\,}};
		\draw (1.2, 1.6) node(E) [circle,inner sep=0pt, minimum size=0.2cm, fill=gray!60,draw] {{\footnotesize\,$E$\,}};
		\draw (2.4, 1.6) node(Q) [circle,inner sep=0pt, minimum size=0.2cm, fill=gray!60,draw] {{\footnotesize\,$Q$\,}};
		\draw[-arcsq] (S) -- (E) node[pos=0.5,sloped,above] {};
		\draw[-arcsq] (E) -- (Q) node[pos=0.5,sloped,above] {};
		\draw [->] (S) edge[bend right=-30] (Q);
		\draw (-0.4, 0.8) node(S2) [] {{\footnotesize\,$S_2$\,}};
		\draw (0.4, 0.8) node(S3) [] {{\footnotesize\,$S_3$\,}};
		\draw (0.8, 0.8) node(E1) [] {{\footnotesize\,$E_1$\,}};
		\draw (1.6, 0.8) node(E2) [] {{\footnotesize\,$E_2$\,}};
		\draw (2.0, 0.8) node(Q1) [] {{\footnotesize\,$Q_{1}$\,}};
		\draw (2.4, 0.8) node(Q2) [] {{\footnotesize\,$Q_{2}$\,}};
		\draw (2.8, 0.8) node(Q3) [] {{\footnotesize\,$Q_{3}$\,}};
		%
		\draw[-arcsq] (S) -- (S2) node[pos=0.5,sloped,above] {};
		\draw[-arcsq] (S) -- (S3) node[pos=0.5,sloped,above] {};
		\draw[-arcsq] (E) -- (E1) node[pos=0.5,sloped,above] {};
		\draw[-arcsq] (E) -- (E2) node[pos=0.5,sloped,above] {};
		\draw[-arcsq] (Q) -- (Q1) node[pos=0.5,sloped,above] {};
		\draw[-arcsq] (Q) -- (Q2) node[pos=0.5,sloped,above] {};
		\draw[-arcsq] (Q) -- (Q3) node[pos=0.5,sloped,above] {};
		\draw (1.2, 0.3) node(con3) [] {{\footnotesize\,(LSTC)\,}};
		\end{tikzpicture}~~~~~~~~
        \begin{tikzpicture}[scale=1.2, line width=0.5pt, inner sep=0.2mm, shorten >=.1pt, shorten <=.1pt]
		%
		\draw (0, 1.6) node(S) [circle,inner sep=0pt, minimum size=0.2cm,  fill=gray!60,draw] {{\footnotesize\,$S$\,}};
		\draw (2.4, 1.6) node(E) [circle,inner sep=0pt, minimum size=0.2cm, fill=gray!60,draw] {{\footnotesize\,$E$\,}};
		\draw (1.2, 1.6) node(Q) [circle,inner sep=0pt, minimum size=0.2cm, fill=gray!60,draw] {{\footnotesize\,$Q$\,}};
		\draw[-] (Q) -- (S) node[pos=0.5,sloped,above] {};
		\draw[-] (E) -- (Q) node[pos=0.5,sloped,above] {};
		\draw (-0.4, 0.8) node(S1) [] {{\footnotesize\,$S_{1}$\,}};
		\draw (0, 0.8) node(S2) [] {{\footnotesize\,$S_2$\,}};
		\draw (0.4, 0.8) node(S3) [] {{\footnotesize\,$S_3$\,}};
		\draw (0.8, 0.8) node(Q1) [] {{\footnotesize\,$Q_{1}$\,}};
		\draw (1.2, 0.8) node(Q2) [] {{\footnotesize\,$Q_{2}$\,}};
		\draw (1.6, 0.8) node(Q3) [] {{\footnotesize\,$Q_{3}$\,}};
		\draw (2.0, 0.8) node(E1) [] {{\footnotesize\,$E_1$\,}};
		\draw (2.4, 0.8) node(E2) [] {{\footnotesize\,$E_2$\,}};
		\draw (2.8,0.8) node(E3) [] {{\footnotesize\,$E_3$\,}};
		\draw[-arcsq] (S) -- (S1) node[pos=0.5,sloped,above] {};
		\draw[-arcsq] (S) -- (S2) node[pos=0.5,sloped,above] {};
		\draw[-arcsq] (S) -- (S3) node[pos=0.5,sloped,above] {};
		\draw[-arcsq] (E) -- (E1) node[pos=0.5,sloped,above] {};
		\draw[-arcsq] (E) -- (E2) node[pos=0.5,sloped,above] {};
		\draw[-arcsq] (E) -- (E3) node[pos=0.5,sloped,above] {};
		\draw[-arcsq] (Q) -- (Q1) node[pos=0.5,sloped,above] {};
		\draw[-arcsq] (Q) -- (Q2) node[pos=0.5,sloped,above] {};
		\draw[-arcsq] (Q) -- (Q3) node[pos=0.5,sloped,above] {};
		\draw (1.2, 0.3) node(con3) [] {{\footnotesize\,(BPC)\,}};
		\end{tikzpicture}\\
		\begin{tikzpicture}[scale=1.2, line width=0.5pt, inner sep=0.2mm, shorten >=.1pt, shorten <=.1pt]
		%
		\draw (0.5, 1.6) node(S) [circle,inner sep=0pt, minimum size=0.2cm,  fill=gray!60,draw] {{\footnotesize\,$S$\,}};
		\draw (2.0, 1.6) node(Q) [circle,inner sep=0pt, minimum size=0.2cm, fill=gray!60,draw] {{\footnotesize\,$Q$\,}};
		\draw[-] (S) -- (Q) node[pos=0.5,sloped,above] {};
		\draw (0, 0.8) node(S1) [] {{\footnotesize\,$S_{1}$\,}};
		\draw (0.5, 0.8) node(S2) [] {{\footnotesize\,$S_2$\,}};
		\draw (1.0, 0.8) node(S3) [] {{\footnotesize\,$S_3$\,}};
		\draw (1.5, 0.8) node(E1) [] {{\footnotesize\,$E_1$\,}};
		\draw (2.0, 0.8) node(Q1) [] {{\footnotesize\,$Q_{1}$\,}};
		\draw (2.5, 0.8) node(Q2) [] {{\footnotesize\,$Q_{2}$\,}};
		\draw (3.0, 0.8) node(Q3) [] {{\footnotesize\,$Q_{3}$\,}};
		\draw[-arcsq] (S) -- (S1) node[pos=0.5,sloped,above] {};
		\draw[-arcsq] (S) -- (S2) node[pos=0.5,sloped,above] {};
		\draw[-arcsq] (S) -- (S3) node[pos=0.5,sloped,above] {};
		\draw[-arcsq] (Q) -- (E1) node[pos=0.5,sloped,above] {};
		\draw[-arcsq] (Q) -- (Q1) node[pos=0.5,sloped,above] {};
		\draw[-arcsq] (Q) -- (Q2) node[pos=0.5,sloped,above] {};
		\draw[-arcsq] (Q) -- (Q3) node[pos=0.5,sloped,above] {};
		\draw (1.2, 0.3) node(con3) [] {{\footnotesize\,(FOFC)\,}};
		\end{tikzpicture}~~~~~~~~
        \begin{tikzpicture}[scale=1.2, line width=0.5pt, inner sep=0.2mm, shorten >=.1pt, shorten <=.1pt]
		\draw (1.2, 2.4) node(Mb) [circle,inner sep=0pt, minimum size=0.2cm, fill=gray!60,draw] {{\footnotesize\,$Mb$\,}};
		\draw (0.4, 1.6) node(Q) [circle,inner sep=0pt, minimum size=0.2cm,  fill=gray!60,draw] {{\footnotesize\,$Q$\,}};
		\draw (1.2, 1.6) node(Q1) [] {{\footnotesize\,$Q_1$\,}};
		\draw (2.0, 1.6) node(E) [circle,inner sep=0pt, minimum size=0.2cm, fill=gray!60,draw] {{\footnotesize\,$E$\,}};
		\draw[-] (Mb) -- (E) node[pos=0.5,sloped,above] {};
		\draw[-] (Mb) -- (Q1) node[pos=0.5,sloped,above] {};
		\draw[-] (Mb) -- (Q) node[pos=0.5,sloped,above] {};
		\draw (-0.6, 0.8) node(S2) [] {{\footnotesize\,$S_2$\,}};
		\draw (-0.0, 0.8) node(S3) [] {{\footnotesize\,$S_3$\,}};
		\draw (0.6,0.8) node(S1) [] {{\footnotesize\,$S_1$\,}};
		\draw (1.0, 0.8) node(Q2) [] {{\footnotesize\,$Q_2$\,}};
		\draw (1.4, 0.8) node(Q3) [] {{\footnotesize\,$Q_3$\,}};

		\draw (2.0, 0.8) node(E1) [] {{\footnotesize\,$E_{1}$\,}};
		\draw (2.4, 0.8) node(E2) [] {{\footnotesize\,$E_{2}$\,}};
		\draw (2.8, 0.8) node(E3) [] {{\footnotesize\,$E_{3}$\,}};
		\draw[-] (Q) -- (S1) node[pos=0.5,sloped,above] {};
		\draw[-] (S1) -- (S3) node[pos=0.5,sloped,above] {};
		\draw[-] (S3) -- (S2) node[pos=0.5,sloped,above] {};
		\draw[-] (E) -- (E1) node[pos=0.5,sloped,above] {};
		\draw[-] (E) -- (E2) node[pos=0.5,sloped,above] {};
		\draw[-] (E) -- (E3) node[pos=0.5,sloped,above] {};
		\draw[-] (Q) -- (Q2) node[pos=0.5,sloped,above] {};
		\draw[-] (Q) -- (Q3) node[pos=0.5,sloped,above] {};
		\draw (1.2, 0.3) node(con3) [] {{\footnotesize\,(CLRG)\,}};
		\end{tikzpicture}~~~~~~~~
        \begin{tikzpicture}[scale=1.2, line width=0.5pt, inner sep=0.2mm, shorten >=.1pt, shorten <=.1pt]
		\draw (1.2, 2.4) node(Mb) [circle,inner sep=0pt, minimum size=0.2cm, fill=gray!60,draw] {{\footnotesize\,$Mb$\,}};
		\draw (0.4, 1.6) node(Q) [circle,inner sep=0pt, minimum size=0.2cm,  fill=gray!60,draw] {{\footnotesize\,$Q$\,}};
		\draw (1.2, 1.6) node(Q1) [] {{\footnotesize\,$Q_1$\,}};
		\draw (2.0, 1.6) node(E) [circle,inner sep=0pt, minimum size=0.2cm, fill=gray!60,draw] {{\footnotesize\,$E$\,}};
		\draw[-] (Mb) -- (E) node[pos=0.5,sloped,above] {};
		\draw[-] (Mb) -- (Q1) node[pos=0.5,sloped,above] {};
		\draw[-] (Mb) -- (Q) node[pos=0.5,sloped,above] {};
		\draw (-0.6, 0.8) node(S2) [] {{\footnotesize\,$S_2$\,}};
		\draw (-0.0, 0.8) node(S3) [] {{\footnotesize\,$S_3$\,}};
		\draw (0.6,0.8) node(S1) [] {{\footnotesize\,$S_1$\,}};
		\draw (1.0, 0.8) node(Q2) [] {{\footnotesize\,$Q_2$\,}};
		\draw (1.4, 0.8) node(Q3) [] {{\footnotesize\,$Q_3$\,}};

		\draw (2.0, 0.8) node(E1) [] {{\footnotesize\,$E_{1}$\,}};
		\draw (2.4, 0.8) node(E2) [] {{\footnotesize\,$E_{2}$\,}};
		\draw (2.8, 0.8) node(E3) [] {{\footnotesize\,$E_{3}$\,}};
		\draw[-] (Q) -- (S1) node[pos=0.5,sloped,above] {};
		\draw[-] (S1) -- (S3) node[pos=0.5,sloped,above] {};
		\draw[-] (S3) -- (S2) node[pos=0.5,sloped,above] {};
		\draw[-] (E) -- (E1) node[pos=0.5,sloped,above] {};
		\draw[-] (E) -- (E2) node[pos=0.5,sloped,above] {};
		\draw[-] (E) -- (E3) node[pos=0.5,sloped,above] {};
		\draw[-] (Q) -- (Q2) node[pos=0.5,sloped,above] {};
		\draw[-] (Q) -- (Q3) node[pos=0.5,sloped,above] {};
		\draw (1.2, 0.3) node(con3) [] {{\footnotesize\,(CLNJ)\,}};
		\end{tikzpicture}\\		
		\caption{The output of LaHiCaSl, LSTC, BPC, FOFC, CLRG, and CLNJ on the multitasking behavior data.}
		\vspace{-0.3cm}
		\label{fig:multitasking-behaviour-structures} 
	\end{center}
\end{figure}

Figure \ref{fig:multitasking-behaviour-structures} gives the performance of different algorithms. The significance levels of LaHiCaSl, BPC, and FOFC were set to $0.001$, $0.0001$, and $0.000001$ respectively. The reason for choosing different significance levels is that BPC and FOFC algorithms will output an empty graph when the significance level is $0.001$. Here, we chose the `better' significance such that the output graphs of BPC and FOFC algorithms are closer to the ground-truth graph.
The result of our output is consistent with the model given in \cite{himi2019multitasking}, which indicates the effectiveness of our method. Note that although GIN and BPC discover three other latent variables that have observed children, neither finds latent factor \emph{Multitasking behavior}. CLRG and CLNJ output the same result and capture the latent variable \emph{Multitasking behavior}; however, They fail to find the latent variable \emph{Speed}. These results again indicate that our algorithm has better performance than other algorithms in learning hierarchical structure.

\subsection{Mental Ability Study}\label{Real-data-Mental-Ability}
We finally apply our LaHiCaSl algorithm to a classic dataset, i.e., Holzinger $\&$ Swineford1939 dataset. This data set consists of mental ability test scores from 301 American 7th- and 8th-grade students. We focus on 9 out of the original 26 tests as done in~\cite{joreskog2016multivariate}. We here use the hypothesized model given in Chapter 8 in ~\cite{joreskog2016multivariate} as a baseline. More specifically, this model contains four latent factors:  \emph{General (G)}, \emph{Visual (Vi)}, \emph{Verbal (Ve)}, and \emph{Speed (S)}, where factor \emph{G} has no observed variables as children, and \emph{Visual (Vi)}, \emph{Verbal (Ve)}, and \emph{Speed (S)} each has three measured variables. A detailed explanation of the data set is given in Table \ref{Table-mental-ability-model}.

\begin{center}
  \begin{table}[htp]
    \small
    \center
     \vspace{-5mm}
    \caption{Details of the hypothesized mental ability model \citep{joreskog2016multivariate}. }
    \label{Table-mental-ability-model}
    \begin{tabular}{p{3cm}p{10.5cm}}
    \toprule
    \textbf{Latent Factors} & \textbf{Children (Indicators)}\\ 
    \midrule
    Visual ($Vs$)       & Visual perception (${Vs}_1$), Cubes (${Vs}_2$), and Lozenges (${Vs}_3$)\\\hline 
    Verbal ($Vb$)      & Paragraph comprehension (${Vb}_1$), Sentence completion (${Vb}_2$), and Word meaning (${Vb}_3$) \\\hline 
    Speed ($S$)       & Speeded addition ($S_1$), Speeded counting of dots ($S_2$), and  Speeded discrimination straight and curved capitals ($S_3$)\\\hline
    General ($G$)   & Visual, Verbal, and  Speed\\
    \bottomrule
    \end{tabular}
    \end{table}  
\end{center}

\begin{figure}[htp]
  \setlength{\abovecaptionskip}{0pt}
	\setlength{\belowcaptionskip}{-6pt}
	\vspace{-0.1cm}
	\begin{center}
	    \begin{tikzpicture}[scale=1.2, line width=0.5pt, inner sep=0.2mm, shorten >=.1pt, shorten <=.1pt]
		\draw (1.2, 2.4) node(G) [circle,inner sep=0pt, minimum size=0.4cm, fill=gray!60,draw] {{\footnotesize\,$G$\,}};
		\draw (0, 1.6) node(S) [circle,inner sep=0pt, minimum size=0.4cm,  fill=gray!60,draw] {{\footnotesize\,$Vs$\,}};
		\draw (1.2, 1.6) node(E) [circle,inner sep=0pt, minimum size=0.4cm, fill=gray!60,draw] {{\footnotesize\,$Vb$\,}};
		\draw (2.4, 1.6) node(Q) [circle,inner sep=0pt, minimum size=0.4cm, fill=gray!60,draw] {{\footnotesize\,$S$\,}};
		\draw[-arcsq] (G) -- (S) node[pos=0.5,sloped,above] {};
		\draw[-arcsq] (G) -- (E) node[pos=0.5,sloped,above] {};
		\draw[-arcsq] (G) -- (Q) node[pos=0.5,sloped,above] {};
		\draw (-0.4, 0.8) node(S1) [] {{\footnotesize\,$Vs_1$\,}};
		\draw (0, 0.8) node(S2) [] {{\footnotesize\,$Vs_2$\,}};
		\draw (0.4, 0.8) node(S3) [] {{\footnotesize\,$Vs_3$\,}};
		\draw (0.8, 0.8) node(E1) [] {{\footnotesize\,$Vb_1$\,}};
		\draw (1.2, 0.8) node(E2) [] {{\footnotesize\,$Vb_2$\,}};
		\draw (1.6,0.8) node(E3) [] {{\footnotesize\,$Vb_3$\,}};
		\draw (2.0, 0.8) node(Q1) [] {{\footnotesize\,$S_1$\,}};
		\draw (2.4, 0.8) node(Q2) [] {{\footnotesize\,$S_2$\,}};
		\draw (2.8, 0.8) node(Q3) [] {{\footnotesize\,$S_3$\,}};
		\draw[-arcsq] (S) -- (S1) node[pos=0.5,sloped,above] {};
		\draw[-arcsq] (S) -- (S2) node[pos=0.5,sloped,above] {};
		\draw[-arcsq] (S) -- (S3) node[pos=0.5,sloped,above] {};
		\draw[-arcsq] (E) -- (E1) node[pos=0.5,sloped,above] {};
		\draw[-arcsq] (E) -- (E2) node[pos=0.5,sloped,above] {};
		\draw[-arcsq] (E) -- (E3) node[pos=0.5,sloped,above] {};
		\draw[-arcsq] (Q) -- (Q1) node[pos=0.5,sloped,above] {};
		\draw[-arcsq] (Q) -- (Q2) node[pos=0.5,sloped,above] {};
		\draw[-arcsq] (Q) -- (Q3) node[pos=0.5,sloped,above] {};
		\draw (1.2, 0.3) node(con3) [] {{\footnotesize\,(LaHiCaSl)\,}};
		\end{tikzpicture}~~~~~~~~
	    \begin{tikzpicture}[scale=1.2, line width=0.5pt, inner sep=0.2mm, shorten >=.1pt, shorten <=.1pt]
		%
		\draw (0, 1.6) node(S) [circle,inner sep=0pt, minimum size=0.4cm,  fill=gray!60,draw] {{\footnotesize\,$Vs$\,}};
		\draw (2.0, 1.6) node(Q) [circle,inner sep=0pt, minimum size=0.4cm, fill=gray!60,draw] {{\footnotesize\,$S$\,}};
		\draw[-arcsq] (Q) -- (S) node[pos=0.5,sloped,above] {};
		\draw (-0.5, 0.8) node(S1) [] {{\footnotesize\,$Vs_1$\,}};
		\draw (0, 0.8) node(S2) [] {{\footnotesize\,$Vs_2$\,}};
		\draw (0.5, 0.8) node(S3) [] {{\footnotesize\,$Vs_3$\,}};
		\draw (1.6, 0.8) node(Q1) [] {{\footnotesize\,$S_1$\,}};
		\draw (2.4, 0.8) node(Q2) [] {{\footnotesize\,$S_2$\,}};
		%
		\draw[-arcsq] (S) -- (S1) node[pos=0.5,sloped,above] {};
		\draw[-arcsq] (S) -- (S2) node[pos=0.5,sloped,above] {};
		\draw[-arcsq] (S) -- (S3) node[pos=0.5,sloped,above] {};
		\draw[-arcsq] (Q) -- (Q1) node[pos=0.5,sloped,above] {};
		\draw[-arcsq] (Q) -- (Q2) node[pos=0.5,sloped,above] {};
		%
		\draw (1.2, 0.3) node(con3) [] {{\footnotesize\,(LSTC)\,}};
		\end{tikzpicture}~~~~~~~~
        \begin{tikzpicture}[scale=1.2, line width=0.5pt, inner sep=0.2mm, shorten >=.1pt, shorten <=.1pt]
		%
		\draw (0, 1.6) node(Vb) [circle,inner sep=0pt, minimum size=0.4cm,  fill=gray!60,draw] {{\footnotesize\,$Vb$\,}};
		\draw (2.0, 1.6) node(S) [circle,inner sep=0pt, minimum size=0.4cm, fill=gray!60,draw] {{\footnotesize\,$S$\,}};
		\draw[-] (Vb) -- (S) node[pos=0.5,sloped,above] {};
		\draw (-0.5, 0.8) node(Vb1) [] {{\footnotesize\,$Vb_1$\,}};
		\draw (0, 0.8) node(Vb2) [] {{\footnotesize\,$Vb_2$\,}};
		\draw (0.5, 0.8) node(Vb3) [] {{\footnotesize\,$Vb_3$\,}};
		\draw (1.4, 0.8) node(S2) [] {{\footnotesize\,$S_2$\,}};
		\draw (2.0, 0.8) node(S3) [] {{\footnotesize\,$S_3$\,}};
		\draw (2.6, 0.8) node(Vs3) [] {{\footnotesize\,$Vs_3$\,}};
		\draw[-arcsq] (Vb) -- (Vb1) node[pos=0.5,sloped,above] {};
		\draw[-arcsq] (Vb) -- (Vb2) node[pos=0.5,sloped,above] {};
		\draw[-arcsq] (Vb) -- (Vb3) node[pos=0.5,sloped,above] {};
		\draw[-arcsq] (S) -- (S2) node[pos=0.5,sloped,above] {};
		\draw[-arcsq] (S) -- (S3) node[pos=0.5,sloped,above] {};
		\draw[-arcsq] (S) -- (Vs3) node[pos=0.5,sloped,above] {};
		\draw (1.2, 0.3) node(con3) [] {{\footnotesize\,(BPC)\,}};
		\end{tikzpicture}\\
		\begin{tikzpicture}[scale=1.2, line width=0.5pt, inner sep=0.2mm, shorten >=.1pt, shorten <=.1pt]
		%
		\draw (0, 1.6) node(Vs) [circle,inner sep=0pt, minimum size=0.4cm,  fill=gray!60,draw] {{\footnotesize\,$Vs$\,}};
		\draw (2.0, 1.6) node(Vb) [circle,inner sep=0pt, minimum size=0.4cm, fill=gray!60,draw] {{\footnotesize\,$Vb$\,}};
		\draw[-] (Vs) -- (Vb) node[pos=0.5,sloped,above] {};
		\draw (-0.5, 0.8) node(Vs1) [] {{\footnotesize\,$Vs_1$\,}};
		\draw (0, 0.8) node(Vs2) [] {{\footnotesize\,$Vs_2$\,}};
		\draw (0.5, 0.8) node(Vs3) [] {{\footnotesize\,$Vs_3$\,}};
		\draw (1.4, 0.8) node(Vb1) [] {{\footnotesize\,$Vb_1$\,}};
		\draw (2.0, 0.8) node(Vb2) [] {{\footnotesize\,$Vb_2$\,}};
		\draw (2.6, 0.8) node(Vb3) [] {{\footnotesize\,$Vb_3$\,}};
		\draw[-arcsq] (Vs) -- (Vs1) node[pos=0.5,sloped,above] {};
		\draw[-arcsq] (Vs) -- (Vs2) node[pos=0.5,sloped,above] {};
		\draw[-arcsq] (Vs) -- (Vs3) node[pos=0.5,sloped,above] {};
		\draw[-arcsq] (Vb) -- (Vb1) node[pos=0.5,sloped,above] {};
		\draw[-arcsq] (Vb) -- (Vb2) node[pos=0.5,sloped,above] {};
		\draw[-arcsq] (Vb) -- (Vb3) node[pos=0.5,sloped,above] {};
		\draw (1.2, 0.3) node(con3) [] {{\footnotesize\,(FOFC)\,}};
		\end{tikzpicture}~~~~~~~~
        \begin{tikzpicture}[scale=1.2, line width=0.5pt, inner sep=0.2mm, shorten >=.1pt, shorten <=.1pt]
		\draw (1.2, 2.4) node(G) [circle,inner sep=0pt, minimum size=0.4cm, fill=gray!60,draw] {{\footnotesize\,$G$\,}};
		\draw (-0.5, 1.6) node(Vs1) [] {{\footnotesize\,${Vs}_1$\,}};
		\draw (0.1, 1.6) node(Vs) [circle,inner sep=0pt, minimum size=0.4cm,  fill=gray!60,draw] {{\footnotesize\,${Vs}$\,}};
		\draw (1.5, 1.6) node(Vb1) [] {{\footnotesize\,${Vb}_1$\,}};
		\draw (0.7, 1.6) node(Vb) [circle,inner sep=0pt, minimum size=0.4cm, fill=gray!60,draw] {{\footnotesize\,${Vb}$\,}};
		\draw (2.0, 1.6) node(S) [circle,inner sep=0pt, minimum size=0.4cm, fill=gray!60,draw] {{\footnotesize\,$S$\,}};
		\draw (2.7, 1.6) node(S11) [circle,inner sep=0pt, minimum size=0.4cm, fill=gray!60,draw] {{\footnotesize\,$S'$\,}};
		\draw[-] (G) -- (Vs1) node[pos=0.5,sloped,above] {};
		\draw[-] (G) -- (Vs) node[pos=0.5,sloped,above] {};
		\draw[-] (G) -- (Vb1) node[pos=0.5,sloped,above] {};
		\draw[-] (Vb1) -- (Vb) node[pos=0.5,sloped,above] {};
		\draw[-] (G) -- (S) node[pos=0.5,sloped,above] {};
		\draw[-] (S) -- (S11) node[pos=0.5,sloped,above] {};
		\draw (-0.4, 0.8) node(Vs2) [] {{\footnotesize\,${Vs}_2$\,}};
		\draw (0.2, 0.8) node(Vs3) [] {{\footnotesize\,${Vs}_3$\,}};
		\draw (0.8, 0.8) node(Vb2) [] {{\footnotesize\,${Vb}_2$\,}};
		\draw (1.3,0.8) node(Vb3) [] {{\footnotesize\,${Vb}_3$\,}};
		\draw (2.1, 0.8) node(S1) [] {{\footnotesize\,$S_1$\,}};
		\draw (2.5, 0.8) node(S2) [] {{\footnotesize\,$S_2$\,}};
		\draw (2.9, 0.8) node(S3) [] {{\footnotesize\,$S_3$\,}};
		\draw[-] (Vs) -- (Vs2) node[pos=0.5,sloped,above] {};
		\draw[-] (Vs) -- (Vs3) node[pos=0.5,sloped,above] {};
		\draw[-] (Vb) -- (Vb2) node[pos=0.5,sloped,above] {};
		\draw[-] (Vb) -- (Vb3) node[pos=0.5,sloped,above] {};
		\draw[-] (S) -- (S1) node[pos=0.5,sloped,above] {};
		\draw[-] (S11) -- (S2) node[pos=0.5,sloped,above] {};
		\draw[-] (S11) -- (S3) node[pos=0.5,sloped,above] {};
		\draw (1.2, 0.3) node(con3) [] {{\footnotesize\,(CLRG)\,}};
		\end{tikzpicture}~~~~~~~~
        \begin{tikzpicture}[scale=1.2, line width=0.5pt, inner sep=0.2mm, shorten >=.1pt, shorten <=.1pt]
		\draw (1.2, 2.4) node(G) [circle,inner sep=0pt, minimum size=0.4cm, fill=gray!60,draw] {{\footnotesize\,$G$\,}};
		\draw (-0.5, 1.6) node(Vs) [circle,inner sep=0pt, minimum size=0.4cm,  fill=gray!60,draw] {{\footnotesize\,${Vs}$\,}};
		\draw (0.1, 1.6) node(Vs1) [] {{\footnotesize\,${Vs}_1$\,}};
		\draw (0.8, 1.6) node(Vb1) [] {{\footnotesize\,${Vb}_1$\,}};
		\draw (1.5, 1.6) node(Vb) [circle,inner sep=0pt, minimum size=0.4cm, fill=gray!60,draw] {{\footnotesize\,${Vb}$\,}};
		\draw (2.0, 1.6) node(S) [circle,inner sep=0pt, minimum size=0.4cm, fill=gray!60,draw] {{\footnotesize\,$S$\,}};
		\draw (2.7, 1.6) node(S11) [circle,inner sep=0pt, minimum size=0.4cm, fill=gray!60,draw] {{\footnotesize\,$S'$\,}};
		\draw[-] (G) -- (Vs1) node[pos=0.5,sloped,above] {};
		\draw[-] (G) -- (Vs) node[pos=0.5,sloped,above] {};
		\draw[-] (Vb1) -- (Vb) node[pos=0.5,sloped,above] {};
		\draw[-] (G) -- (S) node[pos=0.5,sloped,above] {};
		\draw[-] (S) -- (S11) node[pos=0.5,sloped,above] {};
		\draw (-0.5, 0.8) node(Vs2) [] {{\footnotesize\,${Vs}_2$\,}};
		\draw (0.2, 0.8) node(Vs3) [] {{\footnotesize\,${Vs}_3$\,}};
		\draw (0.8, 0.8) node(Vb2) [] {{\footnotesize\,${Vb}_2$\,}};
		\draw (1.3,0.8) node(Vb3) [] {{\footnotesize\,${Vb}_3$\,}};
		\draw (2.1, 0.8) node(S1) [] {{\footnotesize\,$S_1$\,}};
		\draw (2.5, 0.8) node(S2) [] {{\footnotesize\,$S_2$\,}};
		\draw (2.9, 0.8) node(S3) [] {{\footnotesize\,$S_3$\,}};
		\draw[-] (Vs1) -- (Vb1) node[pos=0.5,sloped,above] {};
		\draw[-] (Vs) -- (Vs2) node[pos=0.5,sloped,above] {};
		\draw[-] (Vs) -- (Vs3) node[pos=0.5,sloped,above] {};
		\draw[-] (Vb) -- (Vb2) node[pos=0.5,sloped,above] {};
		\draw[-] (Vb) -- (Vb3) node[pos=0.5,sloped,above] {};
		\draw[-] (S) -- (S1) node[pos=0.5,sloped,above] {};
		\draw[-] (S11) -- (S2) node[pos=0.5,sloped,above] {};
		\draw[-] (S11) -- (S3) node[pos=0.5,sloped,above] {};
		\draw (1.2, 0.3) node(con3) [] {{\footnotesize\,(CLNG)\,}};
		\end{tikzpicture}\\		
		\caption{The output of LaHiCaSl, LSTC, BPC, FOFC, CLRG, and CLNJ on the mental ability data. }
		\vspace{-0.3cm}
		\label{fig:mental-ability-structures} 
	\end{center}
\end{figure}
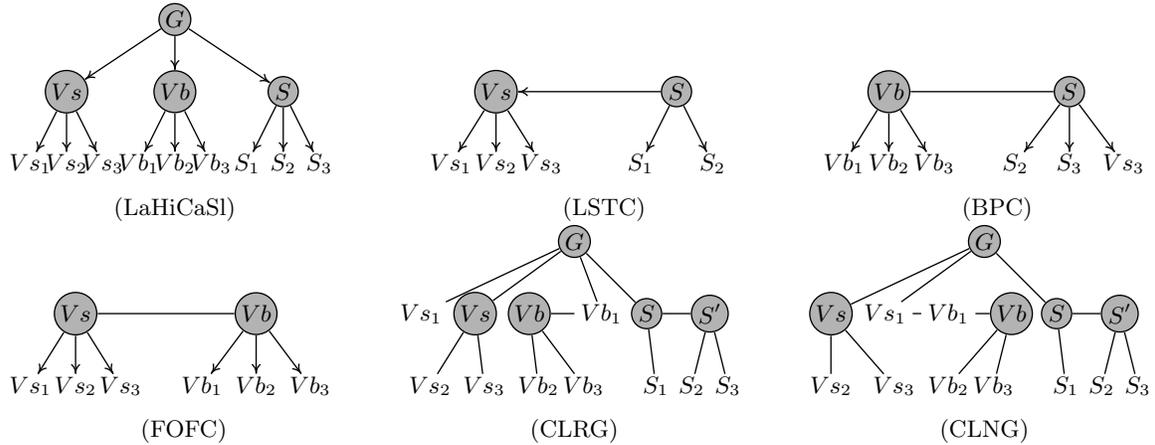

Figure \ref{fig:mental-ability-structures} shows the performance of different algorithms. The significance levels of LaHiCaSl, BPC, and FOFC were set to $0.01$, $0.0001$, and $0.0001$, respectively. Analogous to the process in Section \ref{Subsec-Multi-Bea-Stud}, we chose the small significance level such that the output graphs of BPC and FOFC algorithms are closer to the ground-truth graph. We find that the result of our output is consistent with the model given in \cite{joreskog2016multivariate}, which indicates the effectiveness of our method. It is expected that LSTC, BPC, and FOFC algorithms do not perform well when a latent variable has only unmeasured variables as children (i.e., \emph{General}). 
We notice that CLRG and CLNJ both can capture the latent variable \emph{General}, but they introduce a redundant latent variable $S'$. In summary, the proposed method gives more reliable results than others.

\section{Conclusion}\label{Sec-Conclusion}
In this paper, we proposed the Generalized Independent Noise (GIN) condition in the linear non-Gaussian acyclic causal model, which includes the Independent Noise (IN) condition as a special case. We provide the necessary and sufficient graphical criteria under which the GIN condition holds. Furthermore, we show that GIN can be leveraged to estimate a particular type of latent linear non-Gaussian causal model, LiNGLaH, and show that LiNGLaH is mostly identifiable in terms of GIN conditions under mild assumptions. 
Different from many existing approaches, the latent variables in LiNGLaH may not have observed variables as children, and the graph structure may be beyond a tree.
We proposed a two-phase approach to estimate the LiNGLaH, including locating the latent variables and identifying their causal structure.
Experimental results on both simulation data and real-world data further verified the efficacy of our algorithm.

One of our future research directions is to apply GIN to address more general cases, e.g., eliminating the measurement assumption (no observed variables affect latent variables). 
Another direction of future work is to extend the GIN to the case of a nonlinear latent model, existing techniques \citep{hoyer2009ANM,zhang2009PNL,peters2014causal} may help to address this issue.

\section*{Acknowledgements}
FX would like to acknowledge the support by the Natural Science Foundation of China (62306019) and the Disciplinary funding of Beijing Technology and Business University (STKY202302). KZ would like to acknowledge the support from NSF Grant 2229881, the National Institutes of Health (NIH) under Contract R01HL159805, and grants from Apple Inc., KDDI Research Inc., Quris AI, and Florin Court Capital.
ZM's research was supported by the China Scholarship Council (CSC).

\appendix

\revision{
\section{Differences from Previous Conference Papers}\label{Appendix-difference-previous-conferencepaper}

This paper builds on the conference papers by \cite{xie2020GIN} and \cite{xie2022identification}, and is significantly extended in both theoretical and empirical aspects:
\begin{itemize}[labelindent=.5em,leftmargin=30pt,align=left,itemsep=1pt,topsep=1pt]
    \item[1.] Theoretical results. 
    \begin{itemize}[leftmargin=-10pt,align=left,itemsep=1pt,topsep=0pt]
        \item [] 1.1) We extend the original graphical criteria of the GIN condition presented in our previous work ~\citep{xie2020GIN} and give a more general GIN representation that applies to any linear non-Gaussian acyclic causal models. Specifically, we offer necessary and sufficient graphical conditions under which the GIN condition holds, without relying on the \textit{Purity assumption} (i.e., no direct edge between observed variables), and the \textit{Double-Pure Child Variable assumption} (i.e., latent variable sets can have any number of measurement variables as children"). This result is provided in Theorem 2 of Section \ref{Subsec-graph-criteria}. Additionally, we introduce new graphical criteria for the GIN condition relying on Trek-separation, as presented in Theorem \ref{Theorem-GIN-representation-Trek-separation} of Section \ref{Subsec-graph-criteria}.  
        These graphical criteria further help handle a broader range of latent structures compared to the work by \citet{xie2020GIN,xie2022identification}.
        \item [] 1.2) The proposed algorithm can identify more general latent graphs in this paper compared to that in \citet{xie2022identification} (Section \ref{Sec-Application}). Specifically, we extend the estimation framework in \citet{xie2022identification} to n-factor models, allowing multiple latent confounders behind any two variables. Moreover, we show that the proposed method can be directly extended to cover causal relationships among observed variables as well (Appendix \ref{Appendix-Subsec-infer-observed}). All theoretical results presented in Section 4 regarding the estimation of latent causal graphs are more general than those given in \citet{xie2020GIN, xie2022identification}.
    \end{itemize}
    \item[2.] Empirical estimation and evaluation.
    \begin{itemize}[leftmargin=-10pt,align=left,itemsep=1pt,topsep=0pt]
       \item [] 2.1) The proposed algorithm is more reliable and statistically more efficient in estimating the structure of LiNGLaH with limited samples in practical scenarios. Specifically, we find that in certain cases when identifying global causal clusters, testing the rank (second-order statistics) is equivalent to testing the GIN condition (higher-order statistics). Consequently, in such cases, GIN tests can be replaced with rank deficiency tests, as detailed in Section \ref{Sub-Section-Practical-Algorithm}.

     \item [] 2.2) We provide a more comprehensive empirical evaluation. Specifically, \citet{xie2022identification} only considered 1-factor models in simulated experiments, while, in this paper, we further investigate n-factor models to validate our proposed method (Section \ref{Sec-Experiment}). Additionally, we apply our proposed method to a new real-world data set, providing further evidence of the effectiveness of our proposed method (Section \ref{Real-data-Mental-Ability}).
     \end{itemize}
\end{itemize}
}

\section{Notations and Symbols}\label{appendix-defitions}
\begin{Definition}[Path]
In a DAG, a \textbf{path} is a sequence of nodes $\{V_1, V_2, \dots, V_r\}$ such that $V_i$ and $V_{i+1}$ are adjacent in $\mathcal{G}$, where $1 \le i<r$. Further, if the edges between $V_i$ and $V_{i+1}$ have their arrow pointing to $V_{i+1}$ for $i=1,2, \dots,r-1$, we say that the path is \textbf{directed} from $V_1$ to $V_r$.
\end{Definition}
\begin{Definition}[Collider]
A \textbf{collider} on a path $\{V_1, \dots, V_p\}$ is a node $V_i$ , $1< i < p$, such that $V_{i-1}$ and $V_{i+1}$ are parents of $V_{i}$.
\end{Definition}
\begin{Definition}[Trek]
A \textbf{trek} between $V_i$ and $V_j$ is a path that does not contain any colliders in $\mathcal{G}$.
\end{Definition}
\begin{Definition}[{Source}]
In a trek, a \textbf{source} is a node that has no arrows pointing to it.
\end{Definition}
\begin{Definition}[d-separation]
A path $p$ in a DAG is said to be d-separated (or blocked) by a set of nodes $\mathbf{Z}$ if and only if
\begin{itemize}[leftmargin=30pt,align=left,itemsep=1pt,topsep=0pt]
    \item $p$ contains a chain $V_i \to V_k \to V_j$ or a fork $V_i \leftarrow V_k \rightarrow V_j$ such that the middle node $V_k$ is in $\mathbf{Z}$, or 
    \item $p$ contains a collider $V_i \to V_k \leftarrow V_j$ such that the middle node $V_k$ is not in $\mathbf{Z}$ and such that no descendant of $V_k$ is in $\mathbf{Z}$.
    \vspace{-3mm}
\end{itemize}
\end{Definition}
A set $\mathbf{Z}$ is said to d-separate $\mathbf{A}$ and $\mathbf{B}$ if and only if $\mathbf{Z}$ blocks every path from a node in $\mathbf{A}$ to a node in $\mathbf{B}$.
\revision{
\begin{Definition}[Causal Markov Condition \citep{spirtes2000causation}]
    Given a set of variables $\mathbf{V}$ whose causal structure is represented by a DAG $\mathcal{G}$, every variable in $\mathbf{V}$ is probabilistically independent of its non-descendants in $\mathcal{G}$ given its parents in $\mathcal{G}$.
\end{Definition}
In terms of d-separation, the causal Markov condition states that if $\mathbf{A}$ and $\mathbf{B}$ are d-separated by $\mathbf{Z}$ in true $\mathcal{G}$, then $\mathbf{A}$ and $\mathbf{B}$ are probabilistically independent conditional on $\mathbf{Z}$ according to the true probability measure.
\begin{Definition}[Causal Faithfulness Condition \citep{spirtes2000causation}]
    Given a set of variables $\mathbf{V}$ whose causal structure is represented by a DAG $\mathcal{G}$, the joint probability of $\mathbf{V}$, $P(\mathbf{V})$, is faithful to $\mathcal{G}$ in the sense that $P(\mathbf{V})$ implies no conditional independence relations not already entailed by the causal Markov condition.
\end{Definition}
In terms of d-separation, the causal faithfulness condition states that if $\mathbf{A}$ and $\mathbf{B}$ are probabilistically independent conditional on $\mathbf{Z}$ according to the true probability measure, then $\mathbf{A}$ and $\mathbf{B}$ are d-separated by $\mathbf{Z}$ in true $\mathcal{G}$.
}
\begin{Definition}[\textbf{trek-separation (t-separation \citep{Sullivant-T-separation})}]
Let $\mathbf{A},\mathbf{B},\mathbf{C_{A}}$, and $\mathbf{C_{B}}$ be four subsets of $\mathbf{V}$ in a DAG. We say that the order pair $(\mathbf{C_{A}}, \mathbf{C_{B}})$ {t-separates} $\mathbf{A}$ from $\mathbf{B}$ if, for every trek $(\tau_1;\tau_2)$ from a vertex in $\mathbf{A}$ to a vertex in $\mathbf{B}$, either $\tau_1$ contains a vertex in $\mathbf{C_{A}}$ or $\tau_2$ contains a vertex in $\mathbf{C_{B}}$.
\end{Definition}
It is worth noting that the concept of \emph{t-separation} is a more general separation criterion than \emph{d-separation} in a linear causal model \citep{Sullivant-T-separation}.

\section{Proofs}\label{Appendix-proofs}
Before we present the proofs of our results, we first introduce an important theorem that characterizes the independence of two linear statistics.

\begin{Theorem}[{Darmois-Skitovitch Theorem \citep{Kagan-1973book}}]
Define two random variables ${V_1}$ and ${V_2}$ as linear combinations of independent random variables $n_1, \dots, n_p$:
\begin{flalign}
{V_1} = \sum\limits_{i = 1}^p {{\alpha _i}} {n_i}, \quad \quad {V_2} = \sum\limits_{i = 1}^p {{\beta _i}} {n_i}.
\end{flalign}
where the $\alpha_{i}, \beta_{i}$ are constant coefficients.
If ${V_1}$ and ${V_2}$ are independent, then the random variables ${n_j}$ for which ${\alpha _j}{\beta _j} \ne 0$ are Gaussian. 
\end{Theorem}

In other words, this theorem states that if there exists a non-Gaussian ${n_j}$ for which ${\alpha _j}{\beta _j} \ne 0$, ${V_1}$ and ${V_2}$ are dependent.

We now give a theorem that characterizes the vanishing determinants of a cross-covariance matrix by using the notion of t-separation.

\begin{Theorem}[{Seth-Kellt-Jan Theorem 2.8 in \cite{Sullivant-T-separation}} (SKJ Theorem)]
Let $\mathbf{A},\mathbf{B}$ be four subsets of $\mathbf{V}$. 
The cross-covariance $\Sigma_{\mathbf{A},\mathbf{B}}$ has rank less than or equal to $r$ in all linear directed graphical models if and only if there exist subsets $\mathbf{C_{A}}$ and $\mathbf{C_{B}}$ of $\mathbf{V}$ with $|{\mathbf{C_{A}}}| + | {\mathbf{C_{B}}}| \leq r$ such that $(\mathbf{C}_{A}; \mathbf{C}_{B})$ {t-separates} $A$ from $B$. 
\end{Theorem}

\noindent\textbf{Proof of Proposition \ref{pro-extension-IN}}

\begin{proof}
The proof is straightforward, based on Lemma 1 in \cite{shimizu2011directlingam}, we here omit it.
\end{proof}

\noindent\textbf{Proof of Theorem \ref{Theorem-mathematical-GIN}}

\begin{proof}
Without loss of generality, assume that each component of $\mathcal{S}$ has a zero mean, and that both $\mathbf{E}_Y$ and $\mathbf{E}_Z$ are zero-mean. 

If we can find a non-zero vector $\omega$ such that $\omega^\intercal A = 0$, then $\omega^\intercal \mathbf{Y} =\omega^\intercal A\mathcal{S} + \omega^\intercal \mathbf{E}_Y = \omega^\intercal \mathbf{E}_Y$, which will be independent from $\mathbf{Z}$ in light of conditions 2) and 3), i.e., the GIN condition for $\mathbf{Y}$ given $\mathbf{Z}$ holds true.

We now construct the vector $\omega$. If conditions 2) and 3) hold, we have
$\mathbb{E}[\mathbf{Y}\mathbf{Z}^\intercal] =  A \boldsymbol{\Sigma}_{\mathcal{S},\mathbf{Z}}$, which is determined by $(\mathbf{Y},\mathbf{Z})$.
We now show that under conditions 4), for any non-zero vector $\omega$, $\omega^\intercal A = 0$ if and only if $\omega^\intercal A \boldsymbol{\Sigma}_\mathbf{\mathcal{S},\mathbf{Z}} = 0$ or equivalently $\omega^\intercal \mathbb{E}[\mathbf{Y}\mathbf{Z}^\intercal] = 0$ and that such a vector $\omega$ exists.

Suppose $\omega^\intercal A = 0$, it is trivial to see $\omega^\intercal A \boldsymbol{\Sigma}_{\mathcal{S},\mathbf{Z}} = 0$. Notice that condition 4) implies that $\textrm{rank}(A \boldsymbol{\Sigma}_{\mathcal{S},\mathbf{Z}}) \leq s$ because $\textrm{rank}(A \boldsymbol{\Sigma}_{\mathcal{S},\mathbf{Z}}) \leq \textrm{min}(\textrm{rank}(A), \textrm{rank}(\boldsymbol{\Sigma}_{\mathcal{S},\mathbf{Z}}))$ and $\textrm{rank}(A) = s$. Further according to Sylvester Rank Inequality, we have $\textrm{rank}(A \boldsymbol{\Sigma}_{\mathcal{S},\mathbf{Z}}) \geq \textrm{rank}(A) + \textrm{rank}( \boldsymbol{\Sigma}_{\mathcal{S},\mathbf{Z}}) - s = s$. Therefore, $\textrm{rank}(A \boldsymbol{\Sigma}_{\mathcal{S},\mathbf{Z}}) = s$. Because of condition 1), there must exist a non-zero vector $\omega$, determined by $(\mathbf{Y, Z})$, such that $\omega^\intercal \mathbb{E}[\mathbf{YZ}^\intercal] = \omega^\intercal A \boldsymbol{\Sigma}_{\mathcal{S},\mathbf{Z}}  = 0$. Moreover, this equality implies $\omega^\intercal A = 0$ because $\boldsymbol{\Sigma}_{\mathcal{S},\mathbf{Z}}$ has $s$ rows and has rank $s$. With this $\omega$, we have  $E_{\mathbf{Y}||\mathbf{Z}} = \omega^\intercal \mathbf{E}_Y$ and is independent from $\mathbf{Z}$. Thus the theorem holds.
\end{proof}

\noindent\textbf{Proof of Proposition \ref{Propo-IN-Special_Case}: IN condition is the special case of GIN condition}\label{appendix-IN-sepcial-GIN}

\begin{proof}
For Statement 1, we show that $(\mathbf{Z}, \ddot{{Y}})$ follows the GIN condition implies that $(\mathbf{Z}, {Y})$ follows the IN condition.
If $(\mathbf{Z}, \ddot{{Y}})$ follows the GIN condition, then there must exist a non-zero vector $\ddot{\omega}$ so that $\ddot{\omega}^\intercal \mathbb{E}[\ddot{Y}\mathbf{Z}^\intercal] =0$. This equality implies 
\begin{align}
    \ddot{\omega}^\intercal \mathbb{E}[\left[\begin{matrix}
{Y}\\
{\mathbf{Z}}
\end{matrix}\right]\mathbf{Z}^\intercal]=\ddot{\omega}^\intercal \left[\begin{matrix}
{\mathbb{E}[Y\mathbf{Z}^\intercal]}\\
{\mathbb{E}[\mathbf{Z}\mathbf{Z}^\intercal}]
\end{matrix}\right]  =\mathbf{0}.
\end{align}
Because $\mathbb{E}[\mathbf{ZZ}^\intercal]$ is  non-singular, we further have
\begin{align}
    \ddot{\omega}^\intercal \left[\begin{matrix}
{\mathbb{E}[Y\mathbf{Z}^\intercal]\mathbb{E}^{-1}[\mathbf{Z}\mathbf{Z}^{\intercal}]}\\
{\mathbf{I}}
\end{matrix}\right] = \mathbf{0}.
\end{align}
Let $\omega$ be the first $|Y|$ dimensions of $\ddot{\omega}$ and 
$\tilde{\omega}$ be the second $|\mathbf{Z}|$ dimensions of $\ddot{\omega}$. 
Then we have 
\begin{align}
    \ddot{\omega} {\mathbb{E}[Y\mathbf{Z}^\intercal]\mathbb{E}^{-1}[\mathbf{Z}\mathbf{Z}^{\intercal}]} + \tilde{\omega}{\mathbf{I}} = \mathbf{0}.
\end{align}
Based on the above equation, and without loss of generality, one solution to $\ddot{\omega}^\intercal$ can be expressed as 
\begin{align}
    \ddot{\omega}^\intercal = \{{\omega}, -{\omega}{\mathbb{E}[Y\mathbf{Z}^\intercal]\mathbb{E}^{-1}[\mathbf{Z}\mathbf{Z}^{\intercal}]}\}.
\end{align}
Thus, we have 
\begin{align}
    \ddot{\omega}^\intercal \ddot{Y}= 
        \omega Y  -\omega {\mathbb{E}[Y\mathbf{Z}^\intercal]\mathbb{E}^{-1}[\mathbf{Z}\mathbf{Z}^{\intercal}]} \mathbf{Z}.
\end{align}
Based on the definition of the GIN condition, we have that
$E_{\ddot{Y}||\mathbf{Z}} = \omega Y  -\omega {\mathbb{E}[Y\mathbf{Z}^\intercal]\mathbb{E}^{-1}[\mathbf{Z}\mathbf{Z}^{\intercal}]} \mathbf{Z}$ is independent from $\mathbf{Z}$. It is easy to see that $ Y  - {\mathbb{E}[Y\mathbf{Z}^\intercal]\mathbb{E}^{-1}[\mathbf{Z}\mathbf{Z}^{\intercal}]} \mathbf{Z}$ is independent from $\mathbf{Z}$. Thus, $(\mathbf{Z}, {Y})$ follows the IN condition.

For Statement 2, If $(\mathbf{Z}, {Y})$ follows the IN condition, we have
\begin{align}\label{eq-pro1-2-w}
    \tilde{\omega} = \mathbb{E}[Y \mathbf{Z}^\intercal]\mathbb{E}^{-1}[\mathbf{Z} \mathbf{Z}^\intercal].
\end{align}
Let $\ddot{\omega}=[1^\intercal \; ,-\tilde{\omega}^\intercal]^\intercal$, we get
\begin{align}\noindent
    \ddot{{\omega}}^\intercal \mathbb{E}[{\ddot{Y}}\mathbf{Z}^\intercal]& =[1^\intercal \; ,-\tilde{\omega}^\intercal] \mathbb{E}[\left[\begin{matrix}
{Y}\\
{\mathbf{Z}}
\end{matrix}\right]\mathbf{Z}^\intercal] \\
& =[1^\intercal \; ,-\tilde{\omega}^\intercal] \left[\begin{matrix}
{\mathbb{E}[Y\mathbf{Z}^\intercal]} \noindent\\ \noindent
{\mathbb{E}[\mathbf{Z}\mathbf{Z}^\intercal}]
\end{matrix}\right] 
={\mathbb{E}[Y\mathbf{Z}^\intercal]}-\tilde{\omega}{\mathbb{E}[\mathbf{Z}\mathbf{Z}^\intercal}]. \label{eq-pro1-2-wEYZ}
\end{align}
From Equations \ref{eq-pro1-2-w} and \ref{eq-pro1-2-wEYZ}, we have $\ddot{{\omega}}^\intercal \mathbb{E}[{\ddot{Y}}\mathbf{Z}^\intercal]=0$. That is to say, $\ddot{{\omega}}$ satisfies $\ddot{{\omega}}^{\intercal}\mathbb{E}[{\ddot{Y}}\mathbf{Z}^\intercal]=0$ and that $\ddot{{\omega}}^{\intercal} \neq \mathbf{0}$.

Now, we show that $\ddot{{\omega}}^{\intercal}\ddot{Y}$ is independent from $\mathbf{Z}$. We know that $Y-\tilde{\omega}^{\intercal}\mathbf{Z}$ is independent from $\mathbf{Z}$ based on the definition of the IN condition.
It is easy to see that $\ddot{{\omega}}^{\intercal}\ddot{Y}=[1^\intercal \; ,-\tilde{\omega}^\intercal]\left[\begin{matrix}
{Y}\\
{\mathbf{Z}}
\end{matrix}\right]=Y-\tilde{\omega}^{\intercal}\mathbf{Z}$ is independent from $\mathbf{Z}$. Therefore, $(\mathbf{Z}, \ddot{{Y}})$ follows the GIN condition.
\end{proof}

\noindent\textbf{Proof of Theorem \ref{Theorem:GIN graphical}: Graphical Criteria of GIN Condition}

\revision{
\begin{proof}
The ``if" part: 
To prove this conclusion, we need to demonstrate that under Conditions $1\sim3$, we can transform $\mathbf{Y}$ and $\mathbf{Z}$ into the generative form of Theorem \ref{Theorem-mathematical-GIN}.
First suppose that there exists such a subset of variables, $\mathcal{S}$, that satisfies the three conditions. 
Because of condition 1), i.e., the common cause of $V$ in $\mathbf{Y}$ and each variable in $\mathcal{S}$, if there is any, is in $\mathcal{S}$, condition 2), i.e., $\mathcal{S}$ is a $\mathbf{Y}$-side choke-point set and because according to the linear acyclic model, each $\mathcal{S}_i$ can be expressed as $\mathcal{S}_i = \mathcal{S}'_{i} + \mathcal{S}''_{i}$, where $\mathcal{S}'_i$ represents the contribution from a subset of $\mathcal{S}'_i$'s parent node, consisting of variables that are the shared parents of $\mathcal{S}'_i$ and $\mathbf{Z}$. Meanwhile, $\mathcal{S}''_i$ is a linear function of $\mathrm{Pa}'(\mathcal{S}'_i)$ plus independent noise. Here, $\mathrm{Pa}'(\mathcal{S}'_i)$ denotes the contributions from a specific subset of the parent node of $\mathcal{S}'_i$, characterized by variables that are not the shared parents of $\mathcal{S}'_i$ and $\mathbf{Z}$. 
Thus $\mathcal{S}$ can be written as $\mathcal{S} = \mathcal{S}' + \mathcal{S}''$. It worth noting that $\mathcal{S}''_i \CI \mathcal{S}''$. This is because the common cause of $V$ in $\mathbf{Y}$ and each variable in $\mathcal{S}$, if there is any, is in $\mathcal{S}$ and every active path between $\mathbf{Y}$ and $\mathbf{Z}$ contains a node $S$ in $\mathcal{S}$.
Furthermore, because $\mathcal{S}$ is always causally earlier of the $\mathbf{Y}-$side.
Hence, we know that each component of $\mathbf{Y}$ can be written as a linear function of $\mathcal{S}'$ and some independent errors (which is independent from $\mathcal{S}'$, including $\mathcal{S}''$). By a slight abuse of notation, here we use $\mathcal{S}'$ also to denote the vector of the variables in $\mathcal{S}'$. Then we have 
\begin{equation} \label{eq_t1}
\mathbf{Y} = A \mathcal{S}' + \mathbf{E}'_Y,
\end{equation}
where $A$ is an appropriate linear transformation, $\mathbf{E}'_Y$ is independent from $\mathcal{S}$, but its components are not necessarily independent from each other. In fact, according to the linear acyclic causal model, each observed or unobserved variable is a linear combination of the underlying noise terms $\varepsilon_i$. In equation (\ref{eq_t1}), $\mathcal{S}'$ and $\mathbf{E}'_Y$ are linear combinations of disjoint sets of the noise terms $\varepsilon_i$, implied by the directed acyclic structure over all observed and unobserved variables.

Let us then write $\mathbf{Z}$ as linear combinations of the noise terms. We then show that because of condition 2), i.e., that $\mathcal{S}$ is a $\mathbf{Y}$-side choke-point set, if any noise term $\varepsilon_i$ is present in $\mathbf{E}'_Y$, it will not be among the noise terms in the expression of $\mathbf{Z}$. Otherwise, if $Z_j$ also involves $\varepsilon_i$, then the direct effect of $\varepsilon_i$, among all observed or unobserved variables, is a common cause of $Z_j$ and some component of $\mathbf{Y}$.
This active path between $Z_j$ and that component of $\mathbf{Y}$, however, cannot be blocked by $\mathcal{S}$ because no component of $\mathcal{S}$ is on the active path, as implied by the fact that 
when $\mathcal{S}'$ is written as a linear combination of the underlying noise terms, $\varepsilon_i$ is not among them. Consequently, any noise term in $\mathbf{E}'_Y$ will not contribute to $\mathcal{S}'$ or $\mathbf{Z}$. Hence, we can express $\mathbf{Z}$ as
\begin{equation} \label{eq_t2}
\mathbf{Z} = B \mathcal{S}' + \mathbf{E}'_Z,
\end{equation}
where $\mathbf{E}_Z'$, which is determined by $\mathcal{S}'$ and $\mathbf{Z}$, is independent from $\mathbf{E}_Y'$.

Further considering the condition on the dimensionality of $\mathcal{S}$ and condition 3), one can see that the assumptions in Theorem \ref{Theorem-mathematical-GIN} are satisfied. Therefore, $(\mathbf{Z},\mathbf{Y})$ satisfies the GIN condition.

The ``only-if" part: Then we suppose $(\mathbf{Z, Y})$ satisfies GIN. Consider all sets $\mathcal{S}$ that for any variable $V$ in $\mathbf{Y}$ but not in $\mathcal{S}$, $V$ is not an ancestor of any variable in $\mathcal{S}$ satisfying the conditions in the theorem, and we show that at least one of them satisfies conditions 2) and 3).
Otherwise, if 2) is always violated, then there is an open path between some leaf node in $\mathrm{APa}(\mathbf{Y})$, denoted by $\mathrm{APa}(Y_k)$, and some component of $\mathbf{Z}$, denoted by $Z_j$, and this open path does not go through any common cause of the variables in $\mathrm{APa}(\mathbf{Y})$. Then they have some common cause that does not cause any other variable in  $\mathrm{APa}(\mathbf{Y})$. Consequently, there exists at least one noise term, denoted by $\varepsilon_i$, that contributes to both $\mathrm{APa}(Y_k)$ (and hence $Y_k$) and $Z_j$, but not any other variables in $\mathbf{Y}$. Because of the non-Gaussianity of the noise terms and Darmois-Skitovitch Theorem, if any linear projection of $\mathbf{Y}$, $\omega^\intercal \mathbf{Y}$ is independent from $\mathbf{Z}$, the linear coefficient for $Y_k$ must be zero. Hence $(\mathbf{Z}, \mathbf{Y}\backslash\{Y_k\})$ satisfies GIN, which contradicts the assumption in the theorem. Therefore, there must exist some $\mathcal{S}$ such that 2) holds. 
Next, if 3) is violated, i.e., the rank of the covariance matrix of $\mathcal{S}$ and $\mathbf{Z}$ is smaller than $k$. Then the condition 
$\omega^\intercal \mathbb{E}[\mathbf{YZ}^\intercal] = 0$ does not guarantee that $\omega^\intercal A = 0$. Under the faithfulness assumptions, we then do not have that $\omega^\intercal \mathbf{Y}$ is independent from $\mathbf{Z}$. Hence, condition 3) also holds. 
\end{proof}
}
\noindent\textbf{Proof of Theorem \ref{Theorem-GIN-representation-Trek-separation}}

\begin{proof}
To prove this result, we only need to prove that condition 2) in Theorem \ref{Theorem-GIN-representation-Trek-separation} is equivalent to condition 2) in 
Theorem \ref{Theorem:GIN graphical}.
Because of condition 2), i.e., that the ordered pair $(\emptyset, \mathcal{S})$ $t-$separates $\mathbf{Z}$ and $\mathbf{Y}$ and because according to the definition of Trek separation, i.e., for every trek $(\tau_1;\tau_2)$ from a vertex in $\mathbf{Z}$ to a vertex in $\mathbf{Y}$, either $\tau_1$ contains a vertex in $\emptyset$ or $\tau_2$ contains a vertex in $\mathcal{S}$, the element of $\mathcal{S}$ is causal earlier than $\mathbf{Y}$.
Thus, we know that every active path (trek) between $\mathbf{Y}$ and $\mathbf{Z}$ contains a node $S$ in $\mathcal{S}$, and that $S$ is on the $\mathbf{Y}$ side of every such active path. According to the definition of a side choke-point set, we directly obtain that set $\mathcal{S}$ is a $\mathbf{Y}$-side choke-point set between $\mathbf{Y}$ and $\mathbf{Z}$, i.e., Condition 2 of Theorem \ref{Theorem:GIN graphical} holds. 
\end{proof}

\noindent\textbf{Proof of Theorem \ref{Theorem-global-cluster}: Global Causal Cluster}\label{appendix-proof-theorem-causal-cluster}

\begin{proof}
The "if" part: We will prove this result by contradiction. There are the following three cases:

\noindent \emph{Case 1}: $\mathbf{Y}$ is not a causal cluster and show that $(\mathbf{X} \backslash \mathbf{Y},\tilde{\mathbf{Y}})$ violates the GIN condition, leading to the contradiction. Since $\mathbf{Y}$ is not a causal cluster, without loss of generality, $\mathbf{Pa}(\mathbf{Y})$ must contain at least two different parental latent variable sets, denoted by $\mathcal{L}_1$ and $\mathcal{L}_2$.
Now, we show that there is no non-zero vector $\omega$ such that ${\omega}^\intercal \tilde{\mathbf{Y}}$ is independent from $\mathbf{X} \backslash {\mathbf{Y}}$. 
Because the condition 2 holds, i.e., there is no subset $\tilde{\mathbf{Y}'} \subseteq \tilde{\mathbf{Y}}$ such that $(\mathbf{X} \backslash \tilde{\mathbf{Y}'},\tilde{\mathbf{Y}}')$ follows the GIN condition, the number of pure children of $\mathcal{L}_1$ in $\tilde{\mathbf{Y}}$ is smaller than $|\mathcal{L}_1|+1$ and the number of pure children of $\mathcal{L}_2$ in $\tilde{\mathbf{Y}}$ is smaller than $|\mathcal{L}_2|+1$.
Thus, we obtain that there is no ${\omega}\neq 0$ such that ${\omega}^\intercal \mathbb{E}[\tilde{\mathbf{Y}}((\mathbf{X} \backslash \mathbf{Y})^\intercal]=0$.
That is to say, ${\omega}^\intercal \tilde{\mathbf{Y}}$ is dependent on $\mathbf{X} \backslash \mathbf{Y}$, i.e., $(\mathbf{X} \backslash \mathbf{Y},\tilde{\mathbf{Y}})$ violates the GIN condition, which leads to the contradiction.

\noindent \emph{Case 2}: $\mathbf{Y}$ is a causal cluster but is not global. First, since $\mathbf{Y}$ is not a global causal cluster, we know that there is at least one node, denoted by $V_i$, such that 1) $V_i$ is in $\mathbf{Ch}(\mathbf{Y})$ but not in $\mathbf{Y}$, and that 2) there exists a direct path between $V_i$ and one of node $V_j$ in $\mathbf{Y}$. Without loss of generality, we assume $V_i \to V_j$ and $V_j \in \tilde{\mathbf{Y}}$. Thus, there exists an active path connecting $V_i$ and $V_j$ that does not go through $L(\mathbf{Y})$. Further, according to Theorem \ref{Theorem:GIN graphical}, $(\mathbf{X} \backslash \mathbf{Y},\tilde{\mathbf{Y}})$ violates the GIN condition, which leads to the contradiction (violates condition (2)).

\noindent \emph{Case 3}: $\mathbf{Y}$ is a global causal cluster but $|L(\mathbf{Y})| \neq |\tilde{\mathbf{Y}}|-1$ (Here $|\tilde{\mathbf{Y}}|=\mathrm{Len}+1$). Since $\mathbf{Y}$ is a causal cluster, $|L(\mathbf{Y})|=|L(\tilde{\mathbf{Y}}))|$.
First, we consider the case where $|L(\tilde{\mathbf{Y}}))| < |\tilde{\mathbf{Y}}|-1$.
If $|L(\tilde{\mathbf{Y}}))| < |\tilde{\mathbf{Y}}|-1$, we always can find a subset $\tilde{\mathbf{Y}}' \subseteq \tilde{\mathbf{Y}}$ and
$|\tilde{\mathbf{Y}}'|=|L(\tilde{\mathbf{Y}})| + 1$ such that $(\mathbf{X} \backslash \tilde{\mathbf{Y}'},\tilde{\mathbf{Y}}')$ follows the GIN condition, leading to the contradiction.
We then consider the case where $|L(\tilde{\mathbf{Y}}))| > |\tilde{\mathbf{Y}}|-1$.
According to Theorem \ref{Theorem:GIN graphical}, if $|L(\tilde{\mathbf{Y}}))| > |\tilde{\mathbf{Y}}|-1$, $(\mathbf{X} \backslash \mathbf{Y},\tilde{\mathbf{Y}})$ must violate the GIN condition, which leads to a contradiction.

The "only-if" part: Assume that $\mathbf{Y}$ is a {global causal cluster}. We first know $|L(\mathbf{Y})|=|L(\tilde{\mathbf{Y}}))|$. Since $|L(\mathbf{Y})|=\mathrm{Len}$, $|L(\tilde{\mathbf{Y}}))|=\mathrm{Len}$. According to the definition of the causal cluster, we know
that 1) for any variable $V$ in $\mathbf{Y}$ but not in $L(\mathbf{Y})$, $V$ is not an ancestor of any variable in $L(\mathbf{Y})$, that
2) $L(\mathbf{Y})$ is a $\mathbf{Y}$-side choke-point set.
By Theorem \ref{Theorem:GIN graphical}, we have $(\mathbf{X} \backslash \mathbf{Y},\tilde{\mathbf{Y}})$ follows the GIN condition, i.e., condition 1 holds. Further, by definition of the causal cluster, $L({\tilde{\mathbf{Y}}})$ contains only one parental latent variable set and condition 2 holds.

\end{proof}

\noindent\textbf{Proof of Theorem \ref{Theorem-causal-direction}: Causal Direction among Latent Variables}\label{appendix-causal-direction}

\begin{proof}
For $L(\mathbf{C}_p)$ and $L(\mathbf{C}_q)$, there are two possible causal relations: $L(\mathbf{C}_p) \to L(\mathbf{C}_q)$ and $L(\mathbf{C}_p) \leftarrow L(\mathbf{C}_q)$. Without loss of generality, we assume that the ground-truth causal relationship is $L(\mathbf{C}_p) \to L(\mathbf{C}_q)$.

We will prove this theorem by leveraging Theorem \ref{Theorem:GIN graphical}. That is to say, we show that there exists a set, $L(\mathbf{C}_{p})$ with $0\leq |L(\mathbf{C}_{p})| \leq \textrm{min}(|L(\mathbf{C}_{p})|+|L(\mathbf{C}_{q})|-1, |L(\mathbf{C}_{p})|)$, such that these three conditions of Theorem \ref{Theorem:GIN graphical} hold for $\{P_{|L(\mathbf{C}_p)|+1}, \ldots, P_{2|L(\mathbf{C}_p)|}\}$ and $\{P_{1:|L(\mathbf{C}_p)|}, Q_{1},\ldots, Q_{|L(\mathbf{C}_q)|}\}$. First, since $L(\mathbf{C}_{p})$ and $L(\mathbf{C}_{q})$ are the parents of $P_{1:2|L(\mathbf{C}_p)|}$ and $Q_{1:2|L(\mathbf{C}_q)|}$ respectively, and because $L(\mathbf{C}_p) \to L(\mathbf{C}_q)$,
we know that for any variable $V$ in $\mathbf{C}_{p}$ but not in $L(\mathbf{C}_{p})$, $V$ is not an ancestor of any variable in $L(\mathbf{C}_{p})$ (condition 1 holds).
Next, because there are no confounders behind $L(\mathbf{C}_p)$ and $L(\mathbf{C}_q)$, and $L(\mathbf{C}_p) \cap L(\mathbf{C}_q)=\emptyset$,  we have that all active paths between $\{P_{|L(\mathbf{C}_p)|+1}, \ldots,P_{2|L(\mathbf{C}_p)|}\}$ and $\{P_{1:|L(\mathbf{C}_p)|},Q_{1}, \ldots,Q_{|L(\mathbf{C}_q)|}\}$ go through $L(\mathbf{C}_{p})$ and $L(\mathbf{C}_{p})$ is on the $\{P_{1:|L(\mathbf{C}_p)|},Q_{1}, \ldots,Q_{|L(\mathbf{C}_q)|}\}$ side of all such active paths.
This will imply that $L(\mathbf{C}_{p})$ is a $\{P_{1:|L(\mathbf{C}_p)|},Q_{1}, \ldots,Q_{|L(\mathbf{C}_q)|}\}$-side choke-point set(condition 2 holds). Last, since $\mathbf{C}_p$ and $\mathbf{C}_q$ are two global pure causal cluster in $\mathcal{G}$, the adjacency matrix $\textrm{Adj}_{L(\mathbf{C}_p),\{P_{|L(\mathbf{C}_p)|+1}, \ldots,P_{2|L(\mathbf{C}_p)|}\}}$ has rank $p$, and so dose that of $L(\mathbf{C}_p)$ and $\{P_{1:|L(\mathbf{C}_p)|},Q_{1}, \ldots,Q_{|L(\mathbf{C}_q)|}\}$. According to the rank-faithfulness assumption, we have that the covariance matrix of $L(\mathbf{C}_p)$ and $\{P_{|L(\mathbf{C}_p)|+1}, \ldots, P_{2|L(\mathbf{C}_p)|}\}$ has rank $|L(\mathbf{C}_p|$, and so does that of $L(\mathbf{C}_p)$ and $\{P_{1:|L(\mathbf{C}_p)|}, Q_{1}, \ldots, Q_{|L(\mathbf{C}_q)|}\}$. (condition 3 holds).

For the inverse direction, we next show that $(\{Q_{|L(\mathbf{C}_q)|+1}, \ldots,Q_{2|L(\mathbf{C}_q)|}\}, \{Q_{1:|L(\mathbf{C}_q)|},P_{1}, \ldots,$\\$P_{|L(\mathbf{C}_p)|}\})$ violates the GIN condition. According to Theorem \ref{Theorem:GIN graphical}, we need to show that there exists no set, $\mathcal{S}$ with $0\leq |\mathcal{S}| \leq \textrm{min}(|L(\mathbf{C}_{p})|+|L(\mathbf{C}_{q})|-1, |L(\mathbf{C}_{q})|)$, such that these three conditions of Theorem \ref{Theorem:GIN graphical} hold. We first consider condition 1 of Theorem \ref{Theorem:GIN graphical}. Since $L(\mathbf{C}_{p})$ and $L(\mathbf{C}_{q})$ are the parents of $P_{1:2|L(\mathbf{C}_p)|}$ and $Q_{1:2|L(\mathbf{C}_q)|}$ respectively, we obtain that the minimal set of variables that are parents of any component of $\{Q_{1:|L(\mathbf{C}_q)|}, P_{1}, \ldots, P_{|L(\mathbf{C}_p)|}\}$ is $\{L(\mathbf{C}_{q}), L(\mathbf{C}_{p})\}$. This will imply that $|\mathcal{S}|=|\{L(\mathbf{C}_{q}),L(\mathbf{C}_{p})\}| > |L(\mathbf{C}_{q})|$, which is contradictory to the conditions of Theorem \ref{Theorem:GIN graphical}. Thus, we obtain that $(\{Q_{|L(\mathbf{C}_q)|+1}, \ldots,Q_{2|L(\mathbf{C}_q)|}\},\\ \{Q_{1:|L(\mathbf{C}_q)|},P_{1}, \ldots,P_{|L(\mathbf{C}_p)|}\})$ must violate the GIN condition.

Based on the above analysis, $L(\mathcal{S}_p) \to L(\mathcal{S}_q)$.
\end{proof}

\noindent\textbf{Proof of Proposition of \ref{proposition-identify-global-cluster} }

\begin{proof}
According to Theorem \ref{Theorem-global-cluster} and the updating process of $\mathcal{A}$ (Proposition \ref{Proposition-update-active-data}), the result can be directly proved.
\end{proof}

\noindent\textbf{Proof of Lemma \ref{lemma-pure-impure-cluster}: Pure Cluster}\label{appendix-proof-lemma-1}

\begin{proof}
We will prove this result by contradiction. There are the following two cases:

\emph{Case 1}: $k>|L(\mathbf{C}_1)|+1$ and $\mathbf{C}_1$ is an impure cluster. 
First, since $\mathbf{C}_1$ is an impure cluster, we know there is at least one direct path between $V_i$ and $V_j$. Without loss of generality, we assume that $V_i \to V_j$, and that that $V_i \in \tilde{\mathbf{C}}_1$ and $V_j \notin \tilde{\mathbf{C}}_1$. This will imply that there exists at least one active path connecting $\mathcal{A} \cup \{\mathbf{C}_1 \backslash \{\tilde{\mathbf{C}}_1\}\}$ and $\tilde{\mathbf{C}}_1$ that does not go through $L(\mathbf{C}_1)$. Thus, $L(\mathbf{C}_1)$ is not a choke-point set between $\mathcal{A} \cup \{\mathbf{C}_1 \backslash \{\tilde{\mathbf{C}}_1\}\}$ and $\tilde{\mathbf{C}}_1$. According to condition 2 of Theorem \ref{Theorem:GIN graphical}, we know that $(\mathcal{A} \cup \{\mathbf{C}_1 \backslash \{\tilde{\mathbf{C}}_1\}\},\tilde{\mathbf{C}}_1\})$ violates the GIN condition, which leads to the contradiction.

\emph{Case 2}: $k=|L(\mathbf{C}_1)|+1$ and $\mathbf{C}_1$ is an impure cluster. Because $\mathbf{C}_1$ is an impure cluster, there is at least one direct path between $V_i$ and $V_j$, where $V_i, V_j \in \mathbf{C}_1$. Without loss of generality, we assume $V_i \to V_j$. By leverage Theorem \ref{Theorem:GIN graphical}, we next will show that there does exist $\{\mathbf{P},V_k\} \subset \{\mathcal{A} \backslash \mathbf{C}_1\}$, such that $|\mathbf{P}|=|L(\mathbf{C}_1)|$, and that  $(\{V_i,\mathbf{P}\},\{\mathbf{C}_1, V_k\})$ follows the GIN condition while $(\{V_j,\mathbf{P}\},\{\mathbf{C}_1, V_k\})$ violates the GIN condition. 
We first show the first result, $(\{V_i,\mathbf{P}\},\{\mathbf{C}_1, V_k\})$ follows the GIN condition. Because of \emph{minimal latent hierarchical structure} condition, i.e., each latent set $\mathcal{L}_1$ have at least $2|\mathcal{L}_1|$ pure children relative to $L(\mathbf{C}_1)$, we know that there must exist $V_k \in \{\mathcal{A} \backslash \mathbf{C}_1\}$ and $\mathbf{P} \subset \{\mathcal{A} \backslash \mathbf{C}_1\}$ such that $V_k$ is the child of $L(\mathbf{C}_1)$ and $L(\mathbf{C}_1)$ is a choke point set between $V_i$ between $\{V_k,\mathbf{P}\}$. Therefore, the minimal set of variables that are parents of any component of $\{\mathbf{C}_1, V_k\}$ is $\{L(\mathbf{C}_1), V_i\}$. Hence, we obtain that for any variable $V$ in $\{\mathbf{C}_1,V_k\}$ but not in $\{L(\mathbf{C}_1),V_i\}$, $V$ is not an ancestor of any variable in $\{\mathbf{C}_1,V_k\}$, and that
$\{L(\mathbf{C}_1),V_i\}$ is a choke point set between $\{\mathbf{C}_1, V_k\}$ and $\{V_i,\mathbf{P}\}$. According to the rank-faithfulness assumption, we have that the covariance matrix of $\{L(\mathbf{C}_1), V_i\}$ and $\{\mathbf{C}_1, V_k\}$ has rank $|\{L(\mathbf{C}_1), V_i\}|$, and so does that of $\{L(\mathbf{C}_1), V_i\}$ and $\{V_i,\mathbf{P}\}$. According to Theorem \ref{Theorem:GIN graphical}, we know that $(\{V_i,\mathbf{P}\},\{\mathbf{C}_1, V_k\})$ follows the GIN condition. 
Last, we show $(\{V_j,\mathbf{P}\},\{\mathbf{C}_1, V_k\})$ violates the GIN condition. We have known that the minimal set of variables that are parents of any component of $\{\mathbf{C}_1, V_k\}$ is $\{L(\mathbf{C}_1), V_i\}$. We find that $\{L(\mathbf{C}_1),V_i\}$ is not a choke point set between $\{\mathbf{C}_1, V_k\}$ and $\{V_j,\mathbf{P}\}$. According to Theorem \ref{Theorem:GIN graphical}, we know that $(\{V_j,\mathbf{P}\},\{\mathbf{C}_1, V_k\})$ violates the GIN condition.
\end{proof}

\noindent\textbf{Proof of Lemma \ref{lemma-pure-impure-variables}: Pure sub-Cluster}\label{appendix-proof-lemma-2}

\begin{proof}
According to the Lemma \ref{lemma-pure-impure-cluster}, condition (1) directly holds. 

Condition (2) can be proved by contradiction, analogous to the analysis process of condition (2) of Lemma \ref{lemma-pure-impure-cluster}. The key difference from condition (2) of Lemma \ref{lemma-pure-impure-cluster} is that $\mathbf{Q}$ is a vector. Because of \emph{minimal latent hierarchical structure} condition, i.e., each latent set $\mathcal{L}_1$ have at least $2|\mathcal{L}_1|$ pure children relative to $L(\mathbf{C}_1)$, we can still find that $\mathbf{Q}$. 
\end{proof}

\noindent\textbf{Proof of Proposition \ref{Proposition-merge-rules-cases}: Merging Rules}

\begin{proof}
We will prove these rules by contradiction. We first prove $\mathcal{R}1$. We need to consider the case where $\mathbf{C}_1$ and $\mathbf{C}_2$ do not share the same set of latent variables as parents and $\mathbf{C}_1$ and $\mathbf{C}_2$ have the same size of latent parent variables.
We will show that condition (b) of $\mathcal{R}1$ will violate.
Because $\mathbf{C}_1$ is a global causal cluster, we know that every trek between $\mathbf{C}_1$ and $\mathcal{A} \backslash \mathbf{C}_1$ goes through $L(\mathbf{C}_1)$. Further, because of \emph{minimal latent hierarchical structure} condition, $L(\mathbf{C}_2)$ has at least $2|L(\mathbf{C}_2)|$ pure children. Thus, there exist a variable $V_k \in \mathbf{C}_2$ and $V_k \in \{\mathcal{A} \backslash \{ {\mathbf{C}}_1 \cup \mathbf{C}_2\} \cup \neg_{\mathbf{C}_1}(\tilde{\mathbf{C}}_1) \cup \neg_{\mathbf{C}_2}(V_i)\}$. This will imply that there is one active path that links $V_k$ and $V_i$ through $L(\mathbf{C}_2)$. That is to say,
the number of common components between $\{V_i, \tilde{\mathbf{C}}_1\}$ and $\{\mathcal{A} \backslash \{ {\mathbf{C}}_1 \cup \mathbf{C}_2\} \cup \neg_{\mathbf{C}_1}(\tilde{\mathbf{C}}_1) \cup \neg_{\mathbf{C}_2}(V_i)\}$ greater than or equal to $|\mathcal{L}_1|+1$ (including $\varepsilon_{L(\mathbf{C}_1)}$, $\varepsilon_{L(\mathbf{C}_2)}$ and so on). By Theorem \ref{Theorem:GIN graphical}, $(\mathcal{A} \backslash \{ {\mathbf{C}}_1 \cup \mathbf{C}_2\} \cup \neg_{\mathbf{C}_1}(\tilde{\mathbf{C}}_1) \cup \neg_{\mathbf{C}_2}(V_i),\{V_i, \tilde{\mathbf{C}}_1\})$ violates the GIN condition, which leads to the contradiction.

Now, we prove $\mathcal{R}2$. This proof is similar to the above case. We need to consider the case where $\mathbf{C}_1$ and $\mathbf{C}_2$ do not share the same set of latent variables as parents and $\mathbf{C}_1$ and $\mathbf{C}_2$ have different sizes of latent parent variables. Without loss of generality, we suppose $|L(\mathbf{C}_1)| > |L(\mathbf{C}_2)|$. We will show that condition (b) of $\mathcal{R}2$ will violate. 
Since $\mathbf{C}_1$ is a global cluster, we know that every trek between $\mathbf{C}_1$ and $\mathcal{A} \backslash \mathbf{C}_1$ goes through $L(\mathbf{C}_1)$. Further, because of \emph{minimal latent hierarchical structure} condition, $L(\mathbf{C}_2)$ has at least $2|L(\mathbf{C}_2)|$ pure children. Thus, there exist a variable $V_k \in \mathbf{C}_2$ and $V_k \in \mathcal{A} \backslash \{ {\mathbf{C}}_1 \cup \mathbf{C}_2\} \cup \neg_{\mathbf{C}_1}(\tilde{\mathbf{C}}_1) \cup \neg_{\mathbf{C}_2}(V_i)$. This will imply that there is one active path that links $V_k$ and $V_i$ through $L(\mathbf{L}_2)$. That is to say,
the number of common components between $\{V_i, \tilde{\mathbf{C}}_1\}$ and $\{\mathcal{A} \backslash \{ {\mathbf{C}}_1 \cup \mathbf{C}_2\} \cup \neg_{\mathbf{C}_1}(\tilde{\mathbf{C}}_1) \cup \neg_{\mathbf{C}_2}(V_i)\}$ greater than or equal to $|\mathcal{L}_1|+1$ (including $\varepsilon_{L(\mathbf{C}_1)}$, $\varepsilon_{L(\mathbf{C}_2)}$ and so on). By Theorem \ref{Theorem:GIN graphical}, $(\mathcal{A} \backslash \{ {\mathbf{C}}_1 \cup \mathbf{C}_2\} \cup \neg_{\mathbf{C}_1}(\tilde{\mathbf{C}}_1) \cup \neg_{\mathbf{C}_2}(V_i),\{V_i, \tilde{\mathbf{C}}_1\})$ violates the GIN condition, which leads to the contradiction.
\end{proof}

\noindent\textbf{Proof of Corollary \ref{Corollary-merge-rules-earlycurrent-cases}}

\begin{proof}
It suffices to notice that all elements in $\mathbf{C}_1$ are the children of $\mathcal{L}_1$ and do not affect the variables in $\mathcal{A}$. 
The $\mathcal{R}3.$ and $\mathcal{R}4.$ of corollary follow immediately from
the $\mathcal{R}1.$ and $\mathcal{R}2.$ of Proposition \ref{Proposition-merge-rules-cases} when we update $\mathcal{A}=\mathcal{A} \cup \mathbf{C}_1$.
\end{proof}

\noindent\textbf{Proof of Proposition of \ref{Propostion-testing-GIN-using-Surrogatevariables}}

\begin{proof}
The proof is straightforward, based on the definition of \emph{Surrogate Sets} of Pair $(\mathbf{Z},\mathbf{Y})$, the linearity assumption, and the transitivity of linear causal relations.
\end{proof}

\noindent\textbf{Proof of Proposition \ref{Proposition-update-active-data}}

\begin{proof}
According to Proposition \ref{proposition-identify-global-cluster} and \ref{Proposition-merge-rules-cases}, all elements in $\mathcal{L}$ must be the parent nodes of some nodes in $\mathcal{A}$. That is to say, all nodes in $\mathcal{L}$ must be the leaves when we remove $\mathbf{Ch}(\mathcal{L})$. Thus, the structure of the other variables in $\mathcal{A}'$ is not changed.
Furthermore, because linear causal models are transitive, each latent variable ${L}'_i$ in $\mathcal{A}'$ still have at least $2|{L}'_i|$ pure ``children''. Thus, for the ordered pair $(\mathbf{Z},\mathbf{Y})$, there must exist the surrogate sets of $(\mathbf{Z},\mathbf{Y})$, denoted by $(\mathbf{Z}',\mathbf{Y}')$, where those surrogate sets can be selected from the cluster identified in the previous iterations. According to Proposition \ref{Propostion-testing-GIN-using-Surrogatevariables}, the GIN test of any pair $(\mathbf{Z},\mathbf{Y})$ in $\mathcal{A}'$ are equivalent to the GIN test for the surrogate sets of $(\mathbf{Z},\mathbf{Y})$.
\end{proof}

\noindent\textbf{Proof of Proposition \ref{Proposition-causa-direction-behind-confounders}}\label{appendix-proof-proposition-4}

\begin{proof}
According to Theorem \ref{Theorem-causal-direction}, one can directly prove this result. The key difference from the analysis process of Theorem \ref{Theorem-causal-direction} is that we need to put the $\mathbf{T}_2$ and $\mathbf{T}_1$ into both sides of the ordered pair $(\mathbf{Z},\mathbf{Y})$ when testing GIN condition.
\end{proof}

\noindent\textbf{Proof of Proposition \ref{Proposition-removing-redundant-edges}}

\begin{proof}
We will show this result through Theorem 2.8 in \cite{Sullivant-T-separation}.

(i) There is no directed edge between $\mathcal{L}_p$ and $\mathcal{L}_q$. Because $\mathcal{L}_t$ be the common parents of $\mathbf{C}_i$ and each latent variable set $\mathcal{L}_{\mathbf{S}_i}$ in $\mathcal{L}_\mathbf{S}$ is causally later than $\mathcal{L}_p$ and is causally earlier than $\mathcal{L}_q$, we know that there exist a set, $\mathcal{L}_t \cup \mathcal{L}_\mathbf{S}$, such that all treks between $\mathbf{P}_1$ and $\mathbf{Q}_1$ go through $\mathcal{L}_t \cup \mathcal{L}_\mathbf{S}$, and that each node in $\mathcal{L}_t \cup \mathcal{L}_\mathbf{S}$ is causal earlier than $\mathbf{P}_1$ and $\mathbf{Q}_1$. Thus, $(\emptyset,\mathcal{L}_t \cup \mathcal{L}_\mathbf{S})$ $t-$separates $\mathbf{P}_1$ and $\mathbf{Q}_1$. Based on Theorem 2.8 in \cite{Sullivant-T-separation}, we have that the rank of the cross-covariance matrix of $\{\mathbf{P}_1,\mathbf{Q}_1\} \cup \{\mathbf{T}_1,\mathbf{T}_2\} \cup \mathbf{S}$ is less than or equal to $|\mathcal{L}_t \cup \mathcal{L}_\mathbf{S}|$.

(ii) There is a directed edge between $\mathcal{L}_p$ and $\mathcal{L}_q$. Without loss of generality, we assume that $\mathcal{L}_p \to \mathcal{L}_q$. Because $\mathcal{L}_t$ be the common parents of $\mathbf{C}_i$ and each latent variable set $\mathcal{L}_{\mathbf{S}_i}$ in $\mathcal{L}_\mathbf{S}$ is causally later than $\mathcal{L}_p$ and is causally earlier than $\mathcal{L}_q$, we know that not all treks between $\mathbf{P}_1$ and $\mathbf{Q}_1$ go through $\mathcal{L}_t \cup \mathcal{L}_\mathbf{S}$, e.g., $\mathbf{P}_1 \leftarrow \mathcal{L}_p \to \mathcal{L}_q \to \mathbf{Q}_1$. Thus, $(\emptyset,\mathcal{L}_t \cup \mathcal{L}_\mathbf{S})$ can not $t-$separate $\mathbf{P}_1$ and $\mathbf{Q}_1$. 
Based on Theorem 2.8 in \cite{Sullivant-T-separation}, we have that the rank of the cross-covariance matrix of $\{\mathbf{P}_1,\mathbf{Q}_1\} \cup \{\mathbf{T}_1,\mathbf{T}_2\} \cup \mathbf{S}$ must be large than $|\mathcal{L}_t \cup \mathcal{L}_\mathbf{S}|$.
\end{proof}

\noindent\textbf{Proof of Lemma \ref{Lemma-Phase1-identifiable}}

\begin{proof}
We first show that all latent variables can be located in Phase I of ReLLS. Specifically, In the first iteration, with Proposition \ref{proposition-identify-global-cluster}, one can detect all global causal clusters by testing for GIN conditions. Then, by Propositions \ref{Proposition-merge-rules-cases} and Corollary \ref{Corollary-merge-rules-earlycurrent-cases}, one can detect all latent variable sets. Next, by Propositions \ref{Proposition-update-active-data}, 
the new active variable set is structurally consistent with the ground-truth one, which will ensure that the recursive search process is correct and all latent variables can be located.

Now, we show the causal structure among the latent parents of pure causal clusters. This result follows immediately from the definition of causal cluster---the nodes in any cluster have a common parent and there are no directed edges between them if this causal cluster is a pure cluster.
\end{proof}

\noindent\textbf{Proof of Theorem \ref{Theorem-Model-Identification}: Identification of the LiNGLaM}

\begin{proof}
Based on Lemma \ref{Lemma-Phase1-identifiable}, all latent variables, as well as the causal structure among the latent parents of pure clusters can be identified in Phase I of ReLLS.

We now demonstrate that the causal structure among latent variables within any impure cluster can be identified in Phase II of ReLLS. Specifically, for an impure cluster, $\mathbf{C}_i$, a local root set in $\mathbf{C}_i$ can be found exactly with the condition described in Proposition \ref{Proposition-causa-direction-behind-confounders}. Due to the acyclic assumption, the order of recursive search of LocallyInferCausalStructure is the causal order of the original variables.  Finally, given the causal order of the latent variable sets, we can use the rank-based independence tests to eliminate redundant edges from the fully connected sub-graph (Proposition \ref{Proposition-removing-redundant-edges}). Note that the causal relationship of latent variables within the same latent set is unidentifiable. This is because the 
locations of any two latent variables within a latent set are the same in terms of structure, except for the names. 

Based on the above analysis, we can conclude that the LiNGLaM is (almost) identifiable.
\end{proof}

\noindent\textbf{Proof of Observation \ref{observation-GIN-Cov-equal-zero}}

\begin{proof}
By the definition of GIN condition, $\omega$ needs to satisfy $\omega^\intercal \mathbb{E}[\mathbf{Y}\mathbf{Z}^\intercal] = 0$ and $\omega \neq 0$. Thus, we have $\mathrm{Cov}(E_{\mathbf{Y}||\mathbf{Z}},\mathbf{Z}) = \mathrm{Cov}(\omega^\intercal \mathbf{Y},\mathbf{Z})=\omega^\intercal \mathrm{Cov}(\mathbf{Y},\mathbf{Z}) \equiv 0$. Because the variables follow Gaussian distribution, $E_{\mathbf{Y}||\mathbf{Z}}$ is statistically independent of $\mathbf{Z}$.
\end{proof}

\noindent\textbf{Proof of Proposition \ref{Proposition-GIN-equal-Rank}}

\begin{proof}
The "if" part: We will prove this result by contradiction. There are the following three cases:

\noindent \emph{Case 1}: $\mathbf{Y}$ is not a causal cluster and shows that the rank of the cross-covariance matrix $(\boldsymbol{\Sigma}_{\{\mathcal{A} \backslash \mathbf{Y}\},\tilde{\mathbf{Y}}})$ is greater than $|\mathbf{Y}|-1$, leading to the contradiction. Since $\mathbf{Y}$ is not a causal cluster, without loss of generality, $\mathbf{Pa}(\mathbf{Y})$ must contain at least two different parental latent variable sets, denoted by $\mathcal{L}_1$ and $\mathcal{L}_2$.
Because condition 2 holds, i.e., there is no subset $\tilde{\mathbf{Y}'} \subseteq \tilde{\mathbf{Y}}$ such that the cross-covariance matrix $(\boldsymbol{\Sigma}_{\{\mathcal{A} \backslash \tilde{\mathbf{Y}'}\},\tilde{\mathbf{Y}'}})$ has rank $|\tilde{\mathbf{Y}'}|-1$, the number of pure children of $\mathcal{L}_1$ in $\tilde{\mathbf{Y}}$ is smaller than $|\mathcal{L}_1|+1$ and the number of pure children of $\mathcal{L}_2$ in $\tilde{\mathbf{Y}}$ is smaller than $|\mathcal{L}_2|+1$. Thus, for any subset $\tilde{\mathbf{Y}}$ of $\mathbf{Y}$ with $|\tilde{\mathbf{Y}}|=\mathrm{Len}$, $(\emptyset,L(\mathbf{Y}))$ can not $t-$separates $\{\mathcal{A} \backslash \mathbf{Y}\}$ from ${\tilde{\mathbf{Y}}}$. Because of Seth-Kellt-Jan Theorem Theorem and rank-faithfulness assumption, we know that the cross-covariance matrix $(\boldsymbol{\Sigma}_{\{\mathcal{A} \backslash \mathbf{Y}\},\tilde{\mathbf{Y}}})$ is greater than $|\mathbf{Y}|-1$, which leads to the contradiction.

\noindent \emph{Case 2}: $\mathbf{Y}$ is a causal cluster but is not global. First, since $\mathbf{Y}$ is not a global causal cluster, we know that there is at least one node, denoted by $V_i$, such that 1) $V_i$ is in $\mathbf{Ch}(\mathbf{Y})$ but not in $\mathbf{Y}$, and that 2) there exists a direct path between $V_i$ and one of node $V_j$ in $\mathbf{Y}$. Without loss of generality, we assume $V_i \to V_j$ and $V_j \in \tilde{\mathbf{Y}}$. Thus, there exists a trek connecting $V_i$ and $V_j$ that does not go through $L(\mathbf{Y})$. Hence, according to Seth-Kellt-Jan Theorem, the rank of the cross-covariance matrix $(\boldsymbol{\Sigma}_{\{\mathcal{A} \backslash \mathbf{Y}\},\tilde{\mathbf{Y}}})$ must be greater than $|\mathbf{Y}|-1$, which leads to the contradiction.

\noindent \emph{Case 3}: $\mathbf{Y}$ is a global causal cluster but $|L(\mathbf{Y})| \neq |\tilde{\mathbf{Y}}|-1$ (Here $|\tilde{\mathbf{Y}}|=\mathrm{Len}$). Since $\mathbf{Y}$ is a causal cluster, $|L(\mathbf{Y})|=|L(\tilde{\mathbf{Y}}))|$.
First, we consider the case where $|L(\tilde{\mathbf{Y}}))| < |\tilde{\mathbf{Y}}|-1$.
Since $|L(\tilde{\mathbf{Y}}))| < |\tilde{\mathbf{Y}}|-1$, we always can find a subset $\tilde{\mathbf{Y}}' \subseteq \tilde{\mathbf{Y}}$ and
$|\tilde{\mathbf{Y}}'|=|L(\tilde{\mathbf{Y}})| + 1$ such that the cross-covariance matrix $(\boldsymbol{\Sigma}_{\{\mathcal{A} \backslash \tilde{\mathbf{Y}'}\},\tilde{\mathbf{Y}'}})$ has rank $|\tilde{\mathbf{Y}'}|-1$, leading to the contradiction.
We then consider the case where $|L(\tilde{\mathbf{Y}}))| > |\tilde{\mathbf{Y}}|-1$.
According to Seth-Kellt-Jan Theorem Theorem, if $|L(\tilde{\mathbf{Y}}))| > |\tilde{\mathbf{Y}}|-1$, the rank of the cross-covariance matrix $(\boldsymbol{\Sigma}_{\{\mathcal{A} \backslash \mathbf{Y}\},\tilde{\mathbf{Y}}})$ must be greater than $|\mathbf{Y}|-1$, which leads to the contradiction.

The "only-if" part: Because $\mathbf{Y}$ is a {global causal cluster}. We know $|L(\mathbf{Y})|=|L(\tilde{\mathbf{Y}}))|$. Since $|L(\mathbf{Y})|=\mathrm{Len}-1$, $|L(\tilde{\mathbf{Y}}))|=\mathrm{Len}-1$. According to the definition of causal cluster, we know that all treks between $\{\mathcal{A} \backslash \mathbf{Y}\}$ and ${\mathbf{Y}}$ go through $L(\mathbf{Y})$, and that $L(\mathbf{Y})$ is the parents of ${\mathbf{Y}}$. Thus, $(\emptyset,L(\mathbf{Y}))$ $t-$separates $\{\mathcal{A} \backslash \mathbf{Y}\}$ from ${\mathbf{Y}}$. Because of Seth-Kellt-Jan Theorem Theorem and rank-faithfulness assumption, we know that the cross-covariance matrix $(\boldsymbol{\Sigma}_{\{\mathcal{A} \backslash \mathbf{Y}\},\tilde{\mathbf{Y}}})$ has rank $|\mathbf{Y}|-1$. 
Further, by definition of the causal cluster, $L({\tilde{\mathbf{Y}}})$ contains only one parental latent variable set and condition 2 holds.
\end{proof}

\noindent\textbf{Proof of Corollary \ref{Corollary-causal-direction-observed-variables-behind-confounders}}\label{appendix-proof-corollary-2}

\begin{proof}
    This result follows immediately from Proposition \ref{Proposition-causa-direction-behind-confounders}. The key difference from the analysis process of Proposition \ref{Proposition-causa-direction-behind-confounders} is that we put the measured $\{\mathbf{X}_p,\mathbf{X}_t\}$ and $\{\mathbf{X}_p, X_q, \mathbf{X}_t\}$ into both sides of the ordered pair $(\mathbf{Z},\mathbf{Y})$ when testing GIN condition.
\end{proof}

\noindent\textbf{Proof of Corollary \ref{Corollary-removing-redundant-edges-observed}}

\begin{proof}
    This result follows immediately from Proposition \ref{Proposition-removing-redundant-edges}.
\end{proof}

\noindent\textbf{Proof of Theorem \ref{Theorem-Model-additional-Identification}}

\begin{proof}
The correctness of the LaHiCaSl algorithm with Algorithm \ref{algorithm-locally-infer-causalorder-plus} originates from the following observations:

\begin{itemize}
    \item Firstly, the correctness of Phase I can be obtained from Theorem \ref{Theorem-Model-Identification}.
    \item Secondly, in Phase II (Algorithm \ref{algorithm-locally-infer-causalorder-plus}), by Corollary \ref{Corollary-causal-direction-observed-variables-behind-confounders} and Proposition \ref{Proposition-causa-direction-behind-confounders}, all the unidentified causal edges in impure clusters are identified (Lines $5 \sim 15$  of Algorithm \ref{algorithm-locally-infer-causalorder-plus}).
    \item Lastly, in Phase II (Algorithm \ref{algorithm-locally-infer-causalorder-plus}), by Corollary \ref{Corollary-removing-redundant-edges-observed} and Proposition \ref{Proposition-removing-redundant-edges}, remove the redundant edges among measured variables, and among latent variables (Lines $16 \sim 28$  of Algorithm \ref{algorithm-locally-infer-causalorder-plus}).
\end{itemize}

Based on the above analysis, the LaHiCaSl algorithm with Algorithm \ref{algorithm-locally-infer-causalorder-plus} can output the true causal structure correctly, including the causal relationships among observed variables, between the observed variables and their corresponding latent variable sets, and the causal relationships between the latent variable sets.
\end{proof}

\revision{
\section{Illustrative Examples of Key Concepts}

\begin{Example-set}[Illustrated example for Proposition \ref{Propo-IN-Special_Case}]\label{Example-Connect-IN-GIN}
Consider the causal graph in Figure \ref{fig:IN-GIN}(a). We can observe that $(X_1, \{X_1, X_2\})$ satisfies GIN, while $(X_2, \{X_1, X_2\})$ violates GIN. That is to say, $(X_1,X_2)$ satisfies IN while $(X_2,X_1)$ violates IN. Thus, we can infer that $X_{1} \to X_{2}$. Now, consider the causal graph in Figure \ref{fig:IN-GIN}(b), where there exists a latent confounder $L_1$ between $X_1$ and $X_2$. We can observe that $(\{X_{1},X_{3}\},\{X_{1},X_{2},X_{4}\})$ satisfies the GIN condition while $(\{X_{2},X_{3}\},\{X_{1},X_{2},X_{4}\})$ violates the 
GIN condition. This indicates that there is an edge between $X_{1}$ and $X_{2}$ and $X_{1} \to X_{2}$.
\end{Example-set}
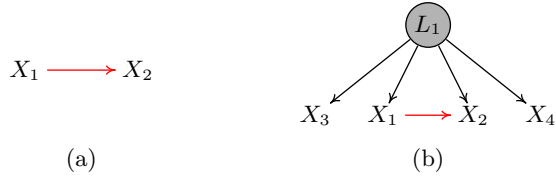
\begin{figure}[htp]
	\begin{center}
	    \begin{tikzpicture}[scale=1.5, line width=0.5pt, inner sep=0.2mm, shorten >=.1pt, shorten <=.1pt]
		\draw (0, -0.4) node(X1) [] {{\footnotesize\,${X}_{1}$\,}};
		\draw (1, -0.4) node(X2) [] {{\footnotesize\,$X_2$\,}};
		\draw (1, -0.8) node(X3) [] {};
		\draw[-arcsq, color=red] (X1) -- (X2) node[pos=0.5,sloped,above] {}; 
        \draw (0.5,-1.2) node(label-ii) [] {{\footnotesize\,(a)\,}};
		\end{tikzpicture}~~~~~~~~~~~~~
        \begin{tikzpicture}[scale=1.5, line width=0.5pt, inner sep=0.2mm, shorten >=.1pt, shorten <=.1pt]
		\draw (1, 0) node(L1) [circle, fill=gray!60,draw] {{\footnotesize\,$L_1$\,}};
		\draw (-0.0, -0.8) node(X1) [] {{\footnotesize\,${X}_{3}$\,}};
		\draw (0.6, -0.8) node(X2) [] {{\footnotesize\,$X_1$\,}};
		\draw (1.4, -0.8) node(X3) [] {{\footnotesize\,$X_2$\,}};
		\draw (2.0, -0.8) node(X4) [] {{\footnotesize\,${X}_{4}$\,}};
		\draw[-arcsq] (L1) -- (X1) node[pos=0.5,sloped,above] {};
		\draw[-arcsq] (L1) -- (X2) node[pos=0.5,sloped,above] {};
		\draw[-arcsq] (L1) -- (X3) node[pos=0.5,sloped,above] {};
		\draw[-arcsq] (L1) -- (X4) node[pos=0.5,sloped,above] {};
		\draw[-arcsq,color=red] (X2) -- (X3) node[pos=0.5,sloped,above] {};
		\draw (1,-1.2) node(label-ii) [] {{\footnotesize\,(b)\,}};
		\end{tikzpicture}
		\caption{An illustrated example for Example \ref{Example-Connect-IN-GIN}. (a) There is no latent confounder between $X_1$ and $X_2$. (b) There is a latent confounder $L_1$ between $X_1$ and $X_2$.} 
		\label{fig:IN-GIN} 
	\end{center}
\end{figure}

\begin{Example-set}[Violation of rank faithfulness]
Consider the causal graph in Figure \ref{fig:explain-assumption-rank-faithfulness}(a), and let $X_1=aL_1+bL_2+\varepsilon_{X_1}$ and $X_2=2aL_1+2bL_2+\varepsilon_{X_2}$. In this case, the rank of the cross-covariance of $\{X_1,X_2\}$ and $\{L_1,L_2\}$ is 1 where the rank-faithfulness is violated,  because $L_1$ and $L_2$ exhibit proportional sets of connection strengths to $X_1$ and $X_2$, i.e, $a/b=2(a/b)$. If letting $L_0=aL_1+bL_2$, then we have $X_1=L_0+\varepsilon_{X_1}$ and $X_2=2L_0+\varepsilon_{X_2}$ as illustrated in Figure \ref{fig:explain-assumption-rank-faithfulness}(b), where the rank is also 1. Because of the violation of rank-faithfulness, we cannot distinguish between the two graphs.
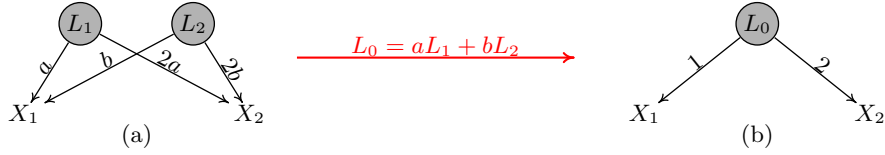
\begin{figure}[htp]
	\begin{center}
		\begin{tikzpicture}[scale=1.5, line width=0.5pt, inner sep=0.1mm, shorten >=.1pt, shorten <=.1pt]
		\draw (0.5, 0.8) node(i-L1) [circle, fill=gray!60,draw] {{\footnotesize\,$L_1$\,}};
		\draw (1.5, 0.8) node(i-L2) [circle, fill=gray!60,draw] {{\footnotesize\,$L_2$\,}};
		
		\draw (0, 0) node(i-X1) [] {{\footnotesize\,$X_1$\,}};
		\draw (2, 0) node(i-X2) [] {{\footnotesize\,$X_2$\,}};
		\draw[-arcsq] (i-L1) -- (i-X1) node[pos=0.5,sloped,above] {\footnotesize\,$a$};
		\draw[-arcsq] (i-L1) -- (i-X2)node[pos=0.5,sloped,above] {\footnotesize\,$2a$}; \draw[-arcsq] (i-L2) -- (i-X1) node[pos=0.5,sloped,above]{\footnotesize\,$b$};
		\draw[-arcsq] (i-L2) -- (i-X2)node[pos=0.5,sloped,above] {\footnotesize\,$2b$};
		
	    \draw (1,-0.2) node(label-ii) [] {{\footnotesize\,(a)\,}};
            \draw (2.4, 0.5) node(arrow1) [] {};
            \draw (4.9, 0.5) node(arrow2) [] {};
            \draw[-arcsq,red,thick] (arrow1) -- (arrow2) node[pos=0.5,sloped,above] {{\footnotesize\,$L_0=aL_1+bL_2$\,}};
		\draw (6.5, 0.8) node(i-L) [circle, fill=gray!60,draw] {{\footnotesize\,$L_0$\,}};
		
		\draw (5.5, 0) node(i-X1) [] {{\footnotesize\,$X_1$\,}};
		\draw (7.5, 0) node(i-X2) [] {{\footnotesize\,$X_2$\,}};
		
		\draw[-arcsq] (i-L) -- (i-X1) node[pos=0.5,sloped,above] {\footnotesize\,$1$};
		\draw[-arcsq] (i-L) -- (i-X2)node[pos=0.5,sloped,above] {\footnotesize\,$2$};
		
		\draw (6.5,-0.2) node(label-ii) [] {{\footnotesize\,(b)\,}};
		\end{tikzpicture}
		\caption{A specification of the causal model that violates the rank-faithfulness assumption. (a) There are two latent variables between $X_1$ and $X_2$. (b) There is one latent variable between $X_1$ and $X_2$.} 
        \vspace{-4mm}
		\label{fig:explain-assumption-rank-faithfulness} 
	\end{center}
\end{figure}
\end{Example-set}

\begin{Example-set}[Purity]
Consider the graph in Figure \ref{fig:simple-main-example}, and let $\mathcal{L}_1=\{L_2,L_3\}$ and $\mathbf{C}_1=\{X_{12},X_1,X_6\}$. $\mathbf{C}_1$ is a pure set relative to $\mathcal{L}_1$. However, if let $\mathcal{L}_1=\{L_1\}$ and $\mathbf{C}_1=\{X_{1},X_{6}\}$, then $\mathbf{C}_1$ is an impure set relative to $\mathcal{L}_1$, because $X_1 \nCI X_{6} | \mathcal{L}_1$. Notice that the \emph{pure variable} or \emph{impure variable} is a relative concept. For example, $L_4$ is an impure variable in $\{L_3,L_4,L_8\}$ relative to $L_1$,  while $L_4$ is a pure variable in $\{L_3,L_4\}$ relative to $L_1$.
\end{Example-set}

\begin{Example-set}[\textbf{Rule 1 }($\mathcal{R}1$)]
Consider the causal graph in Figure \ref{fig:merge-rules-one-cases} (a), where $\mathbf{C}_1=\{X_1,X_2\}$ and $\mathbf{C}_2=\{X_2,X_3\}$ are two impure and global causal clusters. We check the $\mathcal{R}1$ of Proposition \ref{Proposition-merge-rules-cases} as follows: Let $\mathcal{A}=X_{1:5}$. For a subset of $\mathbf{C}_1 \cup \mathbf{C}_2$, e.g., $\tilde{\mathbf{C}}_1=\{X_1,X_3\}$, $\neg_{\mathbf{C}_1}=\{X_5\}$ and $(\{X_5\},\{X_1, X_3\})$ follows the  GIN condition. The same conclusion will hold true for any other subsets of $\mathbf{C}_1 \cup \mathbf{C}_2$. Thus, we have $\mathbf{C}_1$ and $\mathbf{C}_2$ share the same parent, i.e., $L_1$.
\end{Example-set}

\begin{Example-set}[\textbf{Rule 2} ($\mathcal{R}2$)]
Consider the causal graph in Figure \ref{fig:merge-rules-one-cases} (b), where $\mathbf{C}_1=\{X_1, X_2, X_3\}$ and $\mathbf{C}_2=\{X_4, X_5\}$ are two pure causal clusters and have different size of latent parent variables. We check the $\mathcal{R}2$ of Proposition \ref{Proposition-merge-rules-cases} as follows: Let $\mathcal{A}=X_{1:5}$. For a subset of $\mathbf{C}_1$, e.g., $\tilde{\mathbf{C}}_1=\{X_1,X_2\}$, $V_i = X_4 \in \mathbf{C}_2$, $(\{X_3,X_5\},\{X_1,X_2, X_4\})$ follows the  GIN condition. The same conclusion will hold true for any other subset of $\mathbf{C}_1$ and any one variable of $\mathbf{C}_2$. Thus, we have the common parents of $\mathbf{C}_1$ contains the common parents of $\mathbf{C}_2$, i.e., $\{L_2\} \subset \{L_1, L_2\}$.
\end{Example-set}

\begin{figure}[htp]
		\centering
		\begin{tikzpicture}[scale=1.5, line width=0.5pt, inner sep=0.2mm, shorten >=.1pt, shorten <=.1pt]
        \draw [fill=blue!50,thick, fill opacity=0.5, draw=none] (0.25,0.6) ellipse [x radius=0.6cm, y radius=0.3cm];
        \draw [dashed, color=red, fill=red!50,thick, fill opacity=0.5, draw=none] (1.5,0.6) ellipse [x radius=0.65cm, y radius=0.3cm];
		\draw (0.9, 1.4) node(L1) [circle, fill=gray!60,draw] {{\footnotesize\,$L_1$\,}};
		
		\draw (-0.1, 0.6) node(X1) [] {{\footnotesize\,$X_{1}$\,}};
		\draw (0.5, 0.6) node(X2) [] {{\footnotesize\,$X_2$\,}};
		\draw (1.3, 0.6) node(X3) [] {{\footnotesize\,$X_3$\,}};
		\draw (1.9, 0.6) node(X4) [] {{\footnotesize\,$X_4$\,}};
		\draw (2.1, 1.4) node(X5) [] {{\footnotesize\,$X_5$\,}};
		\draw (0.3, 0.6) node(R21A) [] {{\,}};
		\draw (-0.1, 1.2) node(R21B) [] {{\footnotesize\,$\mathbf{C}_{1}$\,}};
        \draw[-arcsq, color=blue!30, thick] (R21B) to [in = 150, out = -80] (R21A);
		\draw (1.5, 0.6) node(R21C) [] {{\,}};
		\draw (1.9, 1.2) node(R21D) [] {{\footnotesize\,$\mathbf{C}_{2}$\,}};
        \draw[-arcsq, color=red!30, thick] (R21D) to [in = 40, out = -80] (R21C);
		\draw[-arcsq] (L1) -- (X1) node[pos=0.5,sloped,above] {};
		\draw[-arcsq] (L1) -- (X2) node[pos=0.5,sloped,above] {};
		\draw[-arcsq] (L1) -- (X3) node[pos=0.5,sloped,above] {};
		\draw[-arcsq] (L1) -- (X4) node[pos=0.5,sloped,above] {};
		\draw[-arcsq] (L1) -- (X5) node[pos=0.5,sloped,above] {};
		%
		\draw [->] (X1) edge[bend right=60] (X2);
		\draw [->] (X3) edge[bend right=60] (X4);
		\draw (1, 0) node(con1) [] {{\footnotesize\,(a) \,}};
		\end{tikzpicture}~~~~~~~~~~~~~
        \begin{tikzpicture}[scale=1.5, line width=0.5pt, inner sep=0.2mm, shorten >=.1pt, shorten <=.1pt]
        \draw [fill=blue!50,thick, fill opacity=0.5, draw=none] (0.55,0.6) ellipse [x radius=0.9cm, y radius=0.3cm];
        \draw [dashed, color=red, fill=red!50,thick, fill opacity=0.5, draw=none] (2.2,0.6) ellipse [x radius=0.65cm, y radius=0.3cm];
		\draw (0.7, 1.4) node(L1) [circle, fill=gray!60,draw] {{\footnotesize\,$L_1$\,}};
		\draw (1.8, 1.4) node(L2) [circle, fill=gray!60,draw] {{\footnotesize\,$L_2$\,}};

		\draw (-0.1, 0.6) node(X1) [] {{\footnotesize\,$X_{1}$\,}};
		\draw (0.5, 0.6) node(X2) [] {{\footnotesize\,$X_2$\,}};
		\draw (1.1, 0.6) node(X3) [] {{\footnotesize\,$X_3$\,}};
		\draw (1.9, 0.6) node(X4) [] {{\footnotesize\,$X_4$\,}};
		\draw (2.5, 0.6) node(X5) [] {{\footnotesize\,$X_5$\,}};
		\draw (0.3, 0.6) node(R21A) [] {{\,}};
		\draw (-0.1, 1.2) node(R21B) [] {{\footnotesize\,$\mathbf{C}_{1}$\,}};
        \draw[-arcsq, color=blue!30, thick] (R21B) to [in = 150, out = -80] (R21A);
		\draw (2.2, 0.6) node(R21C) [] {{\,}};
		\draw (2.5, 1.2) node(R21D) [] {{\footnotesize\,$\mathbf{C}_{2}$\,}};
        \draw[-arcsq, color=red!30, thick] (R21D) to [in = 40, out = -80] (R21C);
		\draw[-arcsq] (L1) -- (X1) node[pos=0.5,sloped,above] {};
		\draw[-arcsq] (L1) -- (X2) node[pos=0.5,sloped,above] {};
		\draw[-arcsq] (L1) -- (X3) node[pos=0.5,sloped,above] {};
		\draw[-arcsq] (L2) -- (X1) node[pos=0.5,sloped,above] {};
		\draw[-arcsq] (L2) -- (X2) node[pos=0.5,sloped,above] {};
		\draw[-arcsq] (L2) -- (X3) node[pos=0.5,sloped,above] {};
		\draw[-arcsq] (L2) -- (X4) node[pos=0.5,sloped,above] {};
		\draw[-arcsq] (L2) -- (X5) node[pos=0.5,sloped,above] {};
		\draw (1.2, 0) node(con1) [] {{\footnotesize\,(b) \,}};
		\end{tikzpicture}
		\vspace{-2mm}
		\caption{The illustrative example for $\mathcal{R}1$ and $\mathcal{R}2$ of Proposition \ref{Proposition-merge-rules-cases}. Graph (a) is an example
of $\mathcal{R}1$ and graph (b) is an example of $\mathcal{R}2$.}
		\label{fig:merge-rules-one-cases} 
\end{figure}

}
\revision{
\begin{Example-set}[\textbf{Rule 3} ($\mathcal{R}3$)]
Consider the causal graph in Figure \ref{fig:merge-rules-three-earlycurrent-cases} (a), where $\mathcal{L}_1=\{L_1\}$ is a latent variable set that is introduced in the first iteration, $\mathbf{C}_1=\{X_1,X_2\}$ is part of its children, and $\mathbf{C}_2=\{L_2,L_3\}$ is a new causal cluster. We check  $\mathcal{R}3$ of Corollary \ref{Corollary-merge-rules-earlycurrent-cases} as follows: Let $\mathcal{A}=\{L_1,L_2,L_3,X_7\}$. For a subset of $\mathbf{C}_1$, e.g., $\tilde{\mathbf{C}}_1=\{X_1\}$, and one node $V_i \in  \mathbf{C}_2$, e.g., $V_i=L_2$, we have $(\{X_7\},\{X_1, L_2\})$ follows the GIN condition. The same conclusion will hold true for any other subset of $\mathbf{C}_1$ and any variable in $\mathbf{C}_2$. This will imply that the latent parent of $\mathbf{C}_2$ is $\mathcal{L}_1$.
\end{Example-set}

\begin{Example-set}[\textbf{Rule 4} ($\mathcal{R}4$)]
Consider the causal graph in Figure \ref{fig:merge-rules-three-earlycurrent-cases} (b), where $\mathcal{L}_1=\{L_1,L_2\}$ is a latent variable set that is introduced in the first iteration, $\mathbf{C}_1=\{X_1,X_2,X_3\}$ is part of its children, and $\mathbf{C}_2=\{L_3,L_4\}$ is a new causal cluster. We check  $\mathcal{R}4$ of Corollary \ref{Corollary-merge-rules-earlycurrent-cases} as follows: Let $\mathcal{A}=\{L_1,L_3,L_4\}$. For a subset of $\mathbf{C}_1$, e.g., $\tilde{\mathbf{C}}_1=\{X_1,X_2\}$, and one node $V_i = L_3 \in \mathbf{C}_2$, $(\{L_4,X_3\},\{X_1,X_2, L_3\})$ follows the  GIN condition. The same conclusion will hold true for any other subset of $\mathbf{C}_1$ and any variable in $\mathbf{C}_2$. Thus, we have the common parents of $\mathbf{C}_1$ contains the common parents of $\mathbf{C}_2$, i.e., $L(\mathbf{C}_2) \subset \{L_1, L_2\}$.
\end{Example-set}

\begin{figure}[htp]
	\begin{center}
		\begin{tikzpicture}[scale=1.5, line width=0.5pt, inner sep=0.2mm, shorten >=.1pt, shorten <=.1pt]
        \draw [fill=blue!50,thick, fill opacity=0.5, draw=none] (0.0,0.7) ellipse [x radius=0.6cm, y radius=0.3cm];
        \draw [dashed, color=red, fill=red!50,rotate=25,thick, fill opacity=0.5, draw=none] (1.4,0.4) ellipse [x radius=0.6cm, y radius=0.3cm];
		\draw (0.6, 1.6) node(L1) [circle, fill=gray!60,draw] {{\footnotesize\,$L_1$\,}};
		\draw (-0.3, 0.7) node(X1) [] {{\footnotesize\,$X_{1}$\,}};
		\draw (0.3, 0.7) node(X2) [] {{\footnotesize\,$X_2$\,}};
		\draw (0.9, 0.8) node(L2) [] {{\footnotesize\,$L_2$\,}};
		\draw (1.4, 1.0) node(L3) [] {{\footnotesize\,$L_3$\,}};
		\draw (0.5, 0.0) node(X3) [] {{\footnotesize\,$X_3$\,}};
		\draw (1.1, 0.0) node(X4) [] {{\footnotesize\,$X_4$\,}};
		\draw (1.4, 0.2) node(X5) [] {{\footnotesize\,$X_5$\,}};
		\draw (1.9, 0.2) node(X6) [] {{\footnotesize\,$X_6$\,}};
		\draw (1.6, 1.9) node(X7) [] {{\footnotesize\,$X_7$\,}};
		\draw (0.0, 0.7) node(R21A) [] {{\,}};
		\draw (-0.3, 1.3) node(R21B) [] {{\footnotesize\,$\mathbf{C}_{1}$\,}};
        \draw[-arcsq, color=blue!30, thick] (R21B) to [in = 150, out = -80] (R21A);
		\draw (1.1, 0.9) node(R21C) [] {{\,}};
		\draw (1.4, 1.5) node(R21D) [] {{\footnotesize\,$\mathbf{C}_{2}$\,}};
        \draw[-arcsq, color=red!30, thick] (R21D) to [in = 40, out = -80] (R21C);
		\draw[-arcsq] (L1) -- (X1) node[pos=0.5,sloped,above] {};
		\draw[-arcsq] (L1) -- (L2) node[pos=0.5,sloped,above] {};
		\draw[-arcsq] (L1) -- (L3) node[pos=0.5,sloped,above] {};
		\draw[-arcsq] (L1) -- (X2) node[pos=0.5,sloped,above] {};
		\draw[-arcsq] (L1) -- (X7) node[pos=0.5,sloped,above] {};
		\draw[-arcsq] (L2) -- (X3) node[pos=0.5,sloped,above] {};
		\draw[-arcsq] (L2) -- (X4) node[pos=0.5,sloped,above] {};
		\draw[-arcsq] (L3) -- (X5) node[pos=0.5,sloped,above] {};
		\draw[-arcsq] (L3) -- (X6) node[pos=0.5,sloped,above] {};
		\draw [->] (X1) edge[bend right=60] (X2);
		\draw [->] (L2) edge[bend right=50] (L3);
		\draw (1, -0.4) node(con1) [] {{\footnotesize\,(a) \,}};
		\end{tikzpicture}~~~~~~~~~~~~~
        \begin{tikzpicture}[scale=1.5, line width=0.5pt, inner sep=0.2mm, shorten >=.1pt, shorten <=.1pt]
        \draw [fill=blue!50,thick, fill opacity=0.5, draw=none] (0.55,0.6) ellipse [x radius=0.9cm, y radius=0.3cm];
        \draw [dashed, color=red, fill=red!50,thick, fill opacity=0.5, draw=none] (2.2,0.6) ellipse [x radius=0.65cm, y radius=0.3cm];
		\draw (0.7, 1.4) node(L1) [circle, fill=gray!60,draw] {{\footnotesize\,$L_1$\,}};
		\draw (1.8, 1.4) node(L2) [circle, fill=gray!60,draw] {{\footnotesize\,$L_2$\,}};
		\draw (1.9, 0.6) node(L3) [] {{\footnotesize\,$L_3$\,}};
		\draw (2.5, 0.6) node(L4) [] {{\footnotesize\,$L_4$\,}};
		
		\draw (-0.1, 0.6) node(X1) [] {{\footnotesize\,$X_{1}$\,}};
		\draw (0.5, 0.6) node(X2) [] {{\footnotesize\,$X_2$\,}};
		\draw (1.1, 0.6) node(X3) [] {{\footnotesize\,$X_3$\,}};
		\draw (1.3, -0.2) node(X4) [] {{\footnotesize\,$X_4$\,}};
		\draw (1.9, -0.2) node(X5) [] {{\footnotesize\,$X_5$\,}};
		\draw (2.5, -0.2) node(X6) [] {{\footnotesize\,$X_6$\,}};
		\draw (3.1, -0.2) node(X7) [] {{\footnotesize\,$X_7$\,}};
		\draw (0.3, 0.6) node(R21A) [] {{\,}};
		\draw (-0.1, 1.2) node(R21B) [] {{\footnotesize\,$\mathbf{C}_{1}$\,}};
        \draw[-arcsq, color=blue!30, thick] (R21B) to [in = 150, out = -80] (R21A);
		\draw (2.2, 0.7) node(R21C) [] {{\,}};
		\draw (2.5, 1.2) node(R21D) [] {{\footnotesize\,$\mathbf{C}_{2}$\,}};
        \draw[-arcsq, color=red!30, thick] (R21D) to [in = 40, out = -80] (R21C);
		\draw[-arcsq] (L1) -- (X1) node[pos=0.5,sloped,above] {};
		\draw[-arcsq] (L1) -- (X2) node[pos=0.5,sloped,above] {};
		\draw[-arcsq] (L1) -- (X3) node[pos=0.5,sloped,above] {};
		\draw[-arcsq] (L2) -- (X1) node[pos=0.5,sloped,above] {};
		\draw[-arcsq] (L2) -- (X2) node[pos=0.5,sloped,above] {};
		\draw[-arcsq] (L2) -- (X3) node[pos=0.5,sloped,above] {};
		\draw[-arcsq] (L2) -- (L3) node[pos=0.5,sloped,above] {};
		\draw[-arcsq] (L2) -- (L4) node[pos=0.5,sloped,above] {};
		\draw[-arcsq] (L3) -- (X4) node[pos=0.5,sloped,above] {};
		\draw[-arcsq] (L3) -- (X5) node[pos=0.5,sloped,above] {};
		\draw[-arcsq] (L4) -- (X6) node[pos=0.5,sloped,above] {};
		\draw[-arcsq] (L4) -- (X7) node[pos=0.5,sloped,above] {};
		\draw (1.2, -0.6) node(con1) [] {{\footnotesize\,(b) \,}};
		\end{tikzpicture}
		\caption{An illustrative example for $\mathcal{R}3$ and $\mathcal{R}4$ of Corollary \ref{Corollary-merge-rules-earlycurrent-cases}. Graph (a) is an example
of $\mathcal{R}3$ and graph (b) is an example of $\mathcal{R}4$.}
            \vspace{-2mm}
		\label{fig:merge-rules-three-earlycurrent-cases}
	\end{center}
\end{figure}
}

\section{Further Allowing Causal Edges among Observed Variables}\label{Appendix-Subsec-infer-observed}

We notice that direct causal interactions may also occur among observed variables. Therefore, now the challenge lies in identifying the causal relationships within this context. In this section, we shift our attention to the specific causal connections between observed variables within the latent atomic structure of a LiNGLaH. Importantly, we show that the aforementioned proposed method can be readily extended to infer the entire causal structure, including specific causal relationships among observed variables.

It is worth noting that the independent noise (IN) condition has been used for discovering causal structures among observed variables in the linear non-Gaussian case, assuming the absence of latent confounders~\citep{shimizu2011directlingam}. Further recall that the IN condition represents a special case of GIN, where the variable set $\mathbf{Z}$ is a subset of $\mathbf{Y}$. In other words, $\mathbf{Y}$ and $\mathbf{Z}$ share certain measured variables (See Proposition \ref{Propo-IN-Special_Case}). Then, it is natural to leverage both the GIN condition and IN condition to estimate the causal structure involving latent variables and causal relations among the measured variables. 
Consequently, the key lies in appropriately allowing the two variable sets, $\mathbf{Y}$ and $\mathbf{Z}$, to share specific measured variables when testing for certain GIN conditions. 
Specifically, we can now identify the causal order between observed variables within the latent atomic structure of a LiNGLaH, as given in the following corollary.

\begin{Corollary}[Identifying Causal Order between Observed Variables]\label{Corollary-causal-direction-observed-variables-behind-confounders}
    Let $\mathbf{X}_p$ be a set of observed variables, and $X_q$ be an observed variable. Suppose $\{\mathbf{X}_{t},\mathcal{L}_t\}$ is the set of confounders of $\mathbf{X}_p$ and $X_q$, where $\mathbf{X}_{t}$ consists of the observed confounders and $\mathcal{L}_{t}$ consists of the latent confounders.
    Let $\mathbf{T}_1$ and $\mathbf{T}_2$ be two sets that contain $|\mathcal{L}'_t|$ pure children of each latent variable set $\mathcal{L}'_t$ in $\mathcal{L}_t$, and $\mathbf{T}_1 \cap \mathbf{T}_2 =\emptyset$. Then if $(\{\mathbf{X}_p,\mathbf{X}_t,\mathbf{T}_2\},\{\mathbf{X}_p,X_q,\mathbf{X}_t,\mathbf{T}_1\})$ follows the GIN condition, $\mathbf{X}_p$ is causally earlier than $X_q$ (denoted by $\mathbf{X}_p \succ X_q$).
\end{Corollary}

Specifically, if $\mathbf{X}_t$, $\mathbf{T}_1$, and $\mathbf{T}_2$ are empty sets, then it is just the original GIN condition, which says that the regression residue of regressing $\mathbf{X}_p$ on $X_q$ is independent of $\mathbf{X}_p$. 
Below, we provide an example to illustrate this case.

\begin{Example-set}
Consider the causal graphs in Figure \ref{fig:discussion}(a), where $\mathbf{C}_t=\{X_{1:4}\}$ is an impure cluster. Suppose $X_p=X_3$ and $X_q=X_4$. Then the set of confounders $\{\mathbf{X}_t,\mathcal{L}_t\}=\{L_1\}$. Let $\mathbf{T}_1=\{X_1\}$ and $\mathbf{T}_2=\{X_2\}$.
According to Corollary \ref{Corollary-causal-direction-observed-variables-behind-confounders}, $(\{X_2,X_3\},\{X_1,X_3,X_4\})$ follows the GIN condition. This implies that $X_3 \succ X_4$.
\end{Example-set}

After identifying the causal order over a set of variables, we next show how to remove redundant edges between observed variables.
\begin{Corollary}[Removing Redundant Edges between Observed Variables]\label{Corollary-removing-redundant-edges-observed}
Let $X_p$ and $X_q$ be two observed variables in an impure cluster $\mathbf{C}_i$. 
Suppose $X_p$ is causally earlier than $X_q$.
Let $\mathcal{L}_t$ be the common parents of $\mathbf{C}_i$ and $\mathbf{X}_\mathbf{S}=\{X_{\mathbf{S}_1},...,X_{\mathbf{S}_s}\}$ be the set of observed variables in $\mathbf{C}_i$ such that each variable $X_{\mathbf{S}_i}$ is causally later than $X_p$ and is causally earlier than $X_q$. Furthermore, let $\{\mathbf{T}_1,\mathbf{T}_2\}$ be pure children of $\mathcal{L}_t$ with $|\mathbf{T}_1|=|\mathbf{T}_2|=|\mathcal{L}_t|$.
Then $X_p$ and $X_q$ are d-separated by $\mathbf{X}_\mathbf{S}\cup \mathcal{L}_t$, i.e., there is no directed edge between $X_p$ and $X_q$ iff the rank of the cross-covariance matrix of $\{\mathbf{P}_1,\mathbf{Q}_1\} \cup \{\mathbf{T}_1,\mathbf{T}_2\} \cup \mathbf{X}_\mathbf{S}$ is less than or equal to $|\mathcal{L}_t \cup \mathbf{X}_\mathbf{S}|$.
\end{Corollary}

Based on the above analysis, we need to slightly modify the original LaHiCaSl algorithm to identify the causal structure between the observed variables. It is important to note that the relationships between observed variables do not affect the localization of latent variables in the original algorithm, specifically in Phase I. This is because in Proposition 3, we do not impose any restrictions on the causal relationships between variables in the active variable set $\mathcal{A}$. Therefore, we only need to update Algorithm 6 in Phase II. Specifically, in addition to the causal structure learning among latent variables, we also need to further determine the causal direction of observed variables according to Corollary \ref{Corollary-causal-direction-observed-variables-behind-confounders} and remove redundant edges in each impure cluster according to Corollary \ref{Corollary-removing-redundant-edges-observed}. The modified algorithm, which allows causal edges among observed variables within a latent atomic structure, is shown in Algorithm \ref{algorithm-locally-infer-causalorder-plus} (LocallyInferCausalStructure$^{+}$), with its correctness shown in the following theorem.

\begin{algorithm}[htb]
	\caption{LocallyInferCausalStructure$^{+}$}
	\label{algorithm-locally-infer-causalorder-plus}
	\begin{algorithmic}[1]
	\REQUIRE
	A set of observed variables $\mathbf{X}$ and partial structure $\mathcal{G}$\\
	\REPEAT
			\STATE Select an impure cluster $\mathbf{C}_i$ from $\mathcal{G}$;
			\STATE Initialize latent confounder set: $\mathbf{LC}= \emptyset$ and observed confounder set: $\mathbf{OC}= \emptyset$;
			\STATE Add the common parent set $\mathcal{L}_t$ of $\mathbf{C}_i$ into $\mathbf{LC}$;
			\WHILE{$|\mathbf{C}_i| > 1$}
                \IF{The elements of $\mathbf{C}_{i}$ are observed variables}
                \STATE Find a local root variable ${X}_{r}$ according to Corollary \ref{Corollary-causal-direction-observed-variables-behind-confounders};
			\STATE $\mathbf{C}_i=\mathbf{C}_i \backslash {X}_r$ and add ${X}_r$ into $\mathbf{OC}$;
			\STATE ${\mathcal{G}}={\mathcal{G}} \cup \{{X}_{r} \succ {X}_{i} | {X}_{i} \in \mathbf{C}_i\}$;
                \ELSE
			\STATE Find a local root variable $\mathcal{L}_{r}$ according to Proposition \ref{Proposition-causa-direction-behind-confounders};
			\STATE $\mathbf{C}_i=\mathbf{C}_i \backslash \mathcal{L}_r$ and add $\mathcal{L}_r$ into $\mathbf{LC}$;
			\STATE ${\mathcal{G}}={\mathcal{G}} \cup \{\mathcal{L}_{r} \succ \mathcal{L}_{i} | \mathcal{L}_{i} \in \mathbf{C}_i\}$;
                \ENDIF
			\ENDWHILE
			\REPEAT
                \IF{The elements of $\mathbf{C}_{i}$ are observed variables}
                \STATE Select an ordered pair of variables $X_p$ and $X_q$ in $\mathbf{C}_i$ with $X_p \succ X_q$;
			\IF{there exists set $\mathbf{X}_\mathbf{S} \subset \mathbf{C}_1$ such that each variable is causally later than $X_p$ and is causally earlier than $X_q$, and the conditions in Corollary \ref{Corollary-removing-redundant-edges-observed} hold.}
			\STATE Remove the directed edge between $X_p$ and $X_q$.
                \ENDIF
                \ELSE
			\STATE Select an ordered pair of variables $\mathcal{L}_p$ and $\mathcal{L}_q$ in $\mathbf{C}_i$ with $\mathcal{L}_p \succ \mathcal{L}_q$;
			\IF{there exists set $\mathcal{L}_\mathbf{S} \subset \mathbf{C}_1$ such that each latent variable set is causally later than $\mathcal{L}_p$ and is causally earlier than $\mathcal{L}_q$, and the conditions in Proposition \ref{Proposition-removing-redundant-edges} hold.}
			\STATE Remove the directed edge between $\mathcal{L}_p$ and $\mathcal{L}_q$.
			\ENDIF
                \ENDIF
			\UNTIL{All ordered pairs of variables in $\mathbf{C}_i$ selected}
			\UNTIL{All impure clusters in $\mathcal{G}$ selected}
	\ENSURE
	Fully identified structure $\mathcal{G}$
	\end{algorithmic}
\end{algorithm}

\begin{Theorem}\label{Theorem-Model-additional-Identification}
Suppose that the input data $\mathbf{X}$ follows LiNGLaH with a minimal latent hierarchical structure. Further, suppose that the causal connections between observed variables only exist in the latent atomic structure of a LiNGLaH. Given infinite samples, the LaHiCaSl algorithm with Algorithm \ref{algorithm-locally-infer-causalorder-plus} will output the true causal structure, including the causal relationships among observed variables, those between the observed variables and their corresponding latent variable sets, and the causal relationships between the latent variable sets.
\end{Theorem}

\begin{figure}[htp]
 \vspace{1mm}
	\label{fig:exp}
	\begin{center}
		\begin{tikzpicture}[scale=0.6]
		\begin{axis}[
		title=(a),
		title style={at={(0.5,-0.6)}},
		width=5.5cm,
		cycle list name=mark list,
		ybar,
		enlargelimits=0.15,
        ymin=0.2,ymax=0.8,
		legend style={at={(0.65,0.2)}, anchor=north,legend columns=-1},
		ylabel={Correct ordering rate},
		xlabel={Sample size},
		symbolic x coords={100,250,500,1k},
		xtick=data,
		nodes near coords,
		nodes near coords align={vertical},
		]
		\addplot [ fill=purple,postaction={pattern=north east lines}] coordinates {(100,0.5) (250,0.54) (500,0.59) (1k,0.68)};
		\addplot [fill=green,postaction={pattern=dots}] coordinates {(100,0.49) (250,0.5) (500,0.5) (1k,0.53)};
		\legend{LaHiCaSl,LSTC}
		\end{axis}
		\end{tikzpicture}~~
		\begin{tikzpicture}[scale=0.6]
		\begin{axis}[
		title=(b),
		title style={at={(0.5,-0.6)}},
		width=5.5cm,
		cycle list name=mark list,
		ybar,
		enlargelimits=0.15,
        ymin=0.2,ymax=0.8,
		legend style={at={(0.65,0.2)}, anchor=north,legend columns=-1},
		ylabel={Correct ordering rate},
		xlabel={Sample size},
		symbolic x coords={100,250,500,1k},
		xtick=data,
		nodes near coords,
		nodes near coords align={vertical},
		]
		\addplot [ fill=purple,postaction={pattern=north east lines}] coordinates {(100,0.5) (250,0.54) (500,0.58) (1k,0.65)};
		\addplot [fill=green,postaction={pattern=dots}] coordinates {(100,0.0) (250,0.0) (500,0.0) (1k,0.0)};
		\legend{LaHiCaSl,LSTC}
		\end{axis}
		\end{tikzpicture}
		\begin{tikzpicture}[scale=0.6]
		\begin{axis}[
		title=(c),
		title style={at={(0.5,-0.6)}},
		width=5.5cm,
		cycle list name=mark list,
		ybar,
		enlargelimits=0.15,
        ymin=0.2,ymax=0.8,
		legend style={at={(0.65,0.2)}, anchor=north,legend columns=-1},
		ylabel={Correct ordering rate},
		xlabel={Sample size},
		symbolic x coords={100,250,500,1k},
		xtick=data,
		nodes near coords,
		nodes near coords align={vertical},
		]
		\addplot [ fill=purple,postaction={pattern=north east lines}] coordinates {(100,0.5) (250,0.54) (500,0.57) (1k,0.66)};
		\addplot [fill=green,postaction={pattern=dots}] coordinates {(100,0.43) (250,0.43) (500,0.47) (1k,0.52)};
		\legend{LaHiCaSl,LSTC}
		\end{axis}
		\end{tikzpicture}~~
		\begin{tikzpicture}[scale=0.6]
		\begin{axis}[
		title=(d),
		title style={at={(0.5,-0.6)}},
		width=5.5cm,
		cycle list name=mark list,
		ybar,
		enlargelimits=0.15,
        ymin=0.2,ymax=0.8,
		legend style={at={(0.65,0.2)}, anchor=north,legend columns=-1},
		ylabel={Correct ordering rate},
		xlabel={Sample size},
		symbolic x coords={100,250, 500,1k},
		xtick=data,
		nodes near coords,
		nodes near coords align={vertical},
		]
		\addplot [ fill=purple,postaction={pattern=north east lines}] coordinates {(100,0.55) (250,0.55) (500,0.58) (1k,0.62)};
		\addplot [fill=green,postaction={pattern=dots}] coordinates {(100,0.25) (250,0.25) (500,0.25) (1k,0.34)};
		\legend{LaHiCaSl,LSTC}
		\end{axis}
		\end{tikzpicture}\\
		\begin{tikzpicture}[scale=0.6]
		\begin{axis}[
		title=(e),
		title style={at={(0.5,-0.6)}},
		width=5.5cm,
		cycle list name=mark list,
		ybar,
		enlargelimits=0.15,
        ymin=0.2,ymax=0.8,
		legend style={at={(0.65,0.2)}, anchor=north,legend columns=-1},
		ylabel={Correct ordering rate},
		xlabel={Sample size},
		symbolic x coords={100, 250, 500,1k},
		xtick=data,
		nodes near coords,
		nodes near coords align={vertical},
		]
		\addplot [ fill=purple,postaction={pattern=north east lines}] coordinates {(100,0.3) (250,0.39) (500,0.5) (1k,0.59)};
		\addplot [fill=green,postaction={pattern=dots}] coordinates {(100,0.0) (250,0.0) (500,0.0) (1k,0.0)};
		\legend{LaHiCaSl,LSTC}
		\end{axis}
		\end{tikzpicture}~~
		\begin{tikzpicture}[scale=0.6]
		\begin{axis}[
		title=(f),
		title style={at={(0.5,-0.6)}},
		width=5.5cm,
		cycle list name=mark list,
		ybar,
		enlargelimits=0.15,
        ymin=0.2,ymax=0.8,
		legend style={at={(0.65,0.2)}, anchor=north,legend columns=-1},
		ylabel={Correct ordering rate},
		xlabel={Sample size},
		symbolic x coords={100,250, 500,1k},
		xtick=data,
		nodes near coords,
		nodes near coords align={vertical},
		]
		\addplot [ fill=purple,postaction={pattern=north east lines}] coordinates {(100,0.33) (250,0.34) (500,0.46) (1k,0.55)};
		\addplot [fill=green,postaction={pattern=dots}] coordinates {(100,0.0) (250,0.0) (500,0.0) (1k,0.0)};
		\legend{LaHiCaSl,LSTC}
		\end{axis}
		\end{tikzpicture}~~
		\begin{tikzpicture}[scale=0.6]
		\begin{axis}[
		title=(g),
		title style={at={(0.5,-0.6)}},
		width=5.5cm,
		cycle list name=mark list,
		ybar,
		enlargelimits=0.15,
        ymin=0.2,ymax=0.8,
		legend style={at={(0.65,0.2)}, anchor=north,legend columns=-1},
		ylabel={Correct ordering rate},
		xlabel={Sample size},
		symbolic x coords={100, 250, 500,1k},
		xtick=data,
		nodes near coords,
		nodes near coords align={vertical},
		]
		\addplot [ fill=purple,postaction={pattern=north east lines}] coordinates {(100,0.38) (250,0.38) (500,0.49) (1k,0.57)};
		\addplot [fill=green,postaction={pattern=dots}] coordinates {(100,0.0) (250,0.0) (500,0.0) (1k,0.0)};
		\legend{LaHiCaSl,LSTC}
		\end{axis}
		\end{tikzpicture}~~
		\begin{tikzpicture}[scale=0.6]
		\begin{axis}[
		title=(h),
		title style={at={(0.5,-0.6)}},
		width=5.5cm,
		cycle list name=mark list,
		ybar,
		enlargelimits=0.15,
        ymin=0.2,ymax=0.8,
		legend style={at={(0.65,0.2)}, anchor=north,legend columns=-1},
		ylabel={Correct ordering rate},
		xlabel={Sample size},
		symbolic x coords={100, 250, 500,1k},
		xtick=data,
		nodes near coords,
		nodes near coords align={vertical},
		]
		\addplot [ fill=purple,postaction={pattern=north east lines}] coordinates {(100,0.32) (250,0.36) (500,0.43) (1k,0.52)};
		\addplot [fill=green,postaction={pattern=dots}] coordinates {(100,0.0) (250,0.0) (500,0.0) (1k,0.0)};
		\legend{LaHiCaSl,LSTC}
		\end{axis}
		\end{tikzpicture}
		\caption{ (a-h) Accuracy of the estimated causal order with LaHiCaSl (purple) and LSTC (green) for Cases 1-8 (the higher the better).}
		\vspace{-0.4cm}
		\label{fig:F1-scores-sample-size}
	\end{center}
\end{figure}
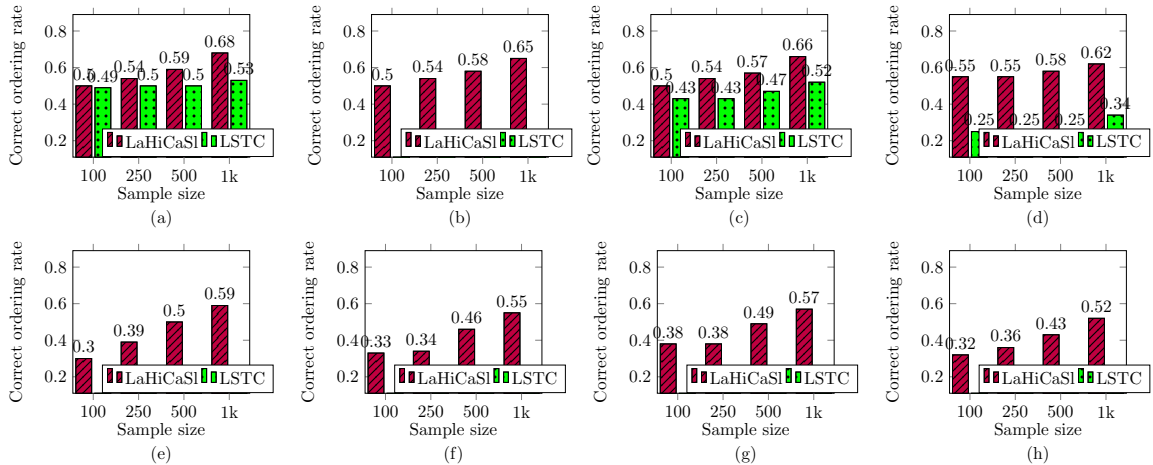

\revision{
\section{More Experimental Results}
\subsection{Small-Sample Experimental Results} \label{Subsec:samll-sample}
In this section, we evaluate the algorithm's performance with small sample sizes. We maintain the experimental settings outlined in Section \ref{Sec-Experiment}, with the only modification being the adjustment of sample sizes to 100, 250, 500, and 1,000. 

The results are presented in Tables \ref{tab:compare-case1-4-small-size} and \ref{tab:compare-case5-8-small-size} and Figure \ref{fig:F1-scores-sample-size}.
We observed that with a sample size of 100, both our method and the comparative methods encountered challenges, resulting in unsatisfactory results. However, as the sample size increased to 1,000, the performance of all methods improved significantly, with our approach showing a significant advantage.
}
\begin{center}
\begin{table}[htp!]
	\small
	\center \caption{Performance of LaHiCaSl, LSTC, FOFC, and BPC on learning measured-based latent structure (the lower the better).}
 \vspace{3mm}
	\label{tab:compare-case1-4-small-size}
	\resizebox{\textwidth}{!}{
	\begin{tabular}{|c|c|c|c|c|c|c|c|c|c|c|c|c|c|}
		\hline  \multicolumn{2}{|c|}{} &\multicolumn{4}{|c|}{\textbf{Latent omission}} & \multicolumn{4}{|c|}{\textbf{Latent commission}} & \multicolumn{4}{|c|}{\textbf{Mismeasurements}}\\
		\hline 
		\multicolumn{2}{|c|}{Algorithm} & LaHiCaSl & LSTC & FOFC & BPC & LaHiCaSl & LSTC & FOFC & BPC & LaHiCaSl & LSTC & FOFC & BPC \\
	\hline 
		 & 100 & 0.18(15) & 0.50(50) & 1.00(50) & 1.00(50) 
		 & 0.15(12) & 0.50(50) & 0.00(0) & 0.00(0) 
		 & 0.10(15) & 0.60(50) & 0.00(0) & 0.00(0) \\
           \cline{2-14}
		 & 250 & 0.14(13) & 0.50(50) & 1.00(50) & 1.00(50) 
		 & 0.09(10) & 0.50(50) & 0.00(0) & 0.00(0) 
		 & 0.09(10) & 0.60(50) & 0.00(0) & 0.00(0) \\
		\cline{2-14}
		{\emph{Case 1}} &500 &  0.12(12) &0.50(50) & 1.00(50) & 
            1.00(50) 
		& 0.10(10) & 0.50(50) & 0.00(0) & 0.00(0) 
		& 0.10(10) & 0.60(50)& 0.00(0) & 0.00(0) \\
		\cline{2-14}
		& 1k & 0.09(15) & 0.49(49) & 0.90(50) & 0.94(50) 
		& 0.05(10) & 0.50(50)& 0.43(20) & 0.55(34) 
		& 0.08(10) & 0.59(50) & 0.66(29) & 0.76(25) \\
		\hline 
		 & 100 
		 & 0.26(25) & 0.50(50) & 1.00(50) & 1.00(50) 
		 & 0.20(15) & 0.00(0) & 0.00(0) & 0.00(0)
		 & 0.25(15) & 0.00(0) & 0.00(0) & 0.00(0)  \\
             \cline{2-14}
		 & 250 & 0.26(20) & 0.50(50) & 1.00(50) & 1.00(50) 
		 & 0.15(15) & 0.00(0) & 0.00(0) & 0.00(0) 
		 & 0.15(15) & 0.00(0) & 0.00(0) & 0.00(0) \\
		\cline{2-14}
		{\emph{Case 2}} &500 
		& 0.22(20) & 0.50(50) & 0.88(50) & 0.90(50) 
		& 0.13(10) & 0.00(0) & 0.00(0) & 0.00(0) 
		& 0.15(10) & 0.00(0) & 0.45(30) & 0.68(49)\\
		\cline{2-14} &1k
		& 0.16(10) & 0.46(45) & 0.79(50) & 0.83(50)
		& 0.10(09) & 0.00(0) & 0.00(0) & 0.00(0)
		& 0.11(09) & 0.00(0) & 0.40(30)  & 0.66(38)\\
		\hline 
		 & 100 
		 & 0.24(20) & 0.67(50) & 1.00(50) & 1.00(50) 
		 & 0.12(10) & 0.35(50) & 0.00(0) & 0.00(0) 
		 & 0.10(10) & 0.00(0) & 0.00(0) & 0.00(0)\\
             \cline{2-14}
		 & 250 & 0.20(20) & 0.67(50) & 1.00(50) & 1.00(50) 
		 & 0.10(12) & 0.33(50) & 0.00(0) & 0.00(0) 
		 & 0.10(16) & 0.00(0) & 0.00(0) & 0.00(0) \\
		\cline{2-14}
		{\emph{Case 3}} &500 
		& 0.15(15) & 0.67(50) & 0.78(50) & 0.77(50) 
		 & 0.12(10) & 0.33(50) & 0.73(48) & 0.76(40) 
		 & 0.10(10) & 0.00(0) & 0.55(50) & 0.60(50)\\
		\cline{2-14}&1k 
		& 0.10(10) & 0.65(50) & 0.69(50) & 0.71(50) 
		 & 0.09(09) & 0.33(50) & 0.67(50) & 0.69(50) 
		 & 0.09(08) & 0.00(0) & 0.49(50) & 0.55(50)\\
		\hline 
		 & 100 
		 &0.25(25) &0.75(50) & 1.00(50) & 1.00(50)
		 &0.20(18) &0.25(50) & 0.00(0) & 0.00(0)
		 &0.20(16) &0.00(0) & 0.00(0) & 0.00(0)\\
             \cline{2-14}
		 & 250 & 0.18(18) & 0.75(50) & 1.00(50) & 1.00(50) 
		 & 0.12(10) & 0.25(50) & 0.00(0) & 0.00(0) 
		 & 0.14(11) & 0.00(0) & 0.00(0) & 0.00(0) \\
		\cline{2-14}
		{\emph{Case 4}} &500
	     &0.14(12) &0.75(50) & 1.00(50) & 1.00(50)
		 &0.10(10) &0.25(50) & 0.00(0) & 0.00(0)
		 &0.08(08) &0.00(0) & 0.00(0) & 0.00(0)\\
		\cline{2-14}
		&1k 
		 &0.10(10) &0.74(50) & 0.90(50) & 0.87(50)
		 &0.10(10) &0.25(50) & 0.00(0) & 0.00(0)
		 &0.08(06) &0.00(0) & 0.13(13) & 0.10(14)\\   
		\hline 
	\end{tabular}}
	\begin{tablenotes}
		\item Note: The number in parentheses indicates the number of occurrences that the current algorithm {\it cannot} correctly solve the problem.
	\end{tablenotes}
	\vspace{-5mm}
\end{table}
\end{center}

\begin{center}
\begin{table}[htp!]
	\small
	\center \caption{Performance of LaHiCaSl, LSTC, CLRG, CLNJ, FOFC, and BPC on learning latent hierarchical structure (the lower the better).}
 \vspace{3mm}
	\label{tab:compare-case5-8-small-size}
	\resizebox{15cm}{!}{
	\begin{tabular}{|c|c|c|c|c|c|c|c|c|c|c|c|c|c|}
		\hline  \multicolumn{2}{|c|}{} &\multicolumn{6}{|c|}{\textbf{Structure Recovery Error Rate}} & \multicolumn{6}{|c|}{\textbf{Error in Hidden Variables}}\\
		\hline 
		\multicolumn{2}{|c|}{Algorithm} & LaHiCaSl & LSTC & CLRG & CLNJ & FOFC & BPC & LaHiCaSl & LSTC & CLRG & CLNJ & FOFC & BPC \\
		\hline 
		 & 100 & 0.84 & 1.0 & 1.0 & 1.0 & 1.0 & 1.0 
		 & 1.4 & 5.0 & 6.0 & 6.0 & 6.0 & 6.0\\
            \cline{2-14}
            & 250 & 0.78 & 1.0 & 1.0 & 1.0 & 1.0 & 1.0 
		 & 1.1 & 5.0 & 6.0 & 6.0 & 6.0 & 6.0\\
		\cline{2-14}
		{\emph{Case 5}} & 500 & 0.72 & 1.0 & 1.0 & 1.0 & 1.0 & 1.0 
		 & 0.8 & 5.0 & 6.0 & 6.0 & 6.0 & 6.0\\
		\cline{2-14}
		& 1k & 0.33 & 1.0 & 1.0 & 1.0 & 1.0 & 1.0 
		 & 0.5 & 4.9 & 6.0 & 6.0 & 5.7 & 5.7\\
		\hline 
		 & 100 & 0.90 & 1.0 & 1.0 & 1.0 & 1.0 & 1.0 
		 & 2.6 & 6.0 & 7.0 & 7.0 & 7.0 & 7.0\\
             \cline{2-14}
            & 250 & 0.88 & 1.0 & 1.0 & 1.0 & 1.0 & 1.0 
		 & 2.2 & 6.0 & 7.0 & 7.0 & 7.0 & 7.0\\
		\cline{2-14}
		{\emph{Case 6}} & 500 & 0.80 & 1.0 & 1.0 & 1.0 & 1.0 & 1.0 
		 & 1.6 & 6.0 & 7.0 & 7.0 & 7.0 & 7.0\\
		\cline{2-14} & 1k & 0.42 & 1.0 & 1.0 & 1.0 & 1.0 & 1.0 
		 & 1.1 & 5.9 & 7.0 & 7.0 & 6.6 & 6.8\\
		\hline 
		 & 100 & 0.88 & 1.0 & 1.0 & 1.0 & 1.0 & 1.0 
		 & 5.4 & 9.0 & 10.0 & 10.0 & 10.0 & 10.0\\
             \cline{2-14}
            & 250 & 0.88 & 1.0 & 1.0 & 1.0 & 1.0 & 1.0 
		 & 5.0 & 9.9 & 10.0 & 10.0 & 10.0 & 10.0\\
		\cline{2-14}
		{\emph{Case 7}} & 500 & 0.76 & 1.0 & 1.0 & 1.0 & 1.0 & 1.0 
		 & 4.4 & 9.0 & 10.0 & 10.0 & 10.0 & 10.0\\
		\cline{2-14}& 1k & 0.37 & 1.0 & 1.0 & 1.0 & 1.0 & 1.0 
		 & 4.3 & 9.0 & 10.0 & 10.0 & 9.8 & 9.8\\
		\hline 
		 & 100 & 1.0 & 1.0 & 1.0 & 1.0 & 1.0 & 1.0 
		 & 6.8 & 8.8 & 9.0 & 9.0 & 9.0 & 9.0\\
             \cline{2-14}
            & 250 & 0.98 & 1.0 & 1.0 & 1.0 & 1.0 & 1.0 
		 & 6.6 & 8.0 & 9.0 & 9.0 & 9.0 & 9.0\\
		\cline{2-14}
		{\emph{Case 8}} & 500 & 0.88 & 1.0 & 1.0 & 1.0 & 1.0 & 1.0 
		 & 6.0 & 8.0 & 9.0 & 9.0 & 9.0 & 9.0\\
		\cline{2-14}
		& 1k & 0.50 & 1.0 & 1.0 & 1.0 & 1.0 & 1.0 
		 & 4.6 & 7.9 & 9.0 &9.0 & 8.5 & 8.8\\   
		\hline 
	\end{tabular}}
\end{table}
\end{center}

\revision{
\subsection{Random Latent Structure Experimental Results}
Here, we analyze the performance of algorithms under randomly generated latent structures. Specifically, we distinguish between Measurement-based structures and latent hierarchical structures. To ensure a fair comparison, we set each latent variable model to have three measurement variables. This model satisfies the assumptions of the BPC and FOFC algorithms under the Measurement-based model. The sample size for all data is set to 5k.

The results are given in Tables \ref{tab:compare-case1-4-randomly-graph} and \ref{tab:compare-case5-8-randomly-graph}, respectively. We observed that our algorithm still outperforms the other comparison methods overall, demonstrating its ability to identify both measurement-based and latent hierarchical structures effectively.

\begin{center}
\begin{table}[htp!]
	\small
	\center \caption{Performance of LaHiCaSl, LSTC, FOFC, and BPC on learning measured-based randomly generated latent structure (the lower the better).}
	 \vspace{3mm}
	\label{tab:compare-case1-4-randomly-graph}
	\resizebox{\textwidth}{!}{
	\begin{tabular}{|c|c|c|c|c|c|c|c|c|c|c|c|c|}
		\hline  \multicolumn{1}{|c|}{} &\multicolumn{4}{|c|}{\textbf{Latent omission}} & \multicolumn{4}{|c|}{\textbf{Latent commission}} & \multicolumn{4}{|c|}{\textbf{Mismeasurements}}\\
		\hline 
		\multicolumn{1}{|c|}{Algorithm} & LaHiCaSl & LSTC & FOFC & BPC & LaHiCaSl & LSTC & FOFC & BPC & LaHiCaSl & LSTC & FOFC & BPC \\
		\hline 
		3 LVs(9 MVs) & 0.08(08) & 0.11(09) & 0.12(10) & 0.14(12) 
		 & 0.02(02) & 0.04(03) & 0.10(10) & 0.14(12) 
		 & 0.06(04) & 0.08(05) & 0.06(06) & 0.03(02) \\
		\cline{1-13}
	   5 LVs(15 MVs) &  0.10(08) &0.14(07) & 0.16(12) & 0.18(13) 
		& 0.06(03) & 0.05(03) & 0.12(10) & 0.08(08) 
		& 0.08(04) & 0.10(04)& 0.11(08) & 0.06(04) \\
		\cline{1-13}
	   10 LVs(30 MVs) & 0.13(10) & 0.20(12) & 0.24(16) & 0.26(16) 
		& 0.10(06) & 0.13(08)& 0.14(13) & 0.16(15) 
		& 0.08(06) & 0.18(08) & 0.16(07) & 0.12(10) \\
		\hline 
	\end{tabular}
}
	\begin{tablenotes}
		\item Note: The number in parentheses indicates the number of occurrences that the current algorithm {\it cannot} correctly solve the problem. LVs denote latent variables and MVs denote measurement variables.
	\end{tablenotes}
	\vspace{-5mm}
\end{table}
\end{center}

\begin{center}
\begin{table}[htp!]
	\small
	\center \caption{Performance of LaHiCaSl, LSTC, CLRG, CLNJ, FOFC, and BPC on learning randomly generated latent hierarchical structure (the lower the better).}
 \vspace{3mm}
	\label{tab:compare-case5-8-randomly-graph}
	\resizebox{15cm}{!}{
	\begin{tabular}{|c|c|c|c|c|c|c|c|c|c|c|c|c|}
		\hline  \multicolumn{1}{|c|}{} &\multicolumn{6}{|c|}{\textbf{Structure Recovery Error Rate}} & \multicolumn{6}{|c|}{\textbf{Error in Hidden Variables}}\\
		\hline 
		\multicolumn{1}{|c|}{Algorithm} & LaHiCaSl & LSTC & CLRG & CLNJ & FOFC & BPC & LaHiCaSl & LSTC & CLRG & CLNJ & FOFC & BPC \\
		\hline 
		 $4$ LVs (9 MVs) & 0.10 & 1.00 & 1.00 & 1.00 & 1.00 & 1.00 
		 & 0.10 & 3.20 & 3.00 & 3.00 & 0.84 & 0.92\\
		\cline{1-13}
	   $6\sim7$ LVs (15 MVs) & 0.16 & 1.00 & 1.00 & 1.00 & 1.00 & 1.00 
		 & 0.32 & 4.12 & 5.00 & 5.00 & 3.00 & 3.20\\
		\cline{1-13}
	   $11 \sim 13$ LVs (30 MVs) & 0.24 & 1.00 & 1.00 & 1.00 & 1.00 & 1.00 
		 & 0.38 & 6.38 & 10.0 & 10.0 & 7.68 & 7.88\\
		\hline 
	\end{tabular}}
    \begin{tablenotes}
	\item Note: LVs denote latent variables and MVs denote measurement variables.
	\end{tablenotes}
\end{table}
\end{center}
}

\revision{
\section{Future Work}
One of our future research directions is to apply GIN to address more general cases, e.g., eliminating the measurement assumption (no observed variables affect latent variables). For instance, consider the two structures in Figure \ref{fig:furture-work}, neither structure (a) nor (b) induces independence constraints over the observed variables. Interestingly, one can find that these two structures entail different GIN conditions over the observed variables when the variables satisfy a linear non-Gaussian model. Specifically, for (a), $(\{X_{1}\},\{X_{3},X_{4}\})$ satisfies the GIN condition while $(\{X_{2}\},\{X_{3},X_{4}\})$ violates the 
GIN condition, and for (b), both $(\{X_{1}\},\{X_{3}, X_{4}\})$ and $(\{X_{2}\},\{X_{3}, X_{4}\})$ satisfy the GIN condition, which means that there must exist a latent variable between $\{X_1, X_2\}$ and $\{X_3, X_4\}$, and $\{X_1, X_2\} \to L_1$.

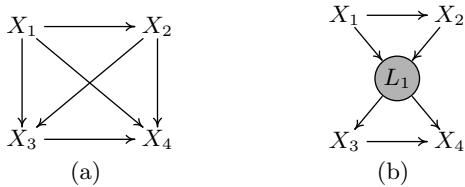
\begin{figure}[htp]
	\begin{center}
	    \begin{tikzpicture}[scale=1.5, line width=0.5pt, inner sep=0.2mm, shorten >=.1pt, shorten <=.1pt]
		\draw (0, 0) node(X3) [] {{\footnotesize\,${X}_{3}$\,}};
		\draw (1.2, 0) node(X4) [] {{\footnotesize\,$X_4$\,}};
		\draw (0, 1.0) node(X1) [] {{\footnotesize\,$X_1$\,}};
		\draw (1.2, 1.0) node(X2) [] {{\footnotesize\,${X}_{2}$\,}};
		\draw[-arcsq] (X1) -- (X2) node[pos=0.5,sloped,above] {};
		\draw[-arcsq] (X1) -- (X3) node[pos=0.5,sloped,above] {};
		\draw[-arcsq] (X1) -- (X4) node[pos=0.5,sloped,above] {};
		\draw[-arcsq] (X2) -- (X3) node[pos=0.5,sloped,above] {};
		\draw[-arcsq] (X2) -- (X4) node[pos=0.5,sloped,above] {};
		\draw[-arcsq] (X3) -- (X4)node[pos=0.5,sloped,above]{};
		\draw (0.6, -0.3) node(con1) [] {{\footnotesize\,(a) \,}};
		\end{tikzpicture}~~~~~~~~~~~~~~
        \begin{tikzpicture}[scale=1.4, line width=0.5pt, inner sep=0.2mm, shorten >=.1pt, shorten <=.1pt]
		\draw (0, 0.0) node(X3) [] {{\footnotesize\,${X}_{3}$\,}};
		\draw (1, 0.0) node(X4) [] {{\footnotesize\,$X_4$\,}};
		\draw (0.5, 0.6) node(L1) [circle, fill=gray!60,draw] {{\footnotesize\,$L_1$\,}};
		\draw (0, 1.2) node(X1) [] {{\footnotesize\,$X_1$\,}};
		\draw (1, 1.2) node(X2) [] {{\footnotesize\,${X}_{2}$\,}};
		\draw[-arcsq] (X1) -- (L1) node[pos=0.5,sloped,above] {};
		\draw[-arcsq] (X2) -- (L1) node[pos=0.5,sloped,above] {};
		\draw[-arcsq] (L1) -- (X3) node[pos=0.5,sloped,above] {};
		\draw[-arcsq] (L1) -- (X4) node[pos=0.5,sloped,above] {}; 
		\draw[-arcsq] (X1) -- (X2) node[pos=0.5,sloped,above] {};
		\draw[-arcsq] (X3) -- (X4) node[pos=0.5,sloped,above] {};
		\draw (0.5, -0.3) node(con1) [] {{\footnotesize\,(b) \,}};
		\end{tikzpicture}
		\caption{Neither structure induces independence constraints while they entail different GIN conditions over the observed variables $X_1,...,X_4$. (a) There are no latent variables in the graph. (b) There is a mediate latent variable in the graph.}
		\label{fig:furture-work} 
	\end{center}
\end{figure}

}
\bibliography{reference}

\end{document}